\documentclass[11pt]{article}

\usepackage[preprint]{acl}

\usepackage{times}
\usepackage{latexsym}

\usepackage[T1]{fontenc}

\usepackage[utf8]{inputenc}

\usepackage{microtype}

\usepackage{inconsolata}

\usepackage{graphicx}
\usepackage{booktabs}
\usepackage{fontawesome5}
\usepackage{tabularx}
\usepackage{makecell}
\usepackage{soul}

\usepackage{threeparttable}
\usepackage{adjustbox}
\usepackage{float}

\usepackage{siunitx}
\usepackage[table,dvipsnames]{xcolor}
\usepackage{ragged2e}
\usepackage{array}
\usepackage{subcaption}  %
\usepackage{multirow} %
\usepackage{longtable} %

\usepackage{enumitem}

\usepackage{placeins} 
\extrafloats{100} %
\usepackage{pdflscape}

\usepackage{tablefootnote}

\definecolor{lkSlight}{HTML}{F8D7DA}
\definecolor{lkFair}{HTML}{FFF3CD}
\definecolor{lkModerate}{HTML}{D1E7DD}
\definecolor{lkSubstantial}{HTML}{CFE2FF}
\definecolor{lkAlmostPerfect}{HTML}{E2D9F3}

\newcommand{\immersiveicon}{%
  \raisebox{-0.2em}{\includegraphics[height=1.25em]{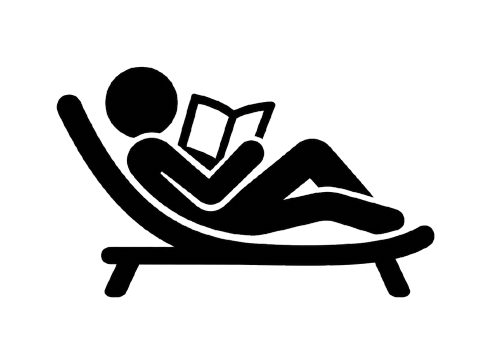}}%
}

\newcommand{\closeicon}{%
  \raisebox{-0.2em}{\includegraphics[height=1.35em]{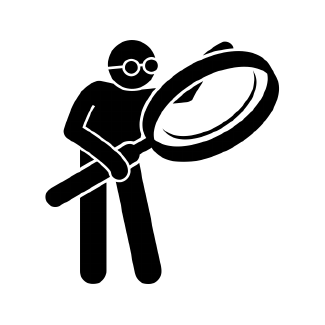}}%
}

\newcommand{\questionref}[1]{\hyperref[#1]{#1}}

\makeatletter
\newcommand*{\codelabel}[1]{%
  \textbf{#1}%
  \def\@currentlabel{#1}%
  \label{#1}%
}
\makeatother

\usepackage[most]{tcolorbox}
\tcbuselibrary{breakable}

\definecolor{srcbg}{HTML}{003380}

\definecolor{mtbg}{HTML}{68049A} %
\definecolor{htbg}{HTML}{BC096F} %

\newtcbox{\mtrbox}{%
  on line,
  colback=mtbg,
  colframe=mtbg,
  arc=2pt,
  boxrule=0pt,
  boxsep=0.5pt,
  left=2pt,
  right=2pt,
  top=1pt,
  bottom=1pt,
}

\newtcbox{\htrbox}{%
  on line,
  colback=htbg,
  colframe=htbg,
  arc=2pt,
  boxrule=0pt,
  boxsep=0.5pt,
  left=2pt,
  right=2pt,
  top=1pt,
  bottom=1pt,
}

\newtcbox{\srcrbox}{%
  on line,
  colback=srcbg,
  colframe=srcbg,
  arc=2pt,
  boxrule=0pt,
  boxsep=0.5pt,
  left=2pt,
  right=2pt,
  top=1pt,
  bottom=1pt,
}

\newcommand{\htr}{\htrbox{\textcolor{white}{HT}}}
\newcommand{\mtr}{\mtrbox{\textcolor{white}{MT}}}
\newcommand{\srcr}{\srcrbox{\textcolor{white}{SRC}}}

\colorlet{goodspan}{teal}
\colorlet{poorspan}{purple}
\newcommand{\good}[1]{\textbf{\textcolor{goodspan}{#1}}}
\newcommand{\poor}[1]{\textbf{\textcolor{poorspan}{#1}}}

\newcommand{\chunkicon}{\raisebox{-0.05em}{\scalebox{0.85}{\faIcon{layer-group}}}}

\newcommand{\name}{\textsc{\small LAIT}}

\newcommand{\laiticon}{%
  \raisebox{-0.15em}{\includegraphics[height=1.05em]{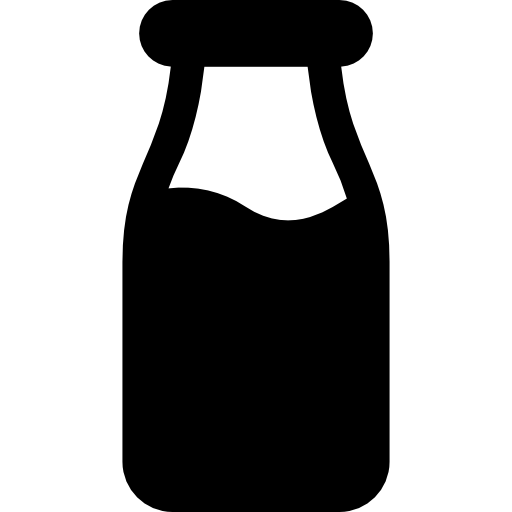}}%
  }

\makeatletter
\newcommand{\labeltablepanel}[2]{%
  \begingroup
  \def\@currentlabel{\thefigure#2}%
  \label{#1}%
  \endgroup
}
\makeatother

\title{ AI translation of literary texts is `fine', but readers still prefer human translations}

\definecolor{sfublue}{HTML}{1B6B8A}    %
\definecolor{uqampurple}{HTML}{7B4FA3}  %
\definecolor{msrgold}{HTML}{C8963E}     %

\newcommand{\affsfu}{\textsuperscript{\textcolor{purple}{\faBook}}}        %
\newcommand{\affuqam}{\textsuperscript{\textcolor{sfublue}{\faHatWizard}}} %
\newcommand{\affmsr}{\textsuperscript{\textcolor{msrgold}{\faMagic}}}        %
\author{
  \textbf{Yves Ferstler\affuqam},
  \textbf{Adam Podoxin\affsfu},
  \textbf{Ty Brassington\affsfu},
  \textbf{Gaëlle Laperrière\affuqam},
  \\
  \textbf{Roman Grundkiewicz\affmsr},
  \textbf{Marie-Jean Meurs\affuqam},
  \textbf{Maite Taboada\affsfu},
  \textbf{Marzena Karpinska\affsfu}
  \\[4pt]
  \affsfu Simon Fraser University \quad
  \affuqam Université du Québec à Montréal \quad
  \affmsr Microsoft
  \\[4pt]
  \small{
    \textbf{Correspondence:} 
    \href{mailto:ferstler.yves@courrier.uqam.ca}{ferstler.yves@courrier.uqam.ca}, \href{mailto:karpinsk@sfu.ca}{karpinsk@sfu.ca}
  }
}

\begin{document}
\maketitle

\begin{abstract}

AI translation of literary works is increasingly common.
While the content may be rendered adequately, we do not know enough about how readers experience it in terms of immersiveness and literary effect--aspects poorly captured by automatic metrics or human evaluation targeting fluency and adequacy.
We ask 15 avid readers to compare recently published human translations (HT) to machine translations (MT) generated with an agentic language model-based pipeline, for 15 recent novels in French, Polish, and Japanese translated into English.
Readers evaluated \textasciitilde8K-word excerpts in two conditions: \textit{immersive reading} of the whole excerpt (30 comparisons) and \textit{close reading} of 386 aligned HT--MT chunk pairs (772 comparisons), with two readers per book. %
Overall, readers find MT ``fine'', but prefer HT (descriptively at the excerpt level, 19/30, and significantly at the chunk level, 522/772) for its ease, clarity, and immersive nature.
Readers' highlights show that MT's quality varies more within one book than HT's does. 
Crucially, readers cannot reliably tell the two apart (17/30 guess correctly) and tend to prefer the version they believe to be human.
Automatic metrics, including LLM-as-a-judge approaches, fail to recover reader preferences and favor MT.
We release \laiticon \name{} ({\small Literary AI Translation}), a reader-centered evaluation dataset with 1K reader comments, 2K judgments and preference ratings, and 7.2K span-level annotations, along with our evaluation protocol and supporting interface.

\begin{table}[h]
\centering
    \renewcommand{\arraystretch}{1.1}
    \begin{tabular}{c@{\hskip.1cm}l}
        \faGithub & \href{https://github.com/Yves575/lait}{\path{github.com/Yves575/lait}} \\
        \faGlobe & \href{http://lait.cs.sfu.ca/}{\path{http://lait.cs.sfu.ca/}} \\
    \end{tabular}
\end{table}

\end{abstract}
\section{Introduction}

Machine-translated literature is reaching real readership.\footnote{We use \emph{AI translation} and \emph{MT} interchangeably for LLM-based translation; questionnaires used \emph{AI} as it is likely the term readers are more familiar with.} 
For instance, Amazon introduced Kindle Translate for self-publishing authors \cite{amazon_staff_kindle_translate_2025}, while services such as GlobeScribe market AI translation to smaller publishers~\citep{globescribe_website}. 
Similarly, Veen Bosch \& Keuning, the largest Dutch trade publisher, announced AI-assisted translation \cite{creamer_dutch_2024}, and Japan's Shogakukan released an app offering AI-assisted translations of light novels to US and Canadian readers \cite{nishioka2024novelous}. 

\begin{figure*}[t]
    \centering
    \includegraphics[width=1.0\textwidth]{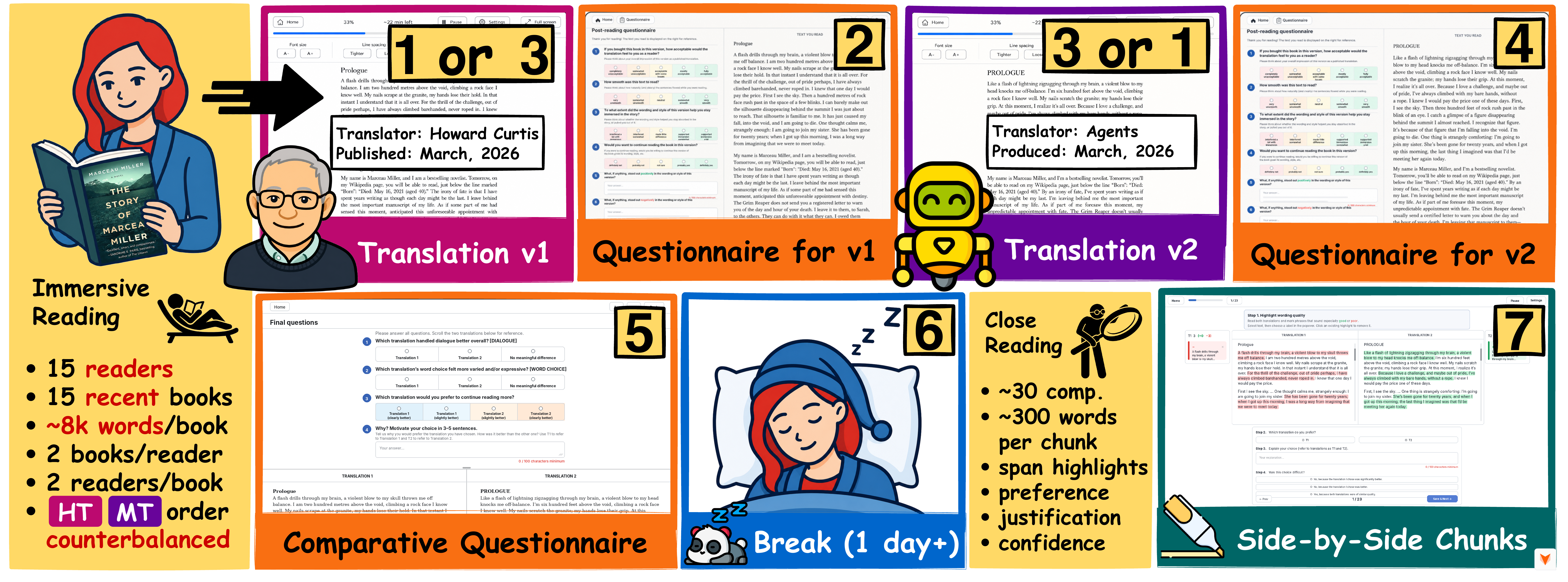}
    \caption{\textbf{Evaluation pipeline:} Avid readers of published fiction evaluate a \textbf{human translation (\htr)} and an \textbf{AI-generated machine translation (\mtr)} of an 8,000-word book excerpt. Participants \textbf{(1)} read the first translation, \textbf{(2)} complete a questionnaire, \textbf{(3--4)} repeat both steps for the competing translation, \textbf{(5)} compare both versions, \textbf{(6)} take a one-day break, and \textbf{(7)} evaluate 300-word chunks side by side, selecting the preferred version with a justification. Presentation order is counterbalanced, i.e., each book is evaluated in both \htr-first and \mtr-first conditions, and each reader evaluates two books, one in each order.}
    \label{fig:eval-pipeline}
\end{figure*}

This raises an interesting question. We know that readers sometimes even prefer fine-tuned AI fiction to expert human writing, at least on short excerpts \citep{chakrabarty2025readers}. So how does AI \emph{translation} shape the reading experience and the perception of the author's work compared to a professional human translation, especially once readers stay with the text for 25--30 pages?
Readers respond to rhythm, emotions, voice, and deliberate word choices \cite{reception-trans2013, eyetrack-experience}, but this experience cannot be captured by current evaluation setups which often concentrate on small units \cite{karpinska-iyyer-2023-large, wu-etal-2025-perhaps, zhang-etal-2025-litransproqa} and break the immersion with segment-level ratings \cite{kocmi-etal-2025-findings} or error annotations \cite{freitag-etal-2021-experts}. 
To capture the value of machine literary translation, we need to pose this question directly to the \textit{reader} \cite{carpuat-etal-2025-interdisciplinary}.

To fill this gap, we recruit 15 avid readers and ask them to read and compare 8K-word-long excerpts (\textasciitilde30 book pages) translated from Japanese, Polish, and French into English\footnote{We choose new books translated recently (2025--2026) into English to mitigate the training data contamination issue.}  by professional literary translators (HT) and an agentic LLM-based pipeline (MT), reading each version in full and uninterrupted, as they would a published book (\textit{immersive reading}).
Then, we ask them to look closer at both texts in 300-word chunks presented side by side and to highlight phrasing they judge as \good{good} or \poor{poor} before indicating which version they prefer (\textit{close reading}). Crucially, besides ratings, preference choices, and chunk-level highlights, we also ask the readers to provide detailed comments and guess which text is MT (\autoref{fig:eval-pipeline}). 

Overall, readers show a slight preference for HT, which is significant under \textit{close reading} (522/772) but only descriptive under \textit{immersive reading} (19/30). 
However, readers do not completely reject MT as they often find it readable and sometimes preferred. 
Notably, chunk-level span annotations show that MT quality varies much more within a single book than HT, and its overall success depends more on the book than the source language. 
Reader comments suggest that HT is praised more for immersion, naturalness, and literary rendering (e.g., ``\textit{having soul}'' (HT) vs.\ ``\textit{translation for dummies}'' (MT)). MT sometimes wins due to readability or local word choices.

Readers also struggle to identify MT reliably. 
Their accuracy remains close to chance even after reading both excerpts (17/30). 
Their explanations show that they are misled by common AI tells \cite{russell-etal-2025-people}, such as em-dashes or assumptions about AI not using swear words. 
Finally, automatic metrics, including LLM-as-a-judge approaches, fail to recover reader preferences and \emph{all} favor MTs.

\noindent To summarize, our contributions are threefold:

\begin{itemize}[leftmargin=2.1em,labelwidth=1.3em,labelsep=0.2em,align=left]
    \item[\faFile] \textbf{\textsc{Data:}} We introduce \laiticon \name, a reader-annotated dataset of 15 \textsc{\small FR/PL/JA$\rightarrow$EN} novel openings (2025--2026), each aligned across source, published HT, and agentic MT at the excerpt, chunk, and paragraph levels. Two readers per book gave 2K ratings and preferences, 1K comments, and 7.2K span-level annotations; an unannotated 16-book development set adds 1.7K paragraph-level alignments across five candidate MT systems.

    \item[\faSearch] \textbf{\textsc{Analysis:}} We show that modern literary MT into English can be readable and hard to detect, but its quality varies within a book more than HT, and readers still value~HT.
    \item[\faGraduationCap] \textbf{\textsc{Methodology:}} We introduce a reader-centered evaluation protocol combining \textit{immersive} and \textit{close} reading, and release our guidelines and annotation software to support reproducibility and future work.
\end{itemize}

\section{Data \& Methods}

\subsection{Constructing \laiticon \name}
\label{subsec:data}

\paragraph{Book selection.} 
We selected recently published, critically acclaimed fiction books originally written in French, Polish, and Japanese, with professional English translations published in 2025--2026, to mitigate data contamination~\cite{jacovi-etal-2023-stop, kocyigit2025overestimation}. 
Overall, we obtained \textbf{15 books} for human evaluation,\footnote{We limit our human evaluation to 15 books due to the prohibitive cost.} with five books per source language (see \autoref{tab:eval_books_chunks_compact_summary_stats}),
and reserved \textbf{16 books} for selecting the MT pipeline, for a total of \textbf{31 books} (see \S\ref{sec:app_mt_litdata} for the full book list).\footnote{All source books and translations were purchased and processed solely for research purposes. To comply with \textit{Fair Dealing} guidelines for research, we have retained no more than 10\% of any individual work.} 

\begin{table}[t!]
    \centering
    \resizebox{\linewidth}{!}{%
    \begin{tabular}{lccccccc}
        \toprule
        & \multicolumn{3}{c}{\htr}
        & \multicolumn{3}{c}{\mtr}
        & \srcr \\
        \cmidrule(lr){2-4} \cmidrule(lr){5-7} \cmidrule(lr){8-8}
        & \textsc{Tokens} & \textsc{Words} & \textsc{Sents}
        & \textsc{Tokens} & \textsc{Words} & \textsc{Sents}
        & \textsc{Tokens} \\
        \midrule
        \multicolumn{8}{l}{\faIcon{book-open} \textbf{Excerpts} \hfill (\textit{n=15})} \\
        \midrule
        \textsc{Mean} & 9,680 & 7,623 & 547 & 9,582 & 7,676 & 539 & 13,307 \\
        \textsc{St. Dev.} & 535 & 449 & 74 & 626 & 529 & 73 & 2,316 \\
        \textsc{Max} & 10,261 & 7,995 & 668 & 10,500 & 8,595 & 683 & 19,116 \\
        \textsc{Min} & 8,600 & 6,747 & 440 & 8,474 & 6,928 & 447 & 9,882 \\
        \midrule
        \multicolumn{8}{l}{\faIcon{file-alt} \textbf{Chunks} \hfill (\textit{n=386})} \\
        \midrule
        \textsc{Mean} & 376 & 296 & 21 & 372 & 298 & 21 & 442 \\
        \textsc{St. Dev.} & 66 & 49 & 8 & 73 & 57 & 8 & 130 \\
        \textsc{Max} & 611 & 486 & 51 & 644 & 530 & 50 & 870 \\
        \textsc{Min} & 118 & 100 & 4 & 121 & 102 & 4 & 79 \\
        \bottomrule
    \end{tabular}
    }
    \caption{Summary statistics for the evaluation dataset. Words are whitespace-delimited counts. Tokens are computed with \texttt{tiktoken} (\texttt{o200k\_base}). Sentence counts for \htr{} and \mtr{} use spaCy (\texttt{en\_core\_web\_trf}).}
    \label{tab:eval_books_chunks_compact_summary_stats}
\end{table}

\paragraph{Source and target languages.} We selected French, Polish, and Japanese as source languages because they span different language families, cultures, and literary traditions. We chose English as the target language to probe the upper bound of translation capability, since modern language models are exceptionally strong in English~\cite{ahuja-etal-2023-mega, zhu-etal-2024-multilingual}. Furthermore, the researchers' familiarity with all the languages facilitated data collection and analysis.

\paragraph{Excerpt extraction.} For each book, we selected an opening excerpt from the English human translation (HT) of \textasciitilde8K words ($\mu=7{,}623$, $SD=449$) under whitespace tokenization while preserving paragraph boundaries.\footnote{Excerpts contained an average of 543 sentences ($SD=72$), equivalent to \textasciitilde30 pages of a typical printed novel. This corresponds to 32,578 sentences or \textasciitilde1,800 pages read by all participants combined.}
This length supported \textit{immersive} reading while keeping the task feasible for readers' evaluation. 
We then manually identified the corresponding source excerpt to obtain source-target mapping with HT.

\paragraph{Dataset Excerpt Preprocessing.}
Before generating MTs, we manually inspected the excerpts and removed editorial notes and footnotes while preserving the original paragraph structure.
Additional normalization, e.g., unifying quotation marks, was applied to MTs and HTs before evaluation to minimize the influence of surface-level features (see \S\ref{sec:eval_interface} for details).

\paragraph{MT pipeline selection.}
To select a strong MT pipeline for the evaluation, we compared five configurations on a 16-book development set: two prompting pipelines, \textsc{\small P1} full-excerpt translation with post-editing and \textsc{\small P2} chunk-level translation with excerpt-level post-editing, each using either GPT-5.4 \cite{openai2026gpt54} or Gemini 3.1 Pro \cite{googledeepmind2026gemini31pro}, and \textsc{\small P3}, an AutoFiction-inspired agentic pipeline~\citep{pham_chang_desai_iyyer_2026} using Claude Code and Codex.\footnote{We chose GPT and Gemini models because their model families performed strongly in WMT25~\citep{kocmi-etal-2025-findings}; we included coding agents because they can operate over long contexts and multiple reference files, and have been shown to produce better long-context literary fiction \cite{huot2025agents}.}

We used a blind five-way preference task: six raters\footnote{Five of the raters were authors and one was a volunteer colleague.} each assessed 2--3 books. For each book, raters read five anonymized translations of the same excerpt presented in random order, selected the version they most preferred as a literary translation, and provided a short comment. The configurations were closely matched, with one exception: \textsc{\small P3} was chosen most often, but only by a small margin, while \textsc{\small P1} with GPT-5.4 was the clear outlier.\footnote{First-choice votes were: \textsc{\small P3} (5/16), \textsc{\small P1} with Gemini 3.1 Pro (4/16), \textsc{\small P2} with Gemini 3.1 Pro (3/16), \textsc{\small P2} with GPT-5.4 (3/16), and \textsc{\small P1} with GPT-5.4 (1/16).} We adopted \textsc{\small P3}, as it produces good translations, not because of evidence that it is decisively better. See \S\ref{app:app_mt_pipe} and \S\ref{app:agentic-pipe} for details.

\begin{figure*}[t]
    \centering
    \includegraphics[width=1.0\textwidth]{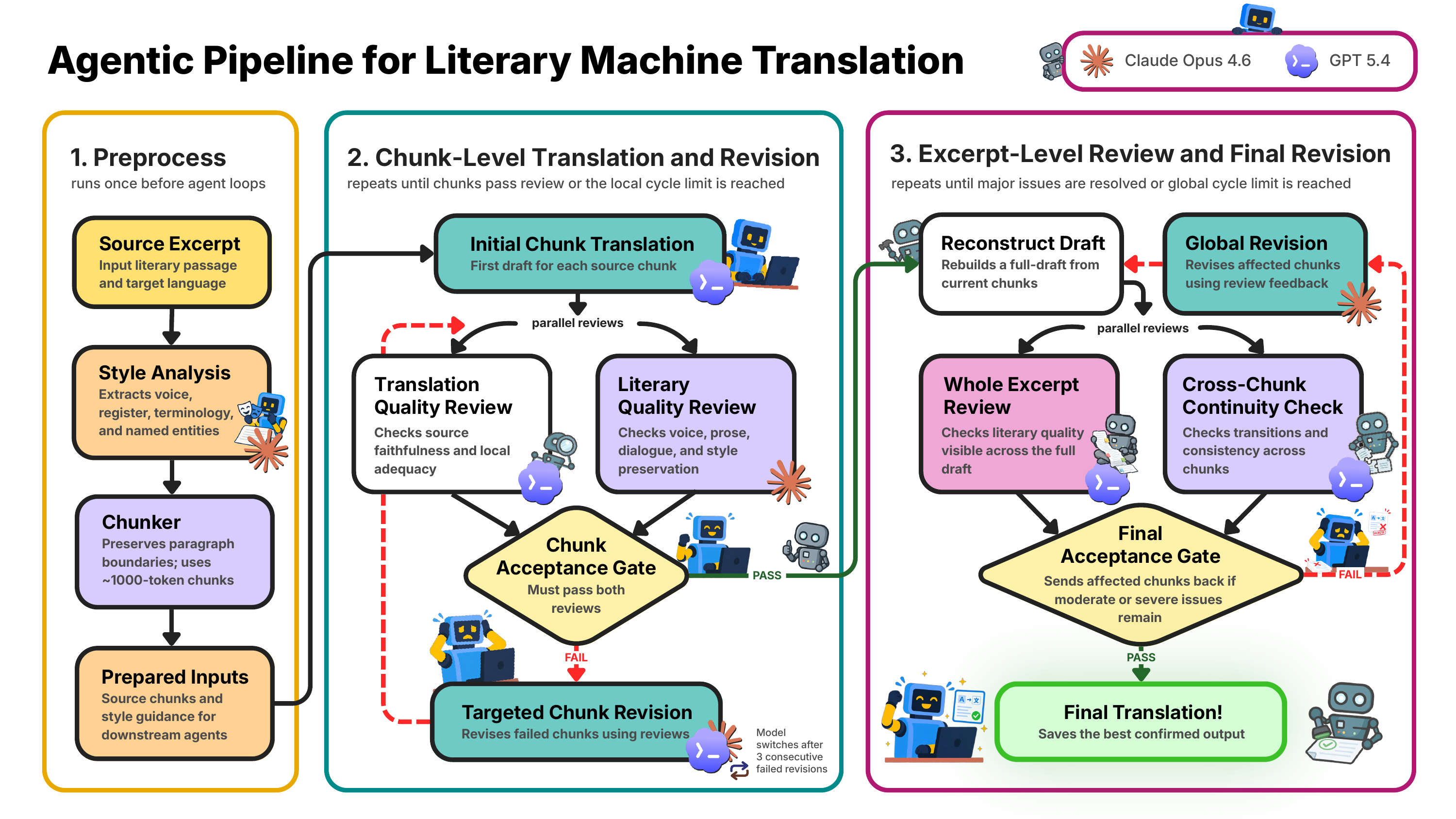}
    \caption{Agentic literary MT pipeline used in this study. Source excerpts are chunked and paired with style guidance, then translated and revised through chunk-level and full-draft review loops.}
    \label{fig:agentic-mt-pipeline}
\end{figure*}

\paragraph{Agentic pipeline.} We adapted the AutoFiction long-form fiction generation framework \cite{pham_chang_desai_iyyer_2026} to literary translation. Specifically, we deployed two coding agents, Claude Code with Claude Opus 4.6 \cite{anthropic2026claudeopus46} and Codex with GPT-5.4,\footnote{We set the reasoning effort to the maximum value, which is \texttt{max} for Claude Code and \texttt{xhigh} for Codex. Both agents were accessed via a monthly subscription at the cost of \$400 USD for the entire experiment.} to produce and validate the translations. Our pipeline consisted of three main stages (see~\autoref{fig:agentic-mt-pipeline}):

\begin{enumerate}
    \item \faIcon{cogs} \textbf{Preprocessing:} Claude Code analyzed the source excerpt to produce translation guidelines (e.g., register, terminology, translation memories). The source was then divided into 1K-token chunks while preserving paragraphs.
    \item \faIcon{file-alt} \textbf{Chunk-level:} Codex translated each chunk, which was then reviewed along two dimensions: (1) translation quality (e.g., faithfulness) using a modified version of {\small LiTransProQA} \citep{zhang-etal-2025-litransproqa} (Codex), and (2) literary quality review (e.g., voice, style) adapted from AutoFiction (Claude Code). Based on the output of these reviews, an \texttt{acceptance gate}\footnote{The gate uses the review rubrics described in \S\ref{app:agentic-pipe}; chunks pass only when the automatic reviewers find no major faithfulness or literary-quality problems.} determined whether the chunk proceeded to the next stage or was sent for revision. We allowed up to three local cycles total due to the high associated cost and small gains.\footnote{Development runs allowed three local cycles; subsequent evaluation and multilingual runs used two because the third cycle offered limited gains. \textit{Make Me Famous} retained its earlier three-cycle run. See Appendix~\ref{app:agentic-pipe} for details.}
    \item \faIcon{book-open} \textbf{Excerpt-level:} Revised chunks were merged into a full draft and reviewed for (1) global literary quality and (2) cross-chunk consistency (Codex). The results passed through a \texttt{final acceptance gate}, which triggered revision by Claude Code if minimum quality thresholds were not met.\footnote{The final gate uses the excerpt-level rubrics described in \S\ref{app:agentic-pipe}; drafts pass only when the automatic reviewers find no major global-quality or cross-chunk consistency problems.} We allowed up to two global cycles.
\end{enumerate}

\paragraph{Chunking.} To prepare the data for the chunk-level \textit{close reading}, 
we split each HT excerpt into \textasciitilde300-word chunks ($\mu=296$, $SD=49$) preserving the paragraph boundaries, yielding \textbf{386} chunks across the 15 evaluation books. We then aligned these chunks using Gemini 3.1 Pro to the corresponding chunks from MT and source text. All alignments were verified manually.

\subsection{Human Evaluation Protocol}
\label{subsec:hum-eval}

\begin{figure}[t]
    \centering
    \includegraphics[width=1\columnwidth]{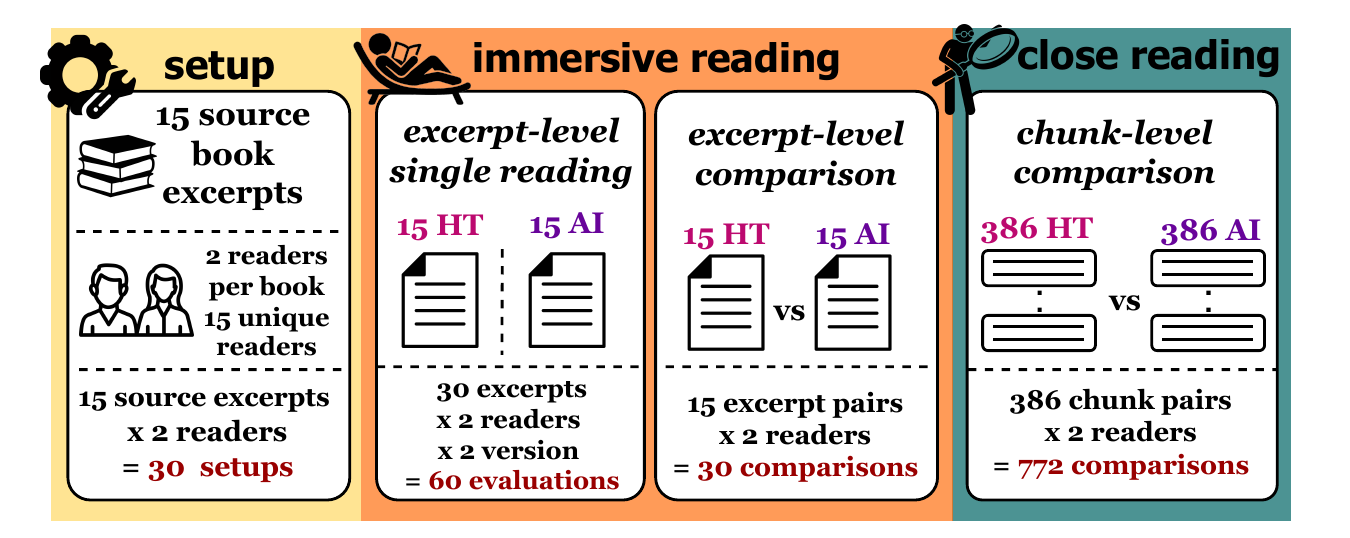}
    \caption{Study design and evaluation counts. Two readers evaluated each of the 15 book excerpts (30 book-reader evaluations).}
    \label{fig:study_counts}
    \vspace{-18pt}
\end{figure}

\paragraph{Overview.} We evaluated 15 book excerpts (\textasciitilde8K words) translated by professional translators (published HT) and by our pipeline (agentic MT). Each excerpt was evaluated by two readers, and each reader evaluated both versions of two books\footnote{Evaluations were consecutive, i.e., readers finished one book before starting the next.} (within-subjects), yielding 60 evaluations and 30 comparisons on the excerpt level (\textit{immersive reading}; \autoref{fig:study_counts}). After a one-day break, readers re-evaluated the same excerpts chunk by chunk, yielding 772 chunk-pair comparisons (\textit{close reading}). \autoref{fig:eval-pipeline} shows the full pipeline.

\paragraph{Participants.}
We recruited 15 avid readers through Upwork\footnote{\url{upwork.com}} (self-reported 12--120 books/year) who primarily read in English and were very comfortable with long literary texts, i.e., the \emph{target readers} of the published translations. Where possible, we matched readers to their preferred genres/story types. Participants were paid \$110 per book (\textasciitilde4h of work; \$27.50/h) plus a \$5--10 completion bonus; the total cost including bonuses and Upwork fees was \textbf{\$4K USD}. The protocol was approved by the Institutional Review Board. See \S\ref{sec:app_humeval} for details.

\begin{table}[t]
\centering
\footnotesize
\setlength{\tabcolsep}{4pt}
\renewcommand{\arraystretch}{1.1}

\resizebox{\linewidth}{!}{%
\begin{tabular}{@{}ll@{}}
\toprule
\textbf{Question} & \textbf{Type} \\
\midrule
\multicolumn{2}{@{}l}{\immersiveicon \textit{\textbf{Immersive reading} --- single-reading}}\\
Q1: Acceptability as a published translation     & 5L \\
Q2: Reading smoothness (fluency)                 & 5L \\
Q3: Wording/style support for immersion          & 5L \\
Q4: Willingness to keep reading this version     & 5L \\
Q5: Positive aspects of wording/style            & Open \\
Q6: Negative aspects of wording/style            & Open \\
Q7: Human- vs.\ machine-translated guess         & 2C \\
Q8: Confidence in guess                          & 5L \\
\addlinespace[2pt]
\multicolumn{2}{@{}l}{\immersiveicon \textit{\textbf{Immersive reading} --- comparison}}\\
Q1: Better dialogue handling                     & 3C \\
Q2: More varied / expressive word choice         & 3C \\
Q3: Version to continue reading                  & 4C \\
Q4: Justification for preferences                & Open \\
Q5: More likely AI-translated                    & 2C \\
Q6: Confidence in choice                         & 5L \\
Q7: Justification for the judgment               & Open \\
\addlinespace[2pt]
\multicolumn{2}{@{}l}{\closeicon \textit{\textbf{Close reading} --- per chunk pair}}\\
Q1: Highlight good / poor phrases                & Span \\
Q2: Preferred translation                        & 2C \\
Q3: Justify choice                               & Open \\
Q4: Difficulty of the choice                     & 3C \\
\addlinespace[2pt]
\multicolumn{2}{@{}l}{\textit{\textbf{Post-study} --- after both assigned books}}\\
Q1: Prior expectations for AI translation quality & Open \\
Q2: Surprise at MT-identification accuracy        & Open \\
Q3: Opinion of AI translation after the study     & Open \\
Q4: Other comments                                & Open \\
\bottomrule
\end{tabular}%
}

\caption{Questionnaire items for human evaluation. Scales: 5-point Likert (5L); categorical with 2/3/4 options (2C/3C/4C); good/poor span labels (Span); free text (Open). See \S\ref{sec:app_humeval} for the full wording.}
\label{tab:questions-reduced}
\end{table}

\paragraph{Procedure.} Each participant provided informed consent and completed a demographic survey on AI use and reading habits. Upon submission of a passphrase embedded in the guidelines the participants received access to the study interface designed to mimic e-book readers. They then completed the \textit{immersive reading} task, the \textit{close reading} task (unlocked after 24h), and a \textit{post-study} questionnaire (see \autoref{tab:questions-reduced} for questions overview).

\paragraph{\immersiveicon\ Immersive reading.}
Participants read one translation of the entire excerpt and answered eight \textit{single-reading} questions, seeing only that version; they then repeated this for the second translation. Only after both readings did they compare the two versions directly (seven comparative questions). Reading order was counterbalanced for each book, i.e., one reader started with HT and the other with MT, and each reader started with HT for one of their books and MT for the other. Participants were not told which version was MT.

\paragraph{\closeicon\ Close reading.} Participants then assessed aligned \textasciitilde300-word HT and MT chunks side by side: for each pair, they highlighted especially good or poor wording, chose the preferred translation, justified the choice, and rated its difficulty.

\paragraph{Questionnaires.}
We used four questionnaires. The \textit{immersive reading} task included two: (1) a single-reading questionnaire presented immediately after each reading, and (2) a comparative questionnaire presented after both readings. Together, these captured global reading experience. The \textit{close reading} task used (3) a questionnaire for each chunk pair, capturing local preference. Finally, after participants completed both books, we administered (4) a \textit{post-study} questionnaire with four open-ended questions about their thoughts on translations. See \autoref{tab:questions-reduced} for questions.

\paragraph{Quality control.} To ensure data quality, we required participants to submit a passphrase embedded in the guidelines before they received their credentials, confirming they had read the instructions. We also recorded screen-reading time and detected anomalies such as fast scrolling.\footnote{Two readers were excluded for insufficient comment detail and fast reading, and replaced by two others to keep the sample size at 15.} Finally, we ran the Pangram AI detector \citep{emi2024technicalreportpangramaigenerated} over participant comments to flag potentially AI-generated text.\footnote{No comments were flagged as AI-generated or AI-assisted.}

\section{Results \& Analysis}

\begin{figure}[t]
    \centering
    \includegraphics[width=1\columnwidth]{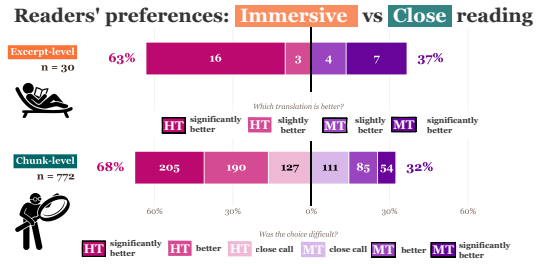}
    \caption{The distribution of readers' preferences between the \textbf{human translation (\htr)} and the \textbf{machine translation (\mtr)} after excerpt-level immersive reading and chunk-level close reading. The readers were judging which version they preferred and how strong the preference was.}
    \label{fig:chunk-res}
\end{figure}

\begin{figure}[t]
    \centering
    \includegraphics[width=1.0\columnwidth]{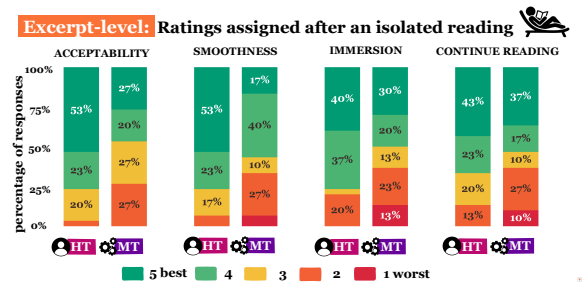}
    \caption{The distribution of readers' ratings after immersive reading of each excerpt \textbf{in isolation}. The readers were judging \textbf{acceptability} (whether it is acceptable as a published translation), \textbf{smoothness}, \textbf{immersiveness} (whether it supported immersion), and \textbf{continue} (whether they would continue reading).}
    \label{fig:single-res}
\end{figure}

\subsection{Readers' ratings and preferences}
\label{subsec:readers-pref}

\paragraph{Readers prefer HT, especially in \textit{close reading}.}
After reading both excerpt-level translations, participants chose which version they would prefer to continue reading (\questionref{co-Q3}).
Across the 30 excerpt-level judgments, 19 favored HT and 11 favored MT (exact 95\% CI for the HT share [44\%, 80\%]), with 16 clear preferences for HT compared to 7 clear preferences for MT (\autoref{fig:chunk-res}). The preference for HT was not statistically significant (order-adjusted HT probability = 63.4\%; OR = 1.73, 95\% CI [0.82, 3.65]; \textit{p}=.148).\footnote{Unless otherwise stated, statistical tests in this section use mixed-effects regression models with random intercepts for reader and book, and, where supported by a likelihood-ratio test, their interaction, with a model family appropriate for the response variable type. When a mixed-effects model is singular, we report the corresponding simpler model. See \S\ref{sec:stat-test-results} for details.}

Preference for HT is clearer in \textit{close reading} (\questionref{ch-Q2}) with 522 of 772 chunk-level judgments favoring HT, including 205 strong preferences, compared to 250 judgments favoring MT, including 54 strong preferences (\autoref{fig:chunk-res}). The chunk-level HT preference was significant (OR = 3.28, 95\% CI [1.41, 7.59], \textit{p}=.0056), and remained significant when preference strength was modeled directly (signed mean = 0.91, 95\% CI [0.24, 1.58], \textit{p}=.010).\footnote{Reading order (HT-first vs.\ MT-first) was included as a predictor in the preference models and was not significant (excerpt-level: \textit{p}=.705, chunk-level: \textit{p}=.327).}

\paragraph{HT is rated higher for quality, especially for smoothness and as a publishable translation.}
After reading each version, participants rated it on a 5-point scale for acceptability as a published translation, smoothness, support for immersion, and willingness to continue reading (\autoref{fig:single-res}).
HT received a large proportion of top ratings (4 or 5) and no rating of 1.
Statistical analysis shows that HT had significantly higher odds of receiving better ratings for acceptability as a published translation and for smoothness. Compared with MT, HT had 4.0 times the odds of a higher acceptability rating (\textit{p}=.0069; at most .043 across all six random-effects specifications, \autoref{tab:ratings-spec-grid}) and 4.3 times the odds of a higher smoothness rating (\textit{p}=.0029; at most .018).
The differences for immersion and willingness to continue reading favored HT but were not significant (\textit{p}=.0815 and .1150).\footnote{Models with random slopes for type (on reader or book) yield comparable results; see \autoref{tab:ratings-spec-grid}.}

\begin{figure}[t]
    \centering
    \includegraphics[width=1\columnwidth]{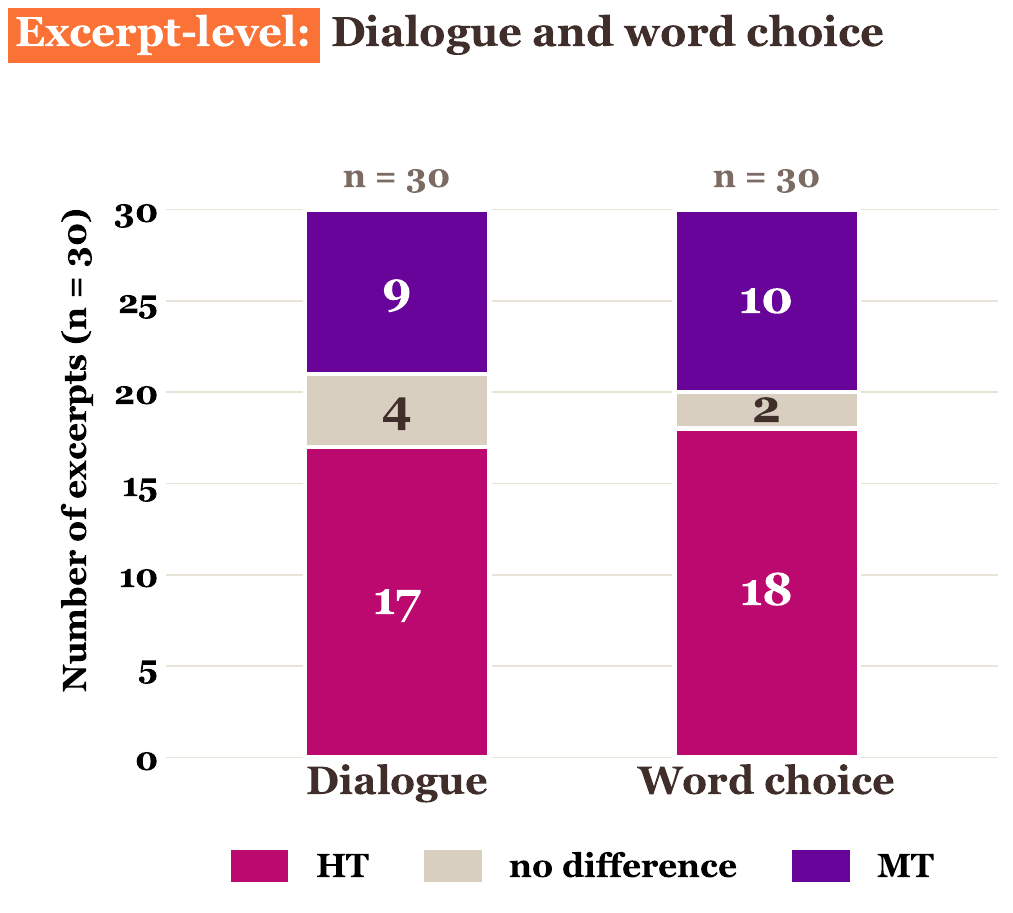}
    \caption{The distribution of readers' excerpt-level preferences for \textbf{dialogue handling} and \textbf{word choice} after reading both the \textbf{human translation (\htr)} and the \textbf{machine translation (\mtr)}. The readers were judging which version handled dialogue better overall and which version had more varied or expressive word choice.}
    \label{fig:excerpt-attribute-pref}
\end{figure}

\paragraph{HT is preferred on dialogue and word choice.}
At the excerpt level, readers also compared the versions on dialogue and word choice (\autoref{fig:excerpt-attribute-pref}), favoring HT on both (17 vs.\ 9 and 18 vs.\ 10, with 4 and 2 ties), though neither difference was significant (signed means = 0.27; \textit{p}=.109 and .131; \autoref{tab:preference-model-effects}).

\paragraph{HT preference does \textit{not} imply MT failure.}
The overall preference for HT does not mean that readers rejected MT across the board.
Of the 11 excerpt-level judgments that favored MT, 7 expressed a clear preference (\autoref{fig:chunk-res}).
After \textit{immersive reading}, 54\% of MT responses indicated moderate to strong willingness to continue reading (a rating of 4 or 5), compared with 66\% for HT (\autoref{fig:single-res}).
And in the \textit{close reading}, roughly one-third (32\%) of chunk-level choices favored MT.

\paragraph{MT preference varies by book but not by language.}
The one-third MT share in close reading was stable across source languages (French 32\%, Japanese 34\%, and Polish 31\%).\footnote{There was no evidence that language predicted MT preference after accounting for reading order and repeated judgments by reader and book (\textit{p}=.928).} By contrast, preference toward MT varied across books, from 4\% to 88\%, and book identity was strongly associated with chunk-level preference (\textit{p}<.001). The clearest case is \textit{Hooked: A Novel of Obsession}, preferred in MT 88\% of the time and consistently by both readers, suggesting that sometimes HT may fall short.

\begin{figure}[t]
    \centering
    \includegraphics[width=1\columnwidth]{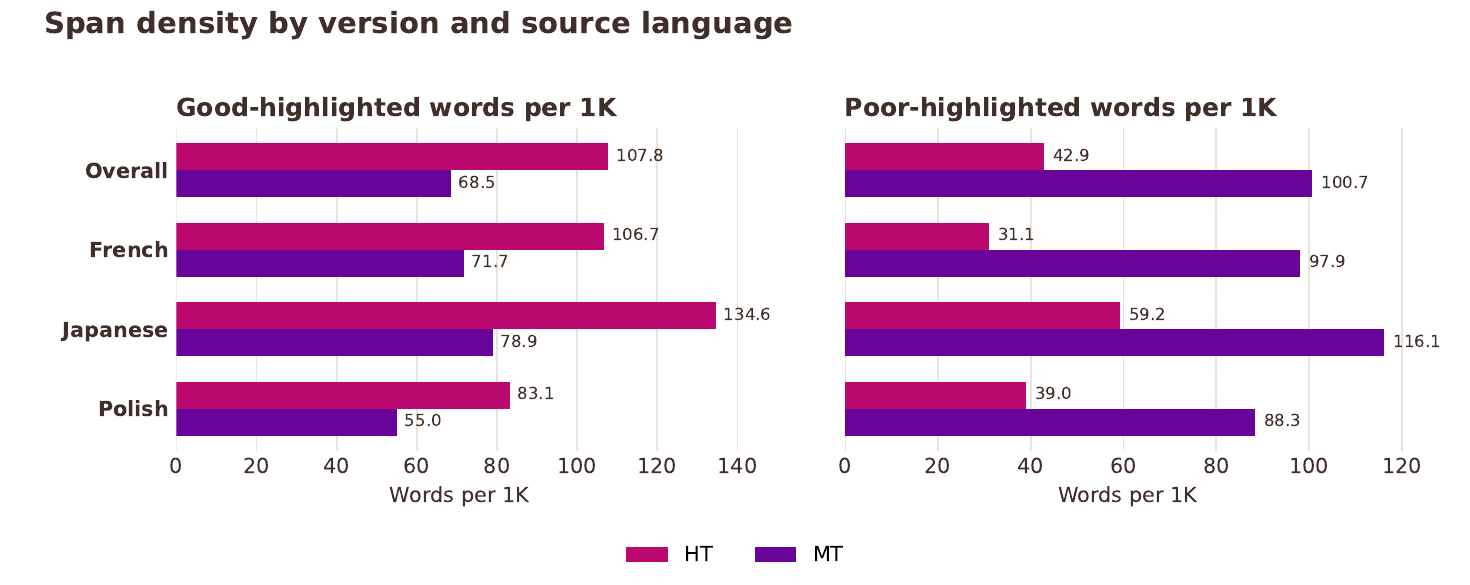}
    \caption{Distribution of \good{good} and \poor{poor} span highlights by source language and translation type.}
    \label{fig:span-density-by-language}
\end{figure}

\begin{figure}[t]
    \centering
    \includegraphics[width=\columnwidth]{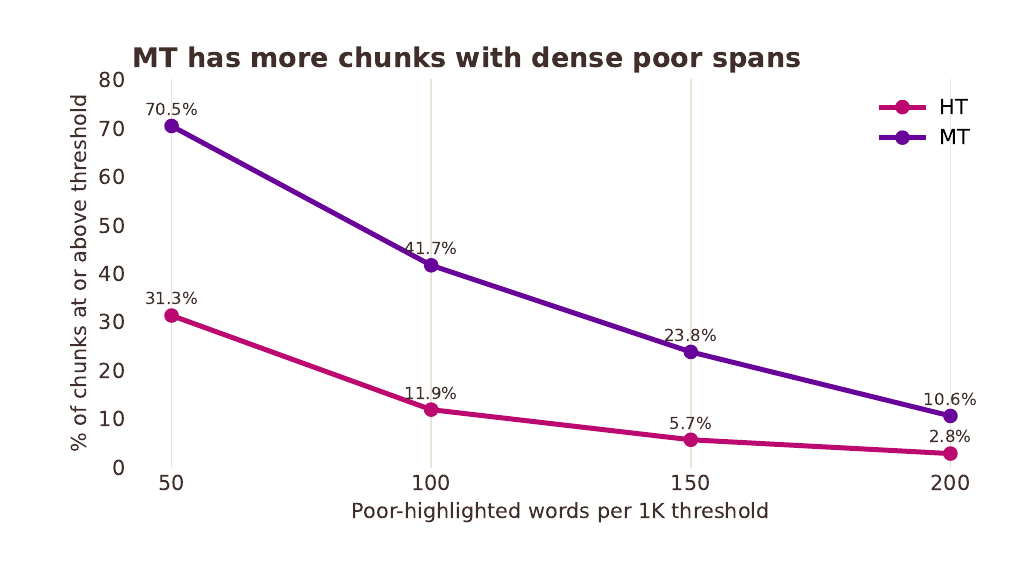}
    \caption{Share of close-reading chunks with dense poor-span evidence. For each threshold, the curve shows the percentage of HT and MT chunks with at least that many poor-highlighted words per 1K words. MT has substantially more high-density poor-span chunks.}
    \label{fig:span-spiky-poor-chunks}
\end{figure}

\begin{figure}[t]
    \centering
    \includegraphics[width=\columnwidth]{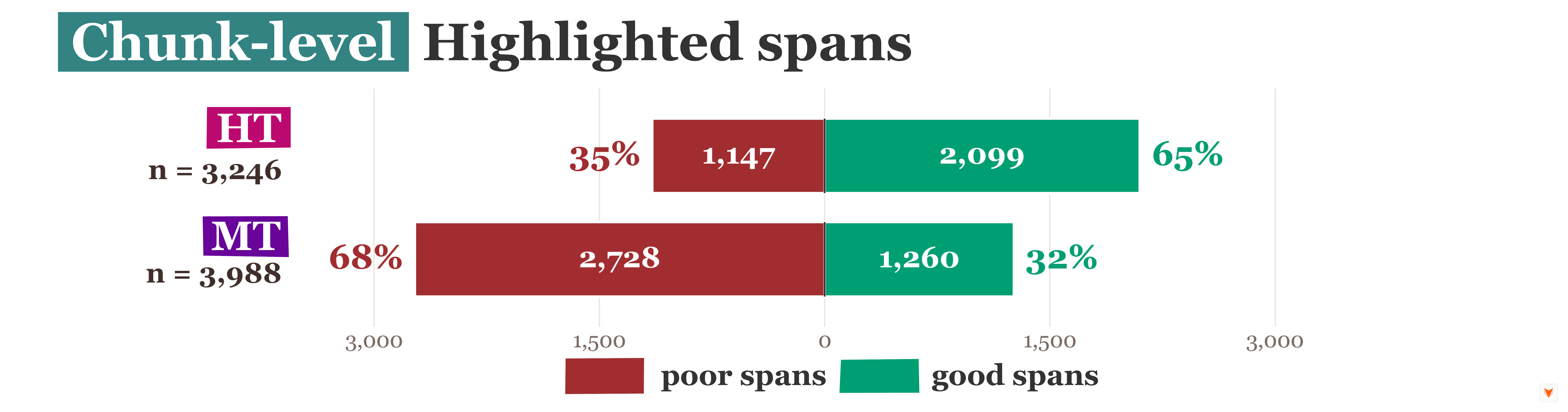}
    \caption{Distribution of span-level highlights in the close-reading task.}
    \label{fig:chunk-highlight}
\end{figure}

\begin{figure*}[t]
    \centering
    \includegraphics[width=\textwidth]{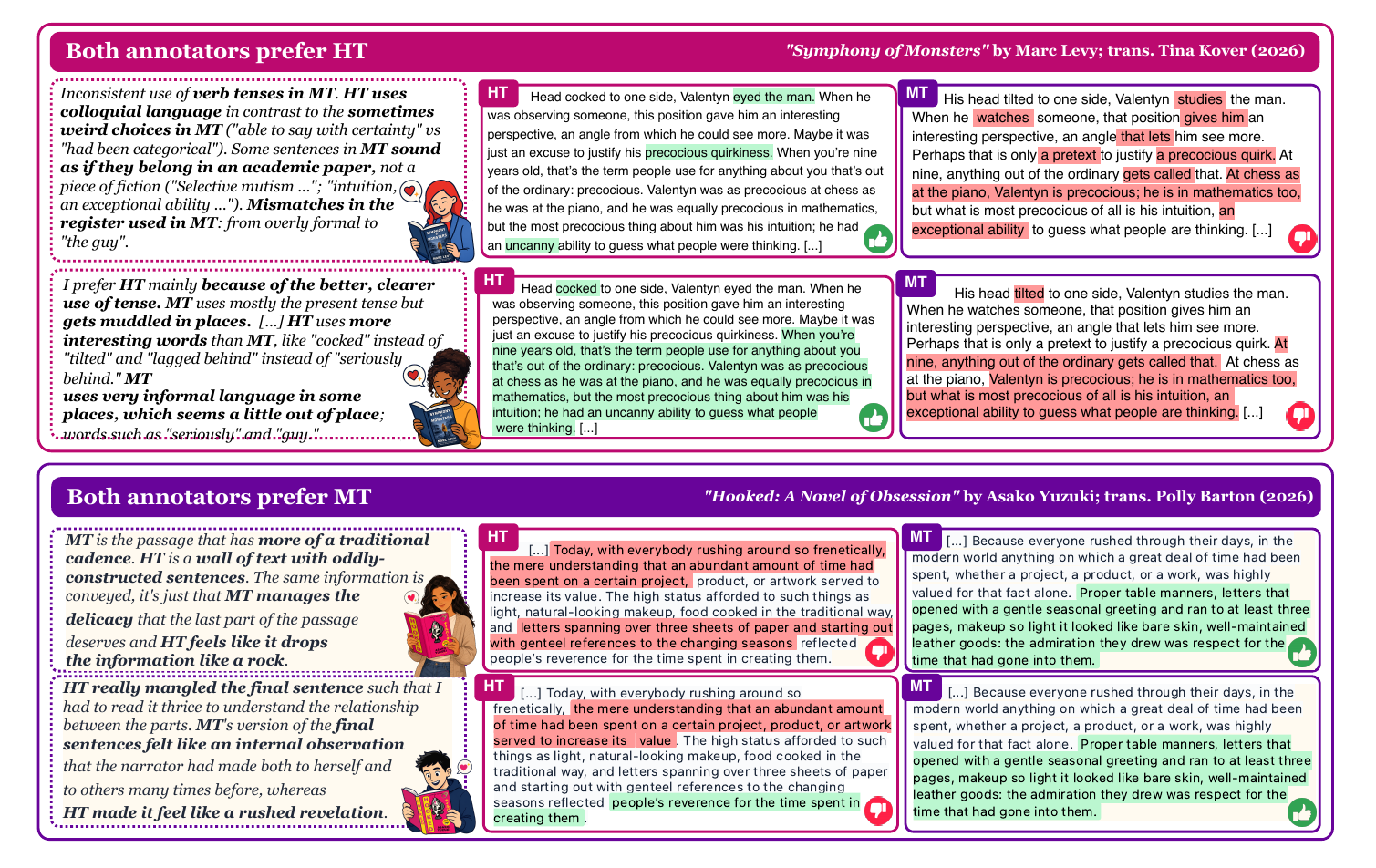}
    \caption{Examples of span-level preference evidence from the side-by-side chunks evaluation. Participants compared aligned HT and MT chunks, labeled spans as \good{good} or \poor{poor}, and chose the preferred translation. Green highlights indicate \good{good} spans and red highlights indicate \poor{poor} spans. The examples show cases where both readers chose the same translation, one favoring HT and one favoring MT, while marking different local evidence for their choices.}
    \label{fig:agreement-span-evidence}
\end{figure*}

\paragraph{Readers highlight more positive evidence in HT and more negative evidence in MT.}
The span annotations show the same pattern as the preference judgments, i.e., readers highlighted more positive evidence in HT than in MT (107.8 vs.\ 68.5 highlighted words per 1K words) and more negative evidence in MT than in HT (100.7 vs.\ 42.9), a gap that holds across all three source languages (\autoref{fig:span-density-by-language}).
Overall, HT received significantly more positive evidence spans (rate ratio = 1.69) and fewer negative spans (rate ratio = 0.43; both \textit{p}<.001; \autoref{fig:chunk-highlight}).
Negative evidence is also more concentrated in MT chunks with 41.7\% having at least 100 negative-highlighted words per 1K, versus 11.9\% of HT chunks (\autoref{fig:span-spiky-poor-chunks}).
\autoref{fig:agreement-span-evidence} (upper row) shows an example where both readers preferred HT, with their comments and highlights.

\paragraph{Span annotations closely track explicit preferences.}
For each chunk judgment, we computed a net span score per version (\good{good} minus \poor{poor} highlighted words); in cases with no ties, the version with the higher score matched the reader's stated preference in 691 of 755 judgments (91.5\%), rising to 692 of 722 (95.8\%) with span counts instead of word counts (both exact binomial \textit{p}<.001). Thus, span annotations directly capture the local textual evidence driving preferences.

\paragraph{Both MT and HT vary in quality, but MT has more variable weak spots.}
Aggregating readers' highlights by chunk within each book, \poor{poor}-wording evidence was more variable for MT than HT in 13 of 15 books (\textit{p}=.0074, paired sign test), and net quality (\good{good} minus \poor{poor} per 1K words) in 12 of 15 (\textit{p}=.0352).\footnote{Chunks containing both \good{good} and \poor{poor} spans were uncommon, but slightly more common for MT (6.7\%) than HT (3.9\%).}

The excerpt-level show similar pattern. MT ratings span the full 5-point scale, whereas HT is never rated 1, and within-participant variability was directionally higher for MT though not significantly (\textit{p}=.285, \S\ref{sec:app_extra_humeval_res}). While readers sometimes find MT smooth or readable, they also encounter chunks (or entire excerpts) with markedly more visible issues.

\paragraph{Literary preference has shared and subjective components.}
Since every excerpt was evaluated by two readers, we further look at inter-reader agreement. Specifically, we look at whether paired readers preferred the same translation and whether they assigned similar ratings.
Excerpt preference was the most subjective case, where readers chose the same version in only 6 of 15 books (40.0\%; AC1 = $-0.120$). This was higher for chunk-level choices (63.7\%; AC1 = 0.355), and even more so once the chunks marked as \textit{similar quality} were excluded (72.2\%; AC1 = 0.546), with the highest agreement when both readers marked one version as \textit{significantly better} (79.5\%; AC1 = 0.667). Perhaps unsurprisingly, this suggests that disagreement is concentrated in global preferences and when the choice between the two translations is less apparent, while clearer local preferences are shared more often between the readers.

For single-reading quality ratings, overall agreement was moderate (AC2 = 0.426) but notably stronger for HT than for MT (0.616 vs.\ 0.216). It was clearest for concrete craft ratings of HT (\texttt{acceptability} 0.765, \texttt{smoothness} 0.725) and weakest for experiential ones on MT (\texttt{immersion} 0.057, \texttt{continue\_reading} 0.089; \S\ref{sec:app-agreement}).
Together, these results suggest readers agree on textual quality, especially locally, but diverge on how much an uneven text still functions for them as literature. Because each reader compared HT and MT directly (within-subjects), this disagreement is unlikely to reflect differences in scale use; we therefore view it as an integral part of literary judgment, combining a shared sensitivity to textual quality with reader-specific priorities \citep{plank2022problem, guerberof2024or, marco-etal-2025-reader}.

\subsection{What do we learn from readers' comments?}
\label{subsec:readers-comment}

\begin{figure*}[t]
    \centering
    \includegraphics[width=0.85\textwidth]{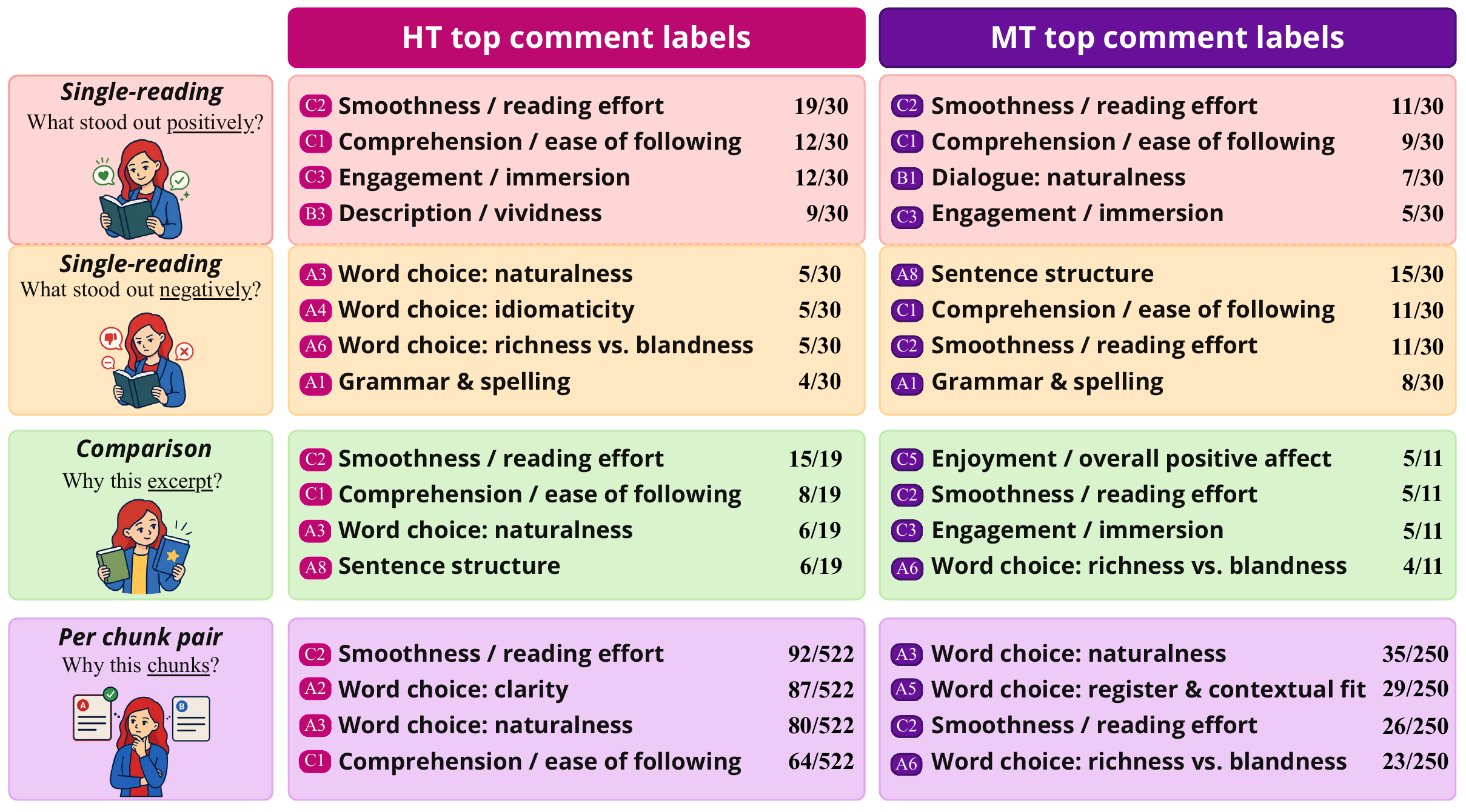}
    \caption{The most frequent labels in readers' comments about the \textbf{human translation (\htr)} and the \textbf{machine translation (\mtr)}. Rows separate single-reading praise, single-reading criticism, excerpt-level preference reasons, and chunk-level preference reasons. Comments could receive multiple labels.}
\label{fig:top-quality-labels}
\end{figure*}

\paragraph{Coding comments.}
We collected 180 comments during \textit{immersive reading} and 772 during \textit{close reading}.\footnote{For \textit{immersive reading}, 120 comments concern the good and bad qualities of each translation, 30 explain a preference choice, and 30 explain why a translation was selected as MT. For \textit{close reading}, all comments explain readers' preferences.} Following an inductive coding approach \cite{inductive}, we developed two coding schemas: (1) one for translation quality (29 categories) and (2) one for why a translation was judged AI-generated (10 categories).\footnote{The quality schema groups its 29 categories into four families: (A) language-level, (B) narrative-level, (C) reader experience, and (D) meta-translation.} The coding schema was developed by first running all comments through GPT-5.4 and Claude Opus 4.7, which helped us to obtain a list of candidate categories. These were used as support for the \textit{inductive coding} conducted by the researchers who further refined the schema.
Two researchers manually coded all \textit{immersive reading} comments, resolving disagreements. For \textit{close reading}, following prior work \citep{ziems-etal-2024-large}, we validated GPT-5.4 against a manually double-coded 10\% sample, then coded the rest automatically.\footnote{We measure agreement as the average Jaccard overlap between the human- and model-annotated label sets (multiple labels), which was 0.727 (high overlap). See \S\ref{sec:topic-model} for details.}  See \S\ref{subsec:app-anno-schema} for labels and details.

\paragraph{Both HT and MT are praised for readability, but HT receives more literary praise.}
In the \textit{single-reading} comments, readers \textbf{praised} both versions for basic readability (\autoref{fig:top-quality-labels}, top row).
Smoothness was the most frequent positive mention for both, though more often for HT (19/30 vs.\ 11/30), and comprehension showed a smaller gap (12/30 vs.\ 9/30), indicating that MT often cleared a basic readability bar.
The gap widened for more literary and experiential qualities: HT received more praise for engagement/immersion (12/30 vs.\ 5/30), and description/vividness was among its top labels (9/30).
MT was praised for dialogue naturalness (7/30), showing that readers did sometimes credit MT with literary strengths.
Thus the comments do not suggest MT was unreadable; rather, HT was more often credited with qualities that made a translation feel engaging, vivid, and literary.
After readers compared both excerpts directly, the reasons for choosing HT or MT became more distinct.

\providecommand{\bookreasonicon}{\raisebox{-0.15em}{\includegraphics[height=1.05em]{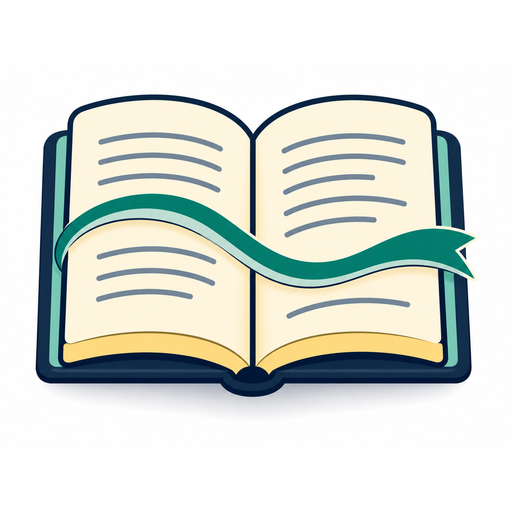}}}
\providecommand{\chunkreasonicon}{\raisebox{-0.15em}{\includegraphics[height=1.05em]{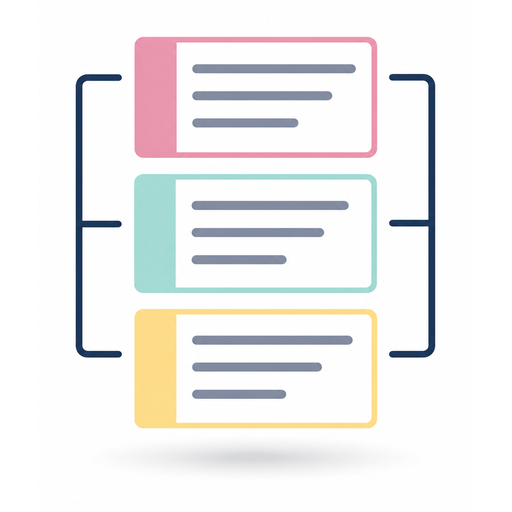}}}
\definecolor{reasonAone}{HTML}{FDEBD0}
\definecolor{reasonAtwo}{HTML}{FADBD8}
\definecolor{reasonAthree}{HTML}{F5CBA7}
\definecolor{reasonAfour}{HTML}{FCF3CF}
\definecolor{reasonAfive}{HTML}{E8DAEF}
\definecolor{reasonAsix}{HTML}{D7BDE2}
\definecolor{reasonAeight}{HTML}{D6EAF8}
\definecolor{reasonBone}{HTML}{D5F5E3}
\definecolor{reasonBtwo}{HTML}{D1F2EB}
\definecolor{reasonBfour}{HTML}{BFE3A4}
\definecolor{reasonCone}{HTML}{D6DBDF}
\definecolor{reasonCtwo}{HTML}{D4E6F1}
\definecolor{reasonCthree}{HTML}{C7CEEA}
\definecolor{reasonCfive}{HTML}{F9E79F}
\definecolor{reasonCsix}{HTML}{F9E79F}
\definecolor{reasonDfour}{HTML}{E8E8E8}
\definecolor{reasonPctA}{HTML}{6EA8DC}
\definecolor{reasonPctB}{HTML}{8DBBE8}
\definecolor{reasonPctC}{HTML}{AED0F0}
\definecolor{reasonPctD}{HTML}{CFE3F8}
\definecolor{reasonPctE}{HTML}{EAF3FC}
\providecommand{\reasonhl}[3]{\begingroup\setlength{\fboxsep}{0.7pt}\colorbox{#1}{#3}\textsuperscript{\textbf{#2}}\endgroup}
\providecommand{\hlAone}[1]{\reasonhl{reasonAone}{A1}{#1}}
\providecommand{\hlAtwo}[1]{\reasonhl{reasonAtwo}{WC-clar.}{#1}}
\providecommand{\hlAthree}[1]{\reasonhl{reasonAthree}{WC-natur.}{#1}}
\providecommand{\hlAfour}[1]{\reasonhl{reasonAfour}{A4}{#1}}
\providecommand{\hlAfive}[1]{\reasonhl{reasonAfive}{WC-regis.}{#1}}
\providecommand{\hlAsix}[1]{\reasonhl{reasonAsix}{A6}{#1}}
\providecommand{\hlAeight}[1]{\reasonhl{reasonAeight}{A8}{#1}}
\providecommand{\hlBone}[1]{\reasonhl{reasonBone}{B1}{#1}}
\providecommand{\hlBtwo}[1]{\reasonhl{reasonBtwo}{B2}{#1}}
\providecommand{\hlBfour}[1]{\reasonhl{reasonBfour}{B4}{#1}}
\providecommand{\hlCone}[1]{\reasonhl{reasonCone}{Comp.}{#1}}
\providecommand{\hlCtwo}[1]{\reasonhl{reasonCtwo}{Smooth.}{#1}}
\providecommand{\hlCthree}[1]{\reasonhl{reasonCthree}{Engag.}{#1}}
\providecommand{\hlCfive}[1]{\reasonhl{reasonCfive}{Enjoy.}{#1}}
\providecommand{\hlCsix}[1]{\reasonhl{reasonCsix}{C6}{#1}}
\providecommand{\hlDfour}[1]{\reasonhl{reasonDfour}{D4}{#1}}
\providecommand{\reasonlegend}[2]{\begingroup\setlength{\fboxsep}{0.7pt}\colorbox{#1}{#2}\endgroup}
\providecommand{\legAone}[1]{\reasonlegend{reasonAone}{#1}}
\providecommand{\legAtwo}[1]{\reasonlegend{reasonAtwo}{#1}}
\providecommand{\legAthree}[1]{\reasonlegend{reasonAthree}{#1}}
\providecommand{\legAfour}[1]{\reasonlegend{reasonAfour}{#1}}
\providecommand{\legAfive}[1]{\reasonlegend{reasonAfive}{#1}}
\providecommand{\legAsix}[1]{\reasonlegend{reasonAsix}{#1}}
\providecommand{\legAeight}[1]{\reasonlegend{reasonAeight}{#1}}
\providecommand{\legBone}[1]{\reasonlegend{reasonBone}{#1}}
\providecommand{\legBtwo}[1]{\reasonlegend{reasonBtwo}{#1}}
\providecommand{\legBfour}[1]{\reasonlegend{reasonBfour}{#1}}
\providecommand{\legCone}[1]{\reasonlegend{reasonCone}{#1}}
\providecommand{\legCtwo}[1]{\reasonlegend{reasonCtwo}{#1}}
\providecommand{\legCthree}[1]{\reasonlegend{reasonCthree}{#1}}
\providecommand{\legCfive}[1]{\reasonlegend{reasonCfive}{#1}}
\providecommand{\legCsix}[1]{\reasonlegend{reasonCsix}{#1}}
\providecommand{\legDfour}[1]{\reasonlegend{reasonDfour}{#1}}
\providecommand{\reasonpct}[2]{\begingroup\setlength{\fboxsep}{0.7pt}\colorbox{#1}{\textbf{#2}}\endgroup}
\providecommand{\pctA}[1]{\reasonpct{reasonPctA}{#1}}
\providecommand{\pctB}[1]{\reasonpct{reasonPctB}{#1}}
\providecommand{\pctC}[1]{\reasonpct{reasonPctC}{#1}}
\providecommand{\pctD}[1]{\reasonpct{reasonPctD}{#1}}
\providecommand{\pctE}[1]{\reasonpct{reasonPctE}{#1}}

\begin{table*}[t!]
    \centering
    \begingroup
    \scriptsize
    \setlength{\tabcolsep}{2pt}
    \renewcommand{\arraystretch}{0.88}
    \setlength{\aboverulesep}{0.25ex}
    \setlength{\belowrulesep}{0.25ex}
    \setlength{\abovecaptionskip}{3pt}
    \setlength{\belowcaptionskip}{0pt}
    \emergencystretch=1em

    \newcommand{\blockhead}[2]{%
        \multicolumn{3}{@{}l@{}}{\textbf{#1}\hspace{0.6em}{\scriptsize #2}}%
    }
    \newcommand{\excomment}[2]{%
        \textbf{Ex.} {\emph{#1}}\par
        \hfill{\scriptsize---{#2}}%
    }
    \newcommand{\sameas}[1]{\textit{\scriptsize same definition as #1}}

    \resizebox{\textwidth}{!}{%
    \begin{tabularx}{\textwidth}{@{}
        >{\RaggedRight\arraybackslash}p{0.23\textwidth}
        >{\RaggedRight\arraybackslash}p{0.25\textwidth}
        >{\RaggedRight\arraybackslash}X
        @{}}
        \toprule
        \textsc{Top labels} & \textsc{Definitions} & \textsc{Representative participant comment} \\
        \midrule

        \blockhead{Top 3 reasons ~\htr~ was chosen}{\bookreasonicon{} \textsc{Excerpt comparison}}\\[0.15em]
        {\scriptsize
            1. \pctA{78.9\%} \legCtwo{Smoothness}\par
            2. \pctD{42.1\%} \legCone{Comprehension}\par
            3. \pctE{31.6\%} \legAthree{Word choice: naturalness}
        }
        &
        {\scriptsize
            \legCtwo{\textbf{Smoothness}}: The prose flowed without friction, without consideration to understandability or clarity.\par
            \legCone{\textbf{Comprehension}}: The reader could understand what was happening: local
            sentence-level understanding and global scene/plot-level comprehension.\par
            \legAthree{\textbf{Word choice: naturalness}}: Natural English words use and phrasings.
        }
        &
        {\scriptsize
            \excomment{
                From the start of the story, HT was a lot \hlCone{clearer to me} and I think a
                lot of that has to do with punctuation and word choice. MT doesn't necessarily
                use obscure or uncommon words, but the translation \hlAthree{doesn't feel quite
                right} at times. For example, near the end of the first paragraph, MT states
                ``his legs refuse obedience and his eyes no longer come together in seeing.''
                I understand what this is supposed to mean, but HT's translation (when your
                legs won't do what they're told to and your eyes can't make out the world) is
                \hlCtwo{so much easier and natural to read}.
            }{Wiesław Myśliwski, \emph{Needle's Eye} (Bill Johnson, Trans.)}
        } \\
        \midrule

        \blockhead{Top 3 reasons ~\mtr~ was chosen}{\bookreasonicon{} \textsc{Excerpt comparison}}\\[0.15em]
        {\scriptsize
            1. \pctC{45.5\%} \legCfive{Enjoyment}\par
            2. \pctC{45.5\%} \legCtwo{Smoothness}\par
            3. \pctE{45.5\%} \legCthree{Engagement}
        }
        &
        {\scriptsize
            \legCtwo{\textbf{Smoothness}}: \sameas{row 1}.\par
            \legCthree{\textbf{Engagement}}: The reader was pulled into the story:
            losing reading awareness, attention focused on the scene.\par
            \legCfive{\textbf{Enjoyment}}: General reader satisfaction with clear statement: ``I liked it'',
            ``pleasant to read'', ``a pleasure'', ``I enjoyed both''.
        }
        &
        {\scriptsize
            \excomment{
                The language in MT is much more natural than HT and is clearer to understand.
                MT \hlCthree{pulls the reader in} with more vivid descriptions and is generally
                \hlCfive{well-written}. HT was boring to read and I felt my
                \hlCthree{mind wandering}.
            }{Haruo Yuki, \emph{The Ark} (Jim Rion, Trans.)}
        } \\
        \midrule

        \blockhead{Top 3 reasons ~\htr~ was chosen}{\chunkreasonicon{} \textsc{Chunk comparison}}\\[0.15em]
        {\scriptsize
            1. \pctB{17.6\%} \legCtwo{Smoothness}\par
            2. \pctC{16.7\%} \legAtwo{Word choice: clarity}\par
            3. \pctD{15.3\%} \legAthree{Word choice: naturalness}
        }
        &
        {\scriptsize
            \legCtwo{\textbf{Smoothness}}: \sameas{row 1}.\par
            \legAtwo{\textbf{Word choice: clarity}}: Clear descriptions, actions and intents due to words and phrasings choices. The reader is not trying to guess what is happening.\par
            \legAthree{\textbf{Word choice: naturalness}}: \sameas{row 1}.
        }
        &
        {\scriptsize
            \excomment{
                Despite one of two better-worded sections in MT, like ``but Cosima will have
                to be patient...'' instead of the repetitive ``Until Cosima grew up...'', HT
                is \hlAtwo{clearer} and \hlAthree{better worded overall}. The part about Cosima
                refusing help is \hlAtwo{clearer in HT}. The part about the cook being
                impressive in MT isn't clear, as it doesn't explain WHAT about her is
                impressive. HT makes it clear that it's her physique. The Latin phrase
                ``manu militari'' seems out of place in MT. ``Roughly shoved by a man in
                uniform'' is clearer. The last sentence is also rendered more clearly and
                \hlCtwo{flows more smoothly in HT}.
            }{Marc Levy, \emph{Symphony Of Monsters} (Tina Kover, Trans.)}
        } \\
        \midrule

        \blockhead{Top 3 reasons ~\mtr~ was chosen}{\chunkreasonicon{} \textsc{Chunk comparison}}\\[0.15em]
        {\scriptsize
            1. \pctD{14.0\%} \legAthree{Word choice: naturalness}\par
            2. \pctE{11.6\%} \legAfive{Word choice: register}\par
            3. \pctE{10.4\%} \legCtwo{Smoothness}
        }
        &
        {\scriptsize
            \legAthree{\textbf{Word choice: naturalness}}: \sameas{row 1}.\par
            \legCtwo{\textbf{Smoothness}}: \sameas{row 1}.\par
            \legAfive{\textbf{Word choice: register}}: Word choices match the context in terms of formality,
            period, dialect/local speech, genre, speaker, situation, or emotional context.
        }
        &
        {\scriptsize
            \excomment{
                MT sounds \hlAthree{slightly more natural}, but it also has phrases that stuck
                out to me as strange. Both have formatting issues. In MT, I would not have
                known that first line was a quote if it weren't next to HT. The address on
                the stationery in MT \hlCtwo{flows much better} in all caps than in HT. I did
                not know what ``a flowery balcony'' meant in HT. The final sentence in both
                is a little long and hard to follow, but MT translates the word as
                \hlAfive{``violin'' which makes more sense in context than HT's ``fiddle''}.
            }{Mikołaj Łoziński, \emph{My Name Is Stramer} (Antonia Lloyd-Jones, Trans.)}
        } \\
        \bottomrule
    \end{tabularx}%
    }

    \caption{Top labels behind reader preferences for ~\htr~ or ~\mtr~ in excerpt- and chunk-level comparisons. Percentages are computed separately within each system and task from positive labels assigned to the chosen translation. Participant comments are shown in full with translation labels normalized to HT/MT, and colored highlights correspond to the code chips in the top-label list.}
    \label{tab:quality-reason-preference-examples}
    \endgroup
\end{table*}

\paragraph{Readers choose HT because its fluency keeps them inside the story.}
In \textit{immersive reading}, readers who preferred HT for smoothness/reading effort (15/19), comprehension (8/19), natural wording (6/19), and sentence structure (6/19; \autoref{fig:top-quality-labels}, third row).
They described HT as \textit{``far easier to follow in terms of flow and word choice,''} especially when MT made it \textit{``difficult to determine who was speaking.''}

In \textit{close reading}, similarly, HT is chosen when its wording is smooth, clear, and natural enough to keep the passage moving as literary prose (92/522 smoothness/reading effort, 87/522 word-choice clarity, 80/522 natural wording; \autoref{fig:top-quality-labels}, fourth row).
For chunks where HT won, readers often also mentioned issues with MT (84\%, 439/522), with criticisms concentrated on word choice, sentence structure, comprehension, and smoothness.

\paragraph{Readers choose MT for engagement and effective local wording.}
In \textit{immersive reading}, readers who preferred MT described it as enjoyable, smooth, or even more expressive than the HT alternative (5/11 enjoyment, 5/11 smoothness/reading effort, 5/11 engagement/immersion, 4/11 word-choice richness; \autoref{fig:top-quality-labels}, third row).
One reader praised MT's \textit{``more intricate and varied''} wording and its \textit{``clearer picture of personalities and relationships''}.

In \textit{close reading}, readers reward particular wording that better matches the scene, register, or character voice (35/250 natural wording, 29/250 register or contextual fit, 26/250 smoothness, 23/250 word-choice richness; \autoref{fig:top-quality-labels}, bottom row). For instance, readers preferred \textit{``bento containers''} over \textit{``plastic containers,''}. 
Together, the excerpt-level and chunk-level comment evidence shows the similar patterns. MT can win when readers experience it as the more engaging excerpt overall, or when a local wording choice creates a better literary effect than the corresponding HT. 
More details in \S\ref{app:preference-comment-evidence}.

\paragraph{Negative comments highlight different issues in HT and MT.}
In \textit{single-reading} criticism, complaints about HT were diffuse and centered on local word choice (naturalness, idiomaticity, and blandness each appeared in 5/30 comments) followed by grammar, register, smoothness, and AI tells (4/30 each; \autoref{fig:top-quality-labels}, second row).
Complaints about MT were more concentrated around reading friction, specifically sentence structure (15/30), smoothness/reading effort (11/30), and comprehension/ease of following (11/30), with grammar and word naturalness also frequent (8/30 each).
Readers described MT as \textit{``abrupt and choppy,''} with \textit{``longwinded run-on sentences,''} or as forcing them to \textit{``stop and reread parts.''}
In six HT readings, readers found nothing negative to report, compared with only two MT readings.

\paragraph{Readers were surprised by the MT quality.} 
In the post-study questionnaire, most readers said they expected the MT to be worse or easier to detect (12/15), and some stated the MT was surprisingly strong (8/15). 
The readers also identified the remaining limitations of MT in literary and contextual qualities (8/15; e.g., \textit{``missing emotional import, poetic sense and failed to take into consideration that some things should be changed in translation based on the cultural and language differences in the output language''}). 
See \S\ref{sec:app_extra_humeval_res} for details.

\begin{figure}[htbp]
    \centering
    \includegraphics[width=1\columnwidth]{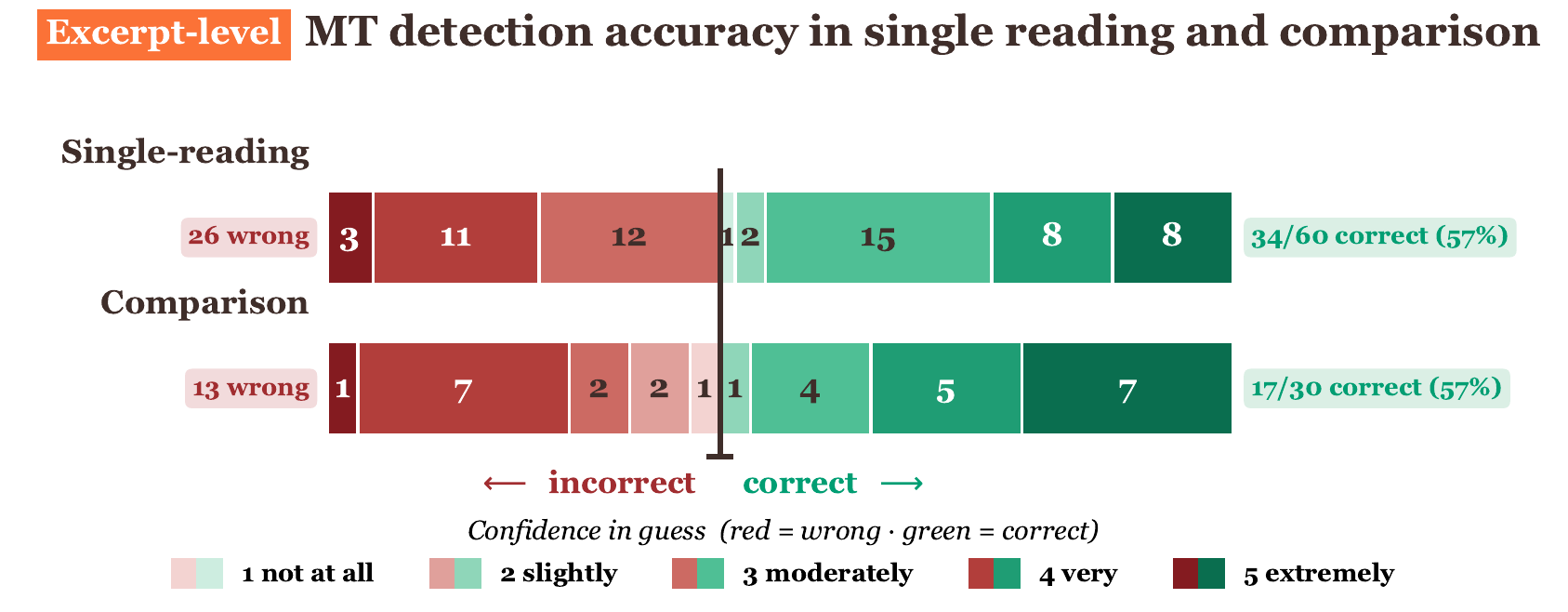}
    \caption{
        Machine-translation (MT) identification accuracy and confidence in the single-reading questionnaire (n=60) and comparative questionnaire (n=30).
        Colors represent confidence levels, from \emph{not at all} to \emph{extremely} confident.
    }
    \label{fig:mt-detection-accuracy-confidence}
\end{figure}

\subsection{Can readers detect machine translation?}
\label{subsec:readers-detection}

\paragraph{Readers cannot reliably tell MT from HT.}
In the single-reading questionnaire, readers correctly labeled the MT excerpt as machine-translated in 16 of 30 cases and the HT excerpt as human-translated in 18 of 30 cases. Overall, we do not find any evidence that the readers were able to distinguish MT from HT in single-reading above the chance level (34/60, \textit{p}=.413) with no difference between the MT and HT accuracy (\textit{p}=.557).

Moreover, even after seeing both translations, readers selected the actual MT version as more likely to be AI-translated only in 17 out of 30 pairs (\textit{p}=.585; \autoref{fig:mt-detection-accuracy-confidence}). While correct guesses were more confident after seeing both translations, that increase was not significant (\textit{p}=.129). Moreover, the high confidence did not guarantee correctness. After reading both excerpts, 8 of the 20 ``very'' or ``extremely'' confident guesses were wrong (40\%), including 1 of 8 ``extremely'' confident guesses (12.5\%). See \S\ref{sec:app_extra_humeval_res} for more details.

\paragraph{Readers prefer the translation they perceive as human.}
Readers' final MT guess was connected to their preferences; in 28 of 30 comparisons they judged their preferred translation to be the human one (all 11 readers who preferred MT believed it was human, as did 17 of 19 who preferred HT).
Only in two cases did the readers prefer HT while still judging it to be MT. In one case the reader was misled by \textit{em-dashes} (i.e., a strong belief that AI uses more em-dashes). In another one the reader noticed French structure in the HT for a book that was originally written in French, which convinced them that this translation, despite being better, is MT.

\paragraph{What cues do readers use to identify MT?}
Word choice was the most frequently cited cue, for both correct (8/17) and incorrect (9/13) guesses, making it non-discriminative. 
Correct guesses described wording as ``awkward,'' incorrect ones as ``generic'' or ``weird'' (\textit{``people just don't speak like that''}). 

More telling was whether the readers relied on concrete textual evidence or on ``folk theories'' about AI (i.e., the common idea of what AI writing is). 
The strongest separator was formatting (M4) such as scene dividers or paragraph breaks, which made it hard to follow. 
Another cue with stronger discriminative power was literal rendering (M1; 6/17 vs.\ 3/13, e.g., MT described as \textit{``like a literal dictionary translation''}) 
along with poor sentence flow or grammar (M3; 7/17 vs.\ 4/13, e.g., run-on and cut-off sentences).

\begin{figure}[t]
    \centering
    \includegraphics[width=1\columnwidth]{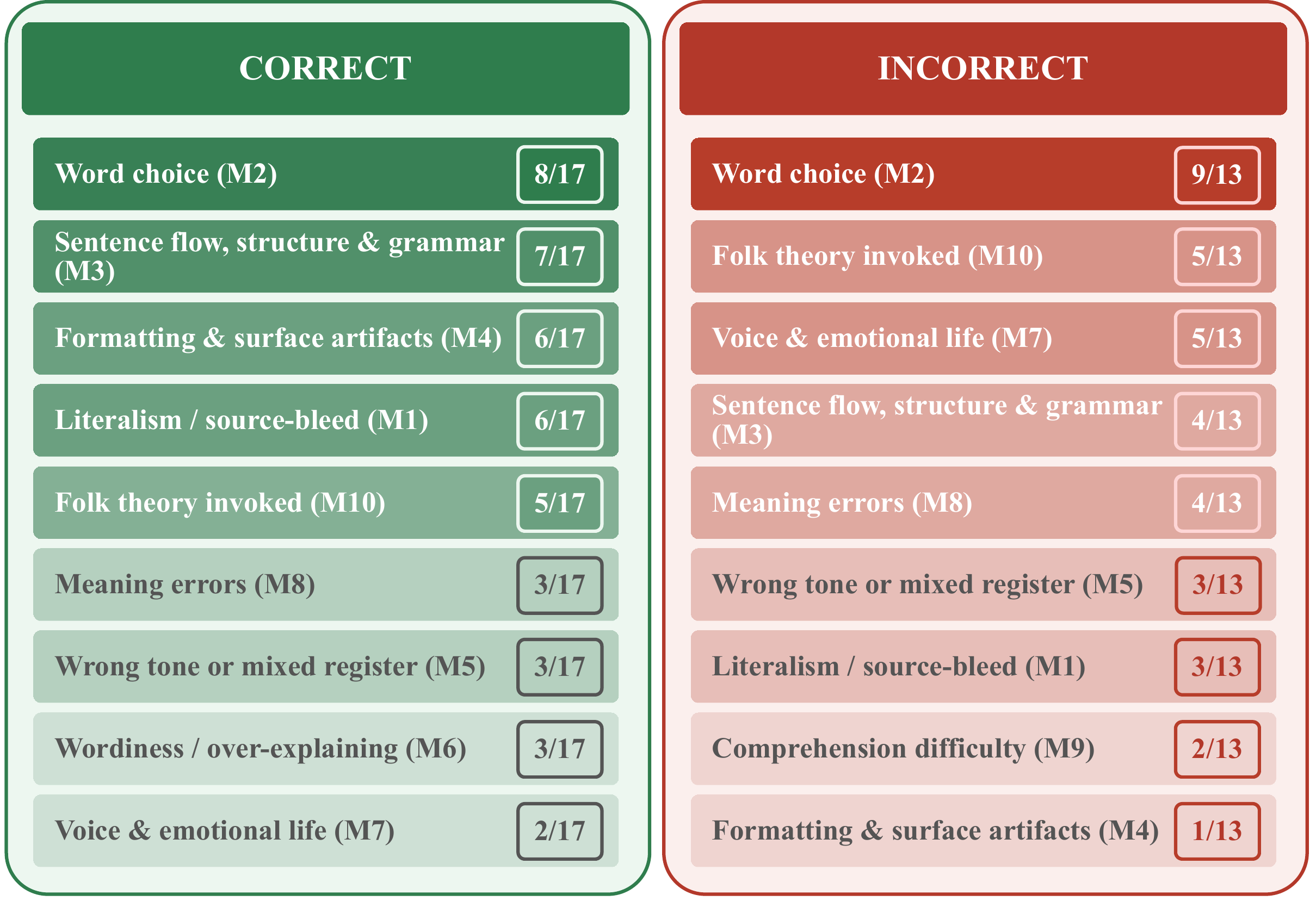}
    \caption{The cues readers mentioned when explaining which translation they believed was machine-translated (\mtr). Correct judgments identify the actual \mtr{} version; incorrect judgments identify the human translation (\htr) as \mtr{}. Comments could receive multiple cue labels.}
    \label{fig:cues-ai}
\end{figure}

The most deceptive were ``folk theories,'' which resulted in accuracy no better than chance (M10; 5/17 vs.\ 5/13). These include em-dashes perceived as \textit{``a classic sign of AI,''} or a belief that AI ``prefers to avoid'' swearing \cite{russell-etal-2025-people}. 
Least reliable were references to subjective literary quality, i.e., voice and emotional tone (M7) was cited mostly by readers who guessed wrong (2/17 vs.\ 5/13), who called the human translation ``flat'' and ``mechanical'' and so took it for the machine. See \S\ref{sec:app_extra_humeval_res} and \autoref{fig:cues-ai}.

\subsection{Can automatic metrics capture readers' preferences?}
\label{subsec:auto-metrics}
We compare readers' chunk-level preferences with three automatic metrics: \textsc{\small{MetricX-QE}} \cite{juraska-etal-2024-metricx}, \textsc{\small{COMETKiwi}} \cite{rei-etal-2022-cometkiwi}, and the LLM-as-a-judge framework \textsc{\small{LiTransProQA}} \cite{zhang-etal-2025-litransproqa}. 

\paragraph{Setup.} We compare chunk-level human judgments with automatic metrics. Since several metrics are constrained by context-window size, we use the \texttt{par3} aligner \cite{thai-etal-2022-exploring} to obtain paragraph-level source--HT--MT alignments for the evaluation data. We then score the paragraphs and map the scores to chunks containing these paragraphs. A chunk-level score was computed by taking the average of paragraph-level scores within each chunk. See \S\ref{app:auto-eval} for details.

\paragraph{Evaluation methods.} We test the three most popular evaluation methods: \textsc{\small MetricX-QE}, \textsc{\small COMETKiwi}, and  \textsc{\small LiTransProQA} with Gemini 3.1 Pro \cite{googledeepmind2026gemini31pro} as a judge. We additionally score the paragraphs and then compute the average chunk score with \textsc{\small COMETKiwi} and \textsc{\small MetricX-QE}.\footnote{We compute \textsc{\small MetricX-QE} over both chunk and paragraph as it is likely better adapted to shorter lengths. For \textsc{\small LiTransProQA} we report only chunk-level evaluation due to the cost and since the model can handle 300-word texts.}

\paragraph{Can the difference between MT and HT be captured with automatic metrics?} 
Across the close-reading chunks, all automatic metric variants prefer MT over HT
({\small \textsc{LiTransProQA}$\uparrow$ MT:~0.996, HT:~0.920};
{\small \textsc{MetricX-QE}$\downarrow$ MT:~8.347, HT:~8.657};
{\small \textsc{COMETKiwi} paragraph mean$\uparrow$ MT:~0.760, HT:~0.722};
{\small \textsc{MetricX-QE} paragraph mean$\downarrow$ MT:~5.184, HT:~5.671}).
Kendall $\tau$~\cite{kendall1938rank} correlation with human judgments at chunk-level is weak to negative for all metric variants
(\textsc{\small LiTransProQA} {\small $\tau=-0.124$},
\textsc{\small MetricX-QE} {\small $\tau=-0.075$},
\textsc{\small COMETKiwi paragraph mean} {\small $\tau=-0.318$},
\textsc{\small MetricX-QE paragraph mean} {\small $\tau=-0.216$}), and is identical whether computed over all judgments or only those where readers agree, since clear human agreement usually favors HT while metrics favor MT.

\begin{table*}[t]
    \centering
    \small
    \setlength{\tabcolsep}{6pt}
    \begin{tabular}{lrrrrr}
        \toprule
        Metric & Chunks & HT mean & MT mean & Oriented HT--MT delta & 95\% CI \\
        \midrule
        LiTransProQA & 386 & 0.920 & 0.996 & -0.076 & $\pm$ 0.017 \\
        MetricX-QE & 385 & 8.657 & 8.347 & -0.309 & $\pm$ 0.128 \\
        COMETKiwi (paragraph mean) & 377 & 0.722 & 0.760 & -0.038 & $\pm$ 0.004 \\
        MetricX-QE (paragraph mean) & 385 & 5.671 & 5.184 & -0.486 & $\pm$ 0.103 \\
        \bottomrule
    \end{tabular}
    \caption{Automatic evaluation of the aligned 300-word human-evaluation chunks: professional \textsc{HT} vs. agentic (P3) \textsc{MT}. Positive oriented deltas favor \textsc{HT}; negative deltas favor agentic/P3 \textsc{MT}. Paragraph-mean rows use mapped par3 paragraph scores.}
    \label{tab:appendix_chunk_metric_results}
\end{table*}

This result is consistent with results reported in \citet{thai-etal-2022-exploring, zhang-etal-2025-litransproqa, zhang-etal-2025-good}.

\subsection{Case study: How about translations into languages other than English?}
\label{subsec:case-study}

To examine whether our findings extend to other target languages, we conducted a small exploratory study with two books translated out of Japanese and Polish into Spanish, French, Polish, and Japanese, evaluated by five native-speaker readers following the main-study protocol.

\paragraph{Setup.} In order to explore whether the strong performance of MT holds beyond English, we run a small multilingual case study in which readers compared HT and MT into Spanish, French, Polish, and Japanese.\footnote{For Spanish, French, and Polish the readers were researchers and volunteers who are native speakers of the given language; we used two readers for the Polish book. For Japanese we followed the English procedure and hired one reader via Upwork, paid \$110 USD.} We translate \textit{My Grandfather, the Master Detective} (Japanese source) into Spanish, French, and Polish, and \textit{The Witcher: Crossroads of Ravens} (Polish source) into Japanese, choosing these two books because both have recently been professionally translated and published in the respective target languages. 

\begin{figure}[t]
    \centering
    \includegraphics[
        width=0.94\columnwidth,
        height=0.24\textheight,
        keepaspectratio
    ]{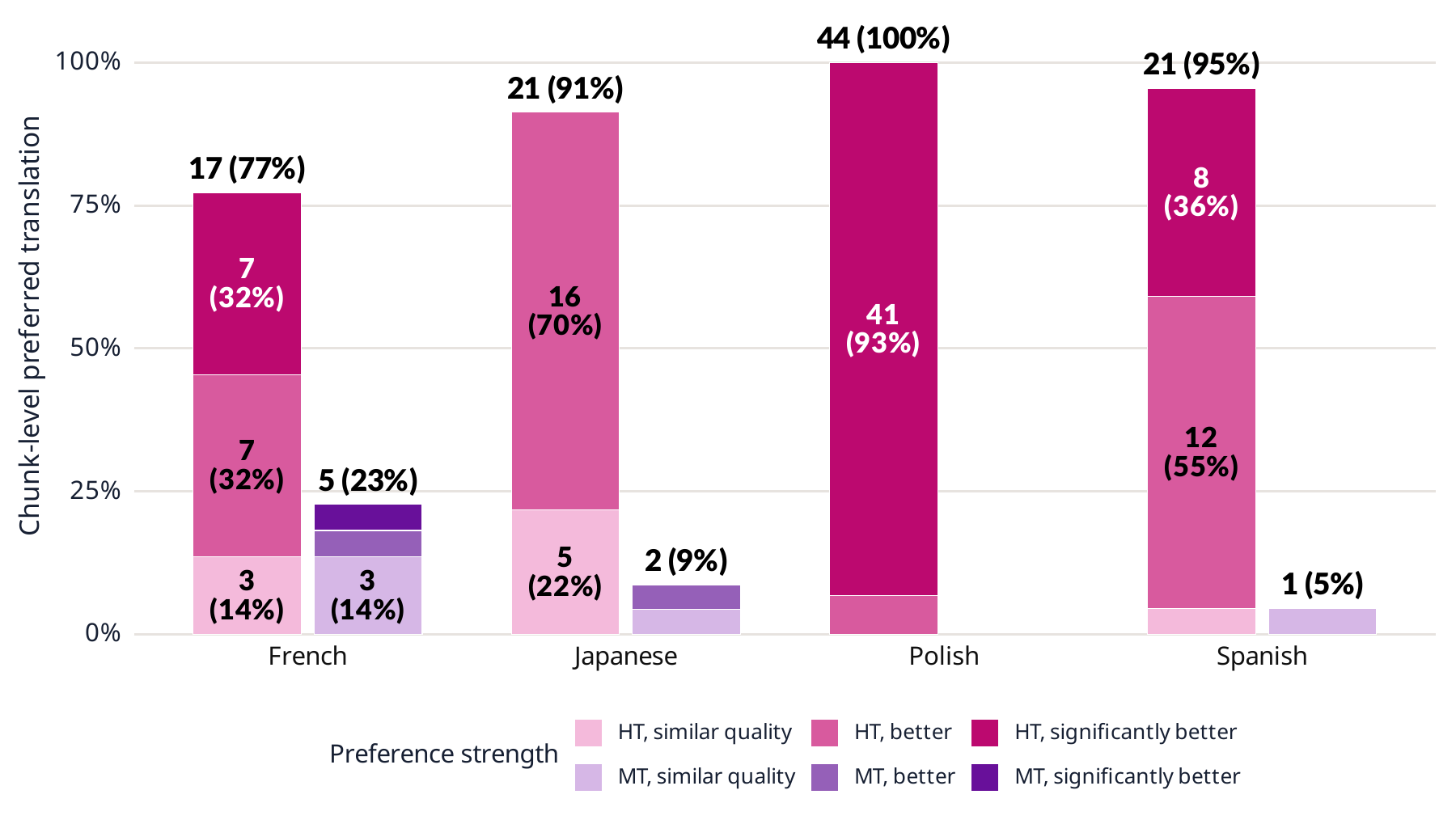}
    \caption{Close-reading preferences in the multilingual case study. Stacked bars show chunk-level preferred translation choices for HT and MT across the four evaluated translation directions, with color intensity indicating preference strength.}
    \label{fig:part2_preference_by_language}
\end{figure}

\paragraph{HT is strongly preferred for translations into languages other than English.}
The HT vs. MT contrast was much sharper than in translation into English. At the chunk level, every reader preferred HT overall, with HT favored in 103 of 111 chunk judgments (92.8\%). At the excerpt level, only the French reader chose MT; however, they still rated HT highly in single-reading, with just a one-point difference between MT and HT (5 vs. 4 for \texttt{acceptability}, \texttt{smoothness}, etc.). The remaining four readers preferred HT, with single-reading ratings favoring HT most for translation into Japanese and Polish. Even the French reader changed their preference toward HT in the close-reading task (\autoref{fig:book_preference_diverging}, \autoref{fig:part2_chunk_choice_heatmap}, and \autoref{fig:part2_preference_by_language}).

The gap was most visible for the Japanese and Polish targets with HT rated 1.75--2 points higher than MT on every single-reading scale; for the Spanish and French targets the rating gap was small or absent, with the French target showing the lowest HT chunk share (17/22).

\paragraph{Readers identified MT through concrete translationese.}
Unlike in the English-facing study, where MT origin was difficult to detect, for translations into other languages the detection accuracy was better with 4 out of 5 readers identifying MT correctly and with strong average confidence (mean 4.4/5). Furthermore, the readers' explanations point to concrete issues such as translationese. One Polish reader states they were 100\% sure that the MT \textit{``cannot possibly be human-translated (...) [at] around 10\% of reading.''}  Readers flagged grammatical and structural problems, such as  sentences \textit{``hanging there as if directly translated from the source, with no predicate,''} as well as source language calques, including \textit{``the third person for Kaede when she speaks about herself.''}\footnote{It is common in Japanese to refer to yourself in the third person but in other languages it often seems either childish or weird.} One recurring cue was also vocabulary, especially in cases where MT failed to use a standard term and instead supplied description \textit{ ``like a student who has not mastered the vocabulary fully yet.''} For example, where the human translator in Polish wrote \textit{nekrolog} (``obituary''), the MT wrote \textit{prasowe artykuły pośmiertne} (``posthumous press articles''). Other times MT used a wrong or non-idiomatic term, for instance, in Spanish, \textit{librerías de viejo} where the human wrote \textit{librerías de segunda mano} (``second-hand bookshops''), a phrasing the reader judged \textit{``not idiomatic in any variety of Spanish.''}

These preliminary results suggest MT's strong English-target performance may not extend to other target languages (see \S\ref{app:multilingual-case-study-results}).

\section{Related Work}

\paragraph{Automatic evaluation.} Automatic evaluation is practical but surface-level \citep{papineni-etal-2002-bleu, popovic-2015-chrf} and captures only partial signals in literary MT \citep{griebel2026fluency}, while trained neural metrics \citep{rei2020comet, rei-etal-2022-cometkiwi, guerreiro-etal-2024-xcomet, juraska-etal-2023-metricx, juraska-etal-2024-metricx} were shown to work poorly on literary machine translation \citep{thai-etal-2022-exploring, zhang-etal-2025-litransproqa} and offer weak signal for what humans perceive as literary quality \citep{gerrits2026creativity}. 
LLM-as-a-judge approaches based on question answering protocols or creative tasks were developed but tested only on short passages \citep{zhang-etal-2025-litransproqa,shafayat20242, zhang2026beyond} and it is unclear how they handle longer dependencies.
Recent document-level and long-document MT work often combines automatic metrics with LLM-as-a-judge evaluation \citep{laubli2018has, laubli2020set, wang2026loong}, but such evaluations still do not directly measure reader experience.

\paragraph{Human evaluation.} Human evaluation often relies on error annotations \citep{burchardt-2013-multidimensional,freitag-etal-2021-experts, kocmi-etal-2024-error} or crowdsourced ratings \citep{graham-etal-2013-continuous, freitag-etal-2021-experts}, which do not capture reading experience \citep{zhang-etal-2025-good}. Most evaluation of literary MT uses short passages or local judgments \citep{wu-etal-2025-perhaps,karpinska-iyyer-2023-large,wang-etal-2023-findings, kocmi-etal-2025-findings}, missing the purpose of literary translation as sustained reading. Prior work on literary MT and reception remains narrower than our reader-facing comparison of professional HT and recent MT over long fiction excerpts \citep{guerberof2020impact,guerberof2022creativity,guerberof2024or,gerrits2025mt,moorkens2018translators,castaldo-etal-2025-extending,matusov2019challenges,taivalkoski2019free,webster2020gutenberg,kosters2023riddle}.
This reader-facing framing also relates to narrative transportation, AI-perception, literary translator voice, and ethics in machine-assisted literary translation \citep{green2000role,porter2024ai,zhu2025human,kenny2020machine,taivalkoski2019ethical}. In contrast, we target readers' experience of long literary MT versus HT in \textit{immersive reading}, and what issues they notice in \textit{close reading}.

\section{Discussion \& Conclusion}

We evaluated literary MT in \textit{immersive} and \textit{close} reading by comparing professional human translations (HT) with a strong agentic LLM pipeline (MT) across 15 recent French, Polish, and Japanese novels translated into English.
Our analysis reveals that readers found MT reasonable and were not able to detect it reliably, but HT was favored descriptively at the excerpt level and significantly in the close reading.
Preferences varied mostly by book, not source language, and readers often preferred the version they believed to be human, while automatic metrics favored MT. We release \name{} and our evaluation pipeline to support future research on literary MT.

\section*{Limitations}

\paragraph{Excerpt scope.} Due to copyrights restrictions and annotation costs we evaluate translations of novel openings, not full books. Literary quality, pacing, and narrative voice may evolve over longer reading, and book-spanning context and background may be lost by LLM translations of full books. These factors may impact judgments over longer translations, however we note that 8K words is already at the length of short stories.

\paragraph{Corpus selection.}
We selected recently published (2025--2026), critically acclaimed fiction with professional English translations available to mitigate training-data contamination. However, this limits the diversity of our books and does not represent the whole set of literature out there.

\paragraph{Languages and directions.}
The main evaluation is French, Polish, and Japanese into English only ($n{=}5$ books per source language). Our multilingual case study covers two additional books and four target languages with a much smaller reader sample and mixed recruitment (Upwork vs.\ researchers/volunteers); those results are exploratory and should not be generalized. A much larger-scale study is required to determine whether the effects observed in the current work apply to other source-target language pairs.

\paragraph{Sample size and agreement.}
Each book is read by only two readers recruited via Upwork with a total of 15 readers. This scale was necessary for immersive, long-form evaluation to be financially feasible, but could limit statical power. To partially mitigate this, we employ a within-subject design, where each reader judges both MT and HT for the same excerpts, reducing individual variability and requiring fewer annotators than a between-subjects design for statistical analysis \citep{Allen2017-withinsubject}.

\section*{Ethics Statement}

\paragraph{Human participants.}
This study was approved by the Simon Fraser University Research Ethics Board (\#30003790). Participants gave informed consent before the evaluation (\S\ref{sec:consent-form}), could withdraw with prorated compensation, and were paid fairly for the amount and difficulty of work (\$110 USD per book, $\sim$\$27.50/hour for $\sim$4 hours of work), plus completion bonuses. Participant identities are anonymized by participant codes. 

\paragraph{Copyright and data sharing.}
Source books were purchased for research use. Following fair-dealing / fair-use principles for scholarly research, we retain no more than 10\% of any individual work. The dataset will be released \textbf{for research purposes only}.

\paragraph{Corpus and societal impact.}
Our corpus over-represents internationally published, commercially successful fiction. 
We study AI-assisted literary translation to inform evaluation practice and publishing transparency: \textbf{not to advocate replacing human translators or deceiving readers}; in fact, we demonstrate that readers often prefer human-authored translations. As LLM-translated fiction enters trade and self-publishing channels, \textbf{we hope that better reader-centered evaluation and full disclosure can support fair labor practices and help readers make informed choices.}

\paragraph{Risks of literary machine translation.}
Literary MT can look acceptable while still changing the reading experience. As our reader comments show, problems with phrasing, context, and voice can affect the reader's impression of a book and its author, especially when AI-translated fiction reaches readers with little human review. Like other LLM output, MT may also carry biased or harmful wording into the translation.

\paragraph{AI use disclosure.} We utilize coding agents to produce plots and edit table layout. We further use language models as writing assistants but no part of this paper was generated from scratch.

\section*{Acknowledgements}
We would like to thank the Upwork participants for their help in evaluating the different book excerpts, and for taking the time to answer our questions with detailed and thoughtful responses that were valuable to our study.
We are grateful to the volunteer and researcher readers who contributed to the multilingual case study.
This work was supported by the following grants and awards:
Simon Fraser University's Rajan Recruitment Award for Women in Computing Science [M Karpinska], Natural Sciences and Engineering Research Council of Canada (NSERC) [MJ Meurs, Discovery Grant RGPIN-2025-07163], [M Taboada, Discovery Grant RGPIN-2022-04481], [M Karpinska, Discovery Grant RGPIN 2026-04987], and the Fonds de recherche du Québec (FRQ) Chaire de recherche du Québec sur la découvrabilité des contenus scientifiques en français [MJ Meurs, Chaire de recherche du Québec sur la découvrabilité des contenus scientifiques en français,  DOI 10.69777/358425].

\bibliography{EMNLP2026/biblio}

\newpage
\appendix
\section{Dataset}
\label{sec:app_mt_litdata}
\newcolumntype{P}[1]{>{\raggedright\arraybackslash}p{#1}}
\newcolumntype{Y}{>{\raggedright\arraybackslash}X}

\begin{table*}[t]
    \centering
    \scriptsize
    \setlength{\tabcolsep}{3pt}
    \renewcommand{\arraystretch}{1.08}
    
    \begin{tabular}{@{}l P{2.65cm} P{2.05cm} P{3.25cm} P{2.45cm} r r@{}}
        \toprule
        Pair & Original Title & Author & Translation Title & Translator & Original Date & Translation Date \\
        \midrule
        \multicolumn{7}{@{}l}{\textbf{Main human evaluation: source to English}}\\
        \addlinespace[1pt]
        
        FR$\rightarrow$EN & \textit{Le Roman de Marceau Miller} & Marceau Miller & \textit{The Story of Marceau Miller} & Howard Curtis & Jan. 2025 & Mar. 2026 \\
        FR$\rightarrow$EN & \textit{La Symphonie des monstres} & Marc Levy & \textit{Symphony of Monsters} & Tina Kover & Oct. 2023 & Jan. 2026 \\
        FR$\rightarrow$EN & \textit{Veiller sur elle} & Jean-Baptiste Andrea & \textit{Watching Over Her} & Frank Wynne & Aug. 2023 & Aug. 2025 \\
        FR$\rightarrow$EN & \textit{Les Yeux de Mona} & Thomas Schlesser & \textit{Mona's Eyes} & Hildegarde Serle & Jan. 2024 & Sept. 2025 \\
        FR$\rightarrow$EN & \textit{C\'el\`ebre} & Maud Ventura & \textit{Make Me Famous} & Gretchen Schmid & Aug. 2024 & May 2025 \\
        
        \addlinespace[2pt]
        
        JA$\rightarrow$EN & \textit{Hakobune} & Haruo Yuki & \textit{The Ark} & Jim Rion & Sept. 2022 & Feb. 2026 \\
        JA$\rightarrow$EN & \textit{Meitantei no Mama de Ite} & Masateru Konishi & \textit{My Grandfather, the Master Detective} & Louise Heal Kawai & Jan. 2023 & Mar. 2026 \\
        JA$\rightarrow$EN & \textit{Kokun, Vol. 1: Nishi kara Kita Shojo} & Nahoko Uehashi & \textit{Kokun: The Girl from the West} & Cathy Hirano & Mar. 2022 & Feb. 2026 \\
        JA$\rightarrow$EN & \textit{Nairu Pachi no Joshikai} & Asako Yuzuki & \textit{Hooked: A Novel of Obsession} & Polly Barton & Mar. 2015 & Mar. 2026 \\
        JA$\rightarrow$EN & \textit{Kiiroi Ie} & Mieko Kawakami & \textit{Sisters in Yellow} & Laurel Taylor and Hitomi Yoshio & Feb. 2023 & Mar. 2026 \\
        
        \addlinespace[2pt]
        
        PL$\rightarrow$EN & \textit{Rozdro\.ze kruk\'ow} & Andrzej Sapkowski & \textit{The Witcher: Crossroads of Ravens} & David French & Oct. 2024 & Sept. 2025 \\
        PL$\rightarrow$EN & \textit{Stacja} & Jakub Szama\l{}ek & \textit{Inner Space} & Kasia Beresford & May 2023 & July 2025 \\
        PL$\rightarrow$EN & \textit{Stramer} & Miko\l{}aj \L{}ozi\'nski & \textit{My Name is Stramer} & Antonia Lloyd-Jones & Oct. 2019 & Aug. 2025 \\
        PL$\rightarrow$EN & \textit{Ucho Igielne} & Wies\l{}aw My\'sliwski & \textit{Needle's Eye} & Bill Johnston & Oct. 2018 & Oct. 2025 \\
        PL$\rightarrow$EN & \textit{Empuzjon. Horror przyrodoleczniczy} & Olga Tokarczuk & \textit{The Empusium: A Health Resort Horror Story} & Antonia Lloyd-Jones & June 2022 & Sept. 2024 \\
        
        \midrule
        \addlinespace[2pt]
        \multicolumn{7}{@{}l}{\textbf{Multilingual target-language case study}}\\
        \addlinespace[1pt]
        
        JA$\rightarrow$FR & \textit{Meitantei no Mama de Ite} & Masateru Konishi & \textit{Les Histoires de Kaede} & Mathilde Tamae-Bouhon & Jan. 2023 & May 2024 \\
        JA$\rightarrow$PL & \textit{Meitantei no Mama de Ite} & Masateru Konishi & \textit{M\'oj dziadek -- genialny detektyw} & Dariusz Lato\'s & Jan. 2023 & Sept. 2025 \\
        JA$\rightarrow$ES & \textit{Meitantei no Mama de Ite} & Masateru Konishi & \textit{La peque\~na habitaci\'on de los misterios} & Juan Francisco Gonz\'alez S\'anchez & Jan. 2023 & Sept. 2025 \\
        PL$\rightarrow$JA & \textit{Rozdro\.ze kruk\'ow} & Andrzej Sapkowski & \textit{Witcher: Karasu no Jujiro} & Yasuko Kawano and Aya Sugiura & Oct. 2024 & Sept. 2025 \\
        
        \midrule
        \addlinespace[2pt]
        \multicolumn{7}{@{}l}{\textbf{Development books for MT pipeline selection and ablations}}\\
        \addlinespace[1pt]
        
        FR$\rightarrow$EN & \textit{Les guerres précieuses} & Perrine Tripier & \textit{Our Precious Wars} & Alison Anderson & Jan. 2023 & Nov. 2025 \\
        FR$\rightarrow$EN & \textit{Naufrage} & Vincent Delecroix & \textit{Small Boat} & Helen Stevenson & Aug. 2023 & Mar. 2025 \\
        FR$\rightarrow$EN & \textit{Déserter} & Mathias Énard & \textit{The Deserters} & Charlotte Mandell & Aug. 2023 & May 2025 \\
        FR$\rightarrow$EN & \textit{Ilaria ou la conquête de la désobéissance} & Gabriella Zalapi & \textit{Ilaria, or The Conquest of Disobedience} & Adriana Hunter & Aug. 2024 & Nov. 2025 \\
        FR$\rightarrow$EN & \textit{Le vieil incendie} & Elisa Shua Dusapin & \textit{The Old Fire} & Aneesa Abbas Higgins & Aug. 2023 & Jan. 2026 \\
        
        \addlinespace[2pt]
        
        JA$\rightarrow$EN & \textit{Jakuson Hitori} & Jose Ando & \textit{Jackson Alone} & Kalau Almony & Nov. 2022 & Jan. 2026 \\
        JA$\rightarrow$EN & \textit{Machi to Sono Futashika na Kabe} & Haruki Murakami & \textit{The City and Its Uncertain Walls} & Philip Gabriel & Apr. 2023 & Nov. 2024 \\
        JA$\rightarrow$EN & \textit{Baba Yaga no Yoru} & Akira Otani & \textit{The Night of Baba Yaga} & Sam Bett & Oct. 2020 & July 2024 \\
        JA$\rightarrow$EN & \textit{Kai ni Tsuzuku Bashonite} & Mai Ishizawa & \textit{The Place of Shells} & Polly Barton & July 2021 & Mar. 2025 \\
        JA$\rightarrow$EN & \textit{Kyūkanbi no kanojotachi} & Emi Yagi & \textit{When the Museum Is Closed} & Yuki Tejima & Mar. 2023 & Jan. 2026 \\
        
        \addlinespace[2pt]
        
        PL$\rightarrow$EN & \textit{Powrót z gwiazd} & Stanisław Lem & \textit{Return from the Stars} & Barbara Marszal & Jan. 1961 & June 1989 \\
        PL$\rightarrow$EN & \textit{Księgi Jakubowe} & Olga Tokarczuk & \textit{The Books of Jacob} & Jennifer Croft & Oct. 2014 & Nov. 2021 \\
        PL$\rightarrow$EN & \textit{Lód} & Jacek Dukaj & \textit{Ice} & Ursula Phillips & Jan. 2007 & Nov. 2025 \\
        PL$\rightarrow$EN & \textit{Mgła} & Kaja Malanowska & \textit{Fog} & Bill Johnston & Jan. 2015 & Mar. 2021 \\
        PL$\rightarrow$EN & \textit{Łakome} & Małgorzata Lebda & \textit{Voracious} & Antonia Lloyd-Jones & Aug. 2023 & Jan. 2025 \\
        PL$\rightarrow$EN & \textit{Poufne} & Mikołaj Grynberg & \textit{Confidential} & Sean Gasper Bye & June 2020 & Jan. 2025 \\
        
        \bottomrule
    \end{tabular}
    \caption{Books used in the human evaluation, multilingual target-language case study, and MT-pipeline development experiments. The table lists the source-target language pair, original title and author, translated publication title and translator, and publication dates for the original and translated editions.}
    \label{tab:translated-works}
\end{table*}

\begin{table*}[t!]
    \centering
    \footnotesize
    \setlength{\tabcolsep}{5pt}
    \resizebox{\textwidth}{!}{%
        \begin{tabular}{lll cc c ccccc}
            \toprule
            & & & \multicolumn{1}{c}{\textbf{Excerpts} \includegraphics[height=1.1em]{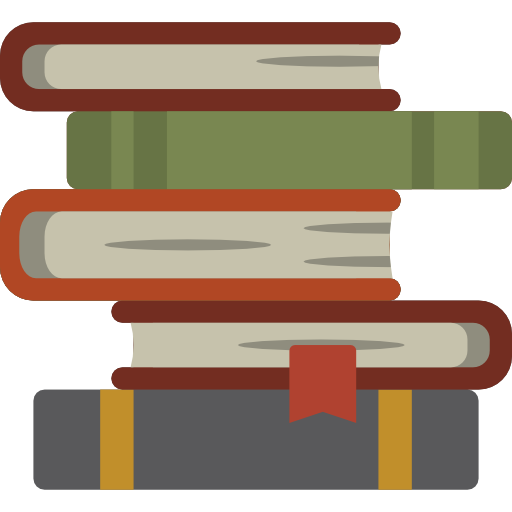}} & \multicolumn{1}{c}{\textbf{Chunks} \chunkicon} &  & \multicolumn{5}{c}{\textbf{Participant Comments} \includegraphics[height=1em]{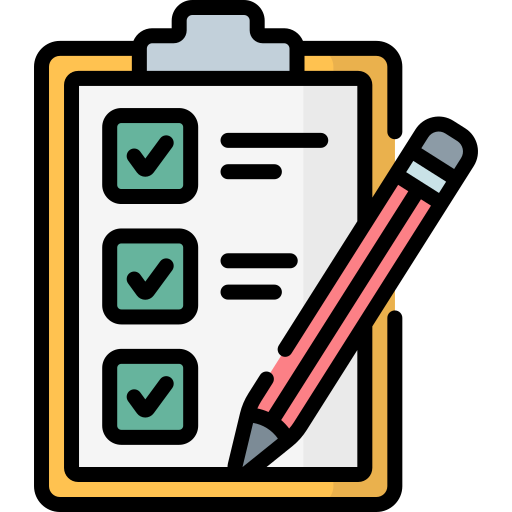}} \\
            & & & \multicolumn{1}{c}{(\textit{n=15})} & \multicolumn{1}{c}{(\textit{n=386})} & & \multicolumn{1}{c}{(\textit{n=60})} & \multicolumn{1}{c}{(\textit{n=60})} & \multicolumn{1}{c}{(\textit{n=30})} & \multicolumn{1}{c}{(\textit{n=30})} & \multicolumn{1}{c}{(\textit{n=772})} \\
            \cmidrule(lr){4-4}\cmidrule(lr){5-5}\cmidrule(lr){6-6}\cmidrule(lr){7-11}
            & & &     &     & \textsc{\#/Ex} & \textsc{Single Q5} & \textsc{Single Q6} & \textsc{Compar. Q4} & \textsc{Compar. Q7} & \textsc{Chunk Justif.} \\
            \midrule
            \multirow{18}{*}{\rotatebox{90}{\textbf{\textsc{Words}}}} & \multirow{3}{*}{\textsc{Mean}} & \htr ~ \includegraphics[height=1em]{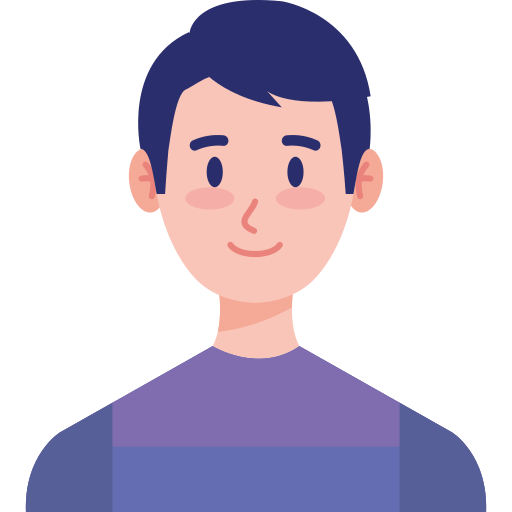} & 7,623.3 & 296.2 & \multirow{3}{*}{25.7} & \multirow{3}{*}{35.6} & \multirow{3}{*}{49.2} & \multirow{3}{*}{76.9} & \multirow{3}{*}{69.2} & \multirow{3}{*}{47.9} \\
             &  & \mtr ~ \includegraphics[height=1em]{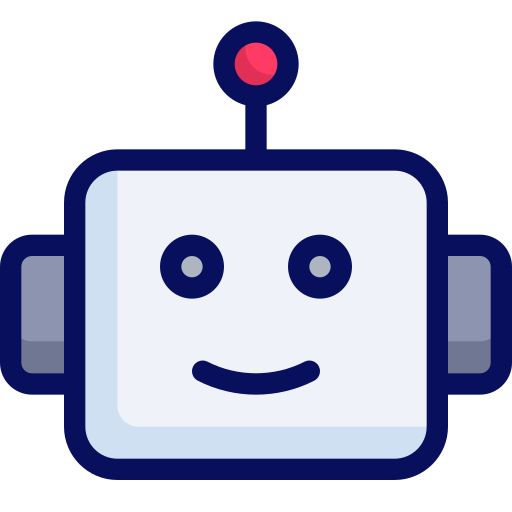} & 7,676.5 & 298.3 &  &  &  &  &  &  \\
             &  & \srcr ~ \includegraphics[height=1em]{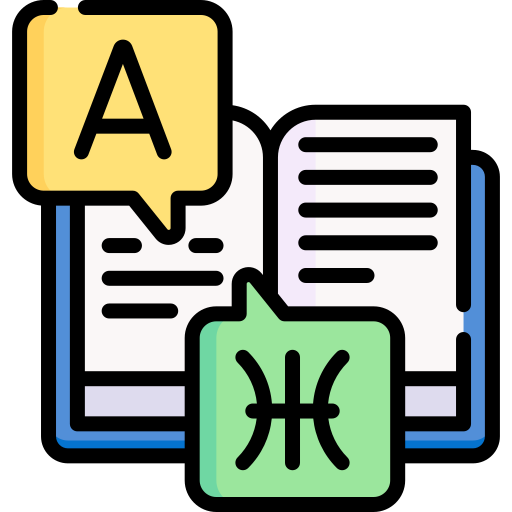} & 4,784.1 & 186.4 &  &  &  &  &  &  \\
            \addlinespace
             & \multirow{3}{*}{\textsc{Median}} & \htr ~ \includegraphics[height=1em]{figs/Icons/human.png} & 7,877 & 307 & \multirow{3}{*}{26} & \multirow{3}{*}{29} & \multirow{3}{*}{41.5} & \multirow{3}{*}{72.5} & \multirow{3}{*}{57} & \multirow{3}{*}{38} \\
             &  & \mtr ~ \includegraphics[height=1em]{figs/Icons/robot.png} & 7,623 & 308 &  &  &  &  &  &  \\
             &  & \srcr ~ \includegraphics[height=1em]{figs/Icons/foreign_book.png} & 6,273 & 240 &  &  &  &  &  &  \\
            \addlinespace
             & \multirow{3}{*}{\textsc{St. Dev.}} & \htr ~ \includegraphics[height=1em]{figs/Icons/human.png} & 449.3 & 49.2 & \multirow{3}{*}{2.2} & \multirow{3}{*}{23.3} & \multirow{3}{*}{32.8} & \multirow{3}{*}{35.9} & \multirow{3}{*}{65.1} & \multirow{3}{*}{29.6} \\
             &  & \mtr ~ \includegraphics[height=1em]{figs/Icons/robot.png} & 529.2 & 57.4 &  &  &  &  &  &  \\
             &  & \srcr ~ \includegraphics[height=1em]{figs/Icons/foreign_book.png} & 3,334.1 & 128.4 &  &  &  &  &  &  \\
            \addlinespace
             & \multirow{3}{*}{\textsc{Max}} & \htr ~ \includegraphics[height=1em]{figs/Icons/human.png} & 7,995 & 486 & \multirow{3}{*}{29} & \multirow{3}{*}{93} & \multirow{3}{*}{176} & \multirow{3}{*}{184} & \multirow{3}{*}{380} & \multirow{3}{*}{215} \\
             &  & \mtr ~ \includegraphics[height=1em]{figs/Icons/robot.png} & 8,595 & 530 &  &  &  &  &  &  \\
             &  & \srcr ~ \includegraphics[height=1em]{figs/Icons/foreign_book.png} & 8,244 & 417 &  &  &  &  &  &  \\
            \addlinespace
             & \multirow{3}{*}{\textsc{Min}} & \htr ~ \includegraphics[height=1em]{figs/Icons/human.png} & 6,747 & 100 & \multirow{3}{*}{22} & \multirow{3}{*}{9} & \multirow{3}{*}{7} & \multirow{3}{*}{24} & \multirow{3}{*}{21} & \multirow{3}{*}{16} \\
             &  & \mtr ~ \includegraphics[height=1em]{figs/Icons/robot.png} & 6,928 & 102 &  &  &  &  &  &  \\
             &  & \srcr ~ \includegraphics[height=1em]{figs/Icons/foreign_book.png} & 141 & 1 &  &  &  &  &  &  \\
            \addlinespace
             & \multirow{3}{*}{\textsc{Total}} & \htr ~ \includegraphics[height=1em]{figs/Icons/human.png} & 114,350 & 114,350 & \multirow{3}{*}{386} & \multirow{3}{*}{2,136} & \multirow{3}{*}{2,955} & \multirow{3}{*}{2,306} & \multirow{3}{*}{2,077} & \multirow{3}{*}{36,964} \\
             &  & \mtr ~ \includegraphics[height=1em]{figs/Icons/robot.png} & 115,147 & 115,147 &  &  &  &  &  &  \\
             &  & \srcr ~ \includegraphics[height=1em]{figs/Icons/foreign_book.png} & 71,762 & 71,762 &  &  &  &  &  &  \\
            \midrule
            \multirow{18}{*}{\rotatebox{90}{\textbf{\textsc{Tokens}}}} & \multirow{3}{*}{\textsc{Mean}} & \htr ~ \includegraphics[height=1em]{figs/Icons/human.png} & 9,680.3 & 376.2 & \multirow{3}{*}{25.7} & \multirow{3}{*}{41.3} & \multirow{3}{*}{60.5} & \multirow{3}{*}{95.2} & \multirow{3}{*}{86.2} & \multirow{3}{*}{61.3} \\
             &  & \mtr ~ \includegraphics[height=1em]{figs/Icons/robot.png} & 9,582.1 & 372.4 &  &  &  &  &  &  \\
             &  & \srcr ~ \includegraphics[height=1em]{figs/Icons/foreign_book.png} & 13,026.4 & 507.5 &  &  &  &  &  &  \\
            \addlinespace
             & \multirow{3}{*}{\textsc{Median}} & \htr ~ \includegraphics[height=1em]{figs/Icons/human.png} & 9,832 & 390 & \multirow{3}{*}{26} & \multirow{3}{*}{34} & \multirow{3}{*}{52} & \multirow{3}{*}{86.5} & \multirow{3}{*}{68.5} & \multirow{3}{*}{48} \\
             &  & \mtr ~ \includegraphics[height=1em]{figs/Icons/robot.png} & 9,648 & 384 &  &  &  &  &  &  \\
             &  & \srcr ~ \includegraphics[height=1em]{figs/Icons/foreign_book.png} & 13,023 & 514 &  &  &  &  &  &  \\
            \addlinespace
             & \multirow{3}{*}{\textsc{St. Dev.}} & \htr ~ \includegraphics[height=1em]{figs/Icons/human.png} & 535.0 & 65.7 & \multirow{3}{*}{2.2} & \multirow{3}{*}{27.8} & \multirow{3}{*}{43.9} & \multirow{3}{*}{46.3} & \multirow{3}{*}{82.5} & \multirow{3}{*}{38.6} \\
             &  & \mtr ~ \includegraphics[height=1em]{figs/Icons/robot.png} & 625.7 & 73.5 &  &  &  &  &  &  \\
             &  & \srcr ~ \includegraphics[height=1em]{figs/Icons/foreign_book.png} & 1,527.0 & 121.7 &  &  &  &  &  &  \\
            \addlinespace
             & \multirow{3}{*}{\textsc{Max}} & \htr ~ \includegraphics[height=1em]{figs/Icons/human.png} & 10,261 & 611 & \multirow{3}{*}{29} & \multirow{3}{*}{113} & \multirow{3}{*}{254} & \multirow{3}{*}{231} & \multirow{3}{*}{482} & \multirow{3}{*}{272} \\
             &  & \mtr ~ \includegraphics[height=1em]{figs/Icons/robot.png} & 10,500 & 644 &  &  &  &  &  &  \\
             &  & \srcr ~ \includegraphics[height=1em]{figs/Icons/foreign_book.png} & 15,967 & 1,139 &  &  &  &  &  &  \\
            \addlinespace
             & \multirow{3}{*}{\textsc{Min}} & \htr ~ \includegraphics[height=1em]{figs/Icons/human.png} & 8,600 & 118 & \multirow{3}{*}{22} & \multirow{3}{*}{11} & \multirow{3}{*}{8} & \multirow{3}{*}{33} & \multirow{3}{*}{23} & \multirow{3}{*}{18} \\
             &  & \mtr ~ \includegraphics[height=1em]{figs/Icons/robot.png} & 8,474 & 121 &  &  &  &  &  &  \\
             &  & \srcr ~ \includegraphics[height=1em]{figs/Icons/foreign_book.png} & 11,155 & 141 &  &  &  &  &  &  \\
            \addlinespace
             & \multirow{3}{*}{\textsc{Total}} & \htr ~ \includegraphics[height=1em]{figs/Icons/human.png} & 145,204 & 145,204 & \multirow{3}{*}{386} & \multirow{3}{*}{2,480} & \multirow{3}{*}{3,633} & \multirow{3}{*}{2,855} & \multirow{3}{*}{2,586} & \multirow{3}{*}{47,326} \\
             &  & \mtr ~ \includegraphics[height=1em]{figs/Icons/robot.png} & 143,732 & 143,732 &  &  &  &  &  &  \\
             &  & \srcr ~ \includegraphics[height=1em]{figs/Icons/foreign_book.png} & 195,396 & 195,396 &  &  &  &  &  &  \\
            \bottomrule
        \end{tabular}
    }
    \caption{Summary statistics for evaluation excerpts, chunks, and participant free-text comments, split into a \textbf{\textsc{Words}} sub-table (whitespace-delimited word counts; we do not report \srcr{} word counts because Japanese cannot be split on whitespaces) and a \textbf{\textsc{Tokens}} sub-table (\texttt{tiktoken} \texttt{o200k\_base} token counts). Excerpts and chunks are reported separately for human-translated (\htr) and machine-translated (\mtr) versions in the \textbf{\textsc{Words}} sub-table; the \textbf{\textsc{Tokens}} sub-table also includes source-language (\srcr) versions. The \textsc{\#/Ex} (chunks per excerpt) column does not depend on the metric and is identical across both sub-tables (and across \htr ~/~ \mtr ~/~ \srcr{}, since chunks are aligned across versions). \textsc{Participant Comments} include: \textit{Single Q5/Q6} are free-text responses from the single-reading questionnaires; \textit{Compar.\ Q4/Q7} are free-text responses from the comparison questionnaire; \textit{Chunk Justif.}\ are per-chunk preference justifications.}
    \label{tab:basic_summary_stats_dataset}
\end{table*}

In this section of the appendix, we provide additional information regarding our dataset. 

\paragraph{Corpus construction.}
A total of 35 long excerpts have been used for this study.
15 for the evaluation, 16 for the development and 4 for the case-study.
Details about the selected books, such as the title, authors, but also translators and release dates are presented in \autoref{tab:translated-works}.

\paragraph{Dataset statistics.}
Additional details about the corpus size for the excerpt and chunks are presented in \autoref{tab:basic_summary_stats_dataset}.
It also includes statistical information about the participants’ comments written during the two reading phases.

\section{Machine Translation}
\label{app:app_mt_pipe}

In this section of the appendix, we provide more details on our machine translation pipeline, including the selection of a reasonable translation approach.

\paragraph{Machine translation pipelines.} Overall, we experiment with three pipelines in five setups:
\begin{enumerate}[label=P\arabic*, nosep]
    \item Full-excerpt translation with post-editing by (a) GPT-5.4 or (b) Gemini 3.1 Pro.
    \item Chunk-level translation with excerpt-level post-editing by (a) GPT-5.4 or (b) Gemini 3.1 Pro.
    \item An agentic pipeline with Claude Code (Opus 4.6) and Codex (GPT-5.4), inspired by AutoFiction~\cite{pham_chang_desai_iyyer_2026}, an agent-based novel-generation pipeline.
\end{enumerate}

We select GPT-5.4 and Gemini 3.1 Pro because these model families were strong performers in the \textsc{WMT 2025} General Translation Shared Task~\citep{kocmi-etal-2025-findings}. We further explore coding agents because they are optimized for long-context tasks and for handling multiple reference documents.

\paragraph{Prompts.} We deploy different prompts for each pipeline. As 
the pipelines include multiple long prompts we refrain from including the entire prompt texts in the appendix and make them accessible through the repository instead (\href{https://github.com/Yves575/lait}{\path{github.com/Yves575/lait}}).

\paragraph{Details on prompting pipelines.}
P1 and P2 used the same two-stage prompting procedure. 
In the first stage, the model was prompted as an expert literary translator and asked to translate the source excerpt into English while preserving meaning, tone, voice, register, rhythm, stylistic effect, ambiguity, characterization, and differences between narrator and character voices. 
The prompt also instructed the model to avoid unjustified omissions or additions and to output only the translation. 
In the second stage, the model received the original source excerpt together with its own first-pass English translation and was tasked with revision. 

\paragraph{[P1] Full-excerpt translation.}
For P1, each complete source excerpt was translated in a single first-pass call. 
This condition therefore gave the model the entire opening excerpt at once, including all available discourse context, recurring names, register shifts, and dialogue patterns. 
The resulting complete English draft was then passed to the post-editing stage together with the complete source excerpt. 
The post-editing pass was also performed at the excerpt level, so both stages in P1 had access to the full source span.

\paragraph{[P2] Chunk-level translation with excerpt-level post-editing.}
For P2, the first-pass translation was performed over smaller source chunks. 
We counted tokens with the \texttt{tiktoken} \texttt{o200k\_base} encoding and targeted a maximum of 1K source tokens per chunk. 
Chunking preserved paragraph boundaries: we split the source excerpt on blank-line paragraph breaks, added whole paragraphs to the current chunk until adding the next paragraph would exceed the token limit, and then started a new chunk. 
If a single paragraph exceeded the limit, it was kept intact as an oversized chunk rather than split mid-paragraph. 
Each chunk was translated independently with the same literary-translation prompt used in P1. 
The chunk translations were then concatenated in source order with paragraph breaks preserved, producing a complete first-pass English draft. 
Finally, as in P1, the post-editing stage received the complete source excerpt and the complete concatenated English draft, allowing the model to revise across chunk boundaries for consistency of voice, terminology, register, narrative perspective, and local fluency.

\paragraph{[P3] Agentic pipeline. } The entire pipeline is presented in \autoref{fig:agentic-mt-pipeline}. We report the agent responsible for each task and the associated checkpoint in \autoref{tab:agentic-pipeline-model-configuration}. The core parts of the pipelines are: (1) data preparation including style analysis, (2) chunk-level translation, and (3) excerpt-level review. For (2) and (3) we implement metrics and guard gates that send translations for edits if any issues are spotted. Below we explain the rubrics and acceptance gate mechanism. 

\begin{table}[t]
\centering
\footnotesize
\resizebox{\columnwidth}{!}{%
\begin{tabular}{@{}llll@{}}
\toprule
Pipeline step & Agent backend & Checkpoint & Runs \\
\midrule
Style Analysis & Claude & claude-opus-4-6 & 35 \\
Initial Chunk Translation & GPT & gpt-5.4 & 31 \\
Translation Quality Review & GPT & gpt-5.4 & 35 \\
Literary Quality Review & Claude & claude-opus-4-6 & 35 \\
Targeted Chunk Revision & GPT & gpt-5.4 & 31 \\
Targeted Chunk Revision \textit{(after 3 failed revisions)} & Claude & claude-opus-4-6 & 4 \\
Whole Excerpt Review & GPT & gpt-5.4 & 35 \\
Cross-Chunk Continuity Check & GPT & gpt-5.4 & 35 \\
Global Revision & Claude & claude-opus-4-6 & 35 \\
\bottomrule
\end{tabular}
}
\caption{Model configuration for the agentic machine-translation pipeline. The table shows which agent backend and checkpoint were used for each translation, review, and revision step, along with the number of runs using that configuration.}
\label{tab:agentic-pipeline-model-configuration}
\end{table}

\paragraph{[P3] Agent review rubrics and acceptance gates.}
The chunk-level acceptance gate combined two automatic reviews. 
First, the modified {\small LiTransProQA} review asked 25 yes/no/maybe questions about translation quality, such as faithfulness, cultural transfer, voice and more.
A chunk passed this quality-estimation component only if it received no \textsc{No} judgments and at most five \textsc{Maybe} judgments. 

Second, a literary review checked whether the chunk preserved the source's accuracy, voice, dialogue, and prose, and recorded findings with \textsc{Low}, \textsc{Medium}, \textsc{High}, or \textsc{Critical} severity.
A chunk passed the literary component only if every lens passed and there were no \textsc{Medium+} findings.\footnote{Following \citet{pham_chang_desai_iyyer_2026} we denote \textsc{Medium+} as any issue of severity medium or higher that would usually require revision. This excludes small issues marked as \textsc{Low}.} 
The local acceptance gate failed any chunk that failed either component and sent only those chunks to the next revision cycle.

The excerpt-level final gate combined a full-draft book review with a cross-chunk audit. 
The book review targeted global translation defects such as repeated translationese, flattening, loss of ambiguity, etc., visible only when reading the excerpt as a whole.
The cross-chunk audit targeted boundary and consistency problems, such as inconsistent terms, narrator or register drift, seam issues, etc.
Both reviews used the same four-level severity scale, and the final gate passed only when neither review found any \textsc{Medium+} issue; otherwise, affected chunks were sent to final revision.

\paragraph{Pipeline selection.} We translate 8K-words\footnote{The book length limit should not exceed 10\% of the original word count, with a maximum limit of 8k-words, and it can be lower.} excerpts from 16 development books under all five setups (\autoref{tab:pipeline-book-count-stats}).\footnote{Because agents make many model calls, running them via API is prohibitively expensive; instead we used subscription services, which cap our translation count.} In a blind five-way preference task, six raters (the authors and volunteer colleagues), each assessing 2--3 books, together ranked the outputs across all 16 development excerpts. The raters were asked to read at least 2K words to assure everyone reads past the chunk boundary. The raters can choose to read more. Besides choosing the best performing translation in their set of five candidates, each rater shares their observations on other translations. Overall, most outputs were comparable, except GPT-5.4 (P1) which noticeably underperformed\footnote{\textsc{\small P3} (5/16), \textsc{\small P1} with Gemini 3.1 Pro (4/16), \textsc{\small P2} with Gemini 3.1 Pro (3/16), \textsc{\small P2} with GPT-5.4 (3/16), and \textsc{\small P1} with GPT-5.4 (1/16).}. We adopted the agentic pipeline (P3), which was preferred by a small margin owing to its more idiomatic vocabulary. See below for the prompts and pipeline details used for this process.

\begin{table}[t]
    \centering
    \small
    \resizebox{\columnwidth}{!}{%
    \begin{tabular}{cll r r r r r r}
        \toprule
        & Pipeline & Model & \textsc{Total} & \textsc{Mean} & \textsc{Median} & \textsc{St. Dev.} & \textsc{Min} & \textsc{Max} \\
        \midrule
        \multirow{5}{*}{\rotatebox{90}{\textbf{\textsc{Words}}}} & \multirow{2}{*}{Pipeline 1} & Gemini & 79,232 & 4,952.0 & 4,019 & 2,397.4 & 2,508 & 9,272 \\
         &  & GPT-5.4 & 78,920 & 4,932.5 & 3,970 & 2,383.9 & 2,498 & 9,218 \\
        \cmidrule(lr){2-9}
         & \multirow{2}{*}{Pipeline 2} & Gemini & 80,614 & 5,038.4 & 4,142 & 2,442.1 & 2,534 & 9,379 \\
         &  & GPT-5.4 & 79,872 & 4,992.0 & 4,058 & 2,448.5 & 2,518 & 9,378 \\
        \cmidrule(lr){2-9}
         & Pipeline 3 & Agents & 79,851 & 4,990.7 & 4,057 & 2,467.6 & 2,476 & 9,012 \\
        \midrule
        \multirow{5}{*}{\rotatebox{90}{\textbf{\textsc{Tokens}}}} & \multirow{2}{*}{Pipeline 1} & Gemini & 100,622 & 6,288.9 & 5,042 & 3,109.0 & 3,150 & 11,974 \\
         &  & GPT-5.4 & 99,219 & 6,201.2 & 4,885 & 3,034.6 & 3,223 & 11,830 \\
        \cmidrule(lr){2-9}
         & \multirow{2}{*}{Pipeline 2} & Gemini & 102,331 & 6,395.7 & 5,185 & 3,180.4 & 3,204 & 12,171 \\
         &  & GPT-5.4 & 100,270 & 6,266.9 & 5,012 & 3,102.4 & 3,206 & 11,938 \\
        \cmidrule(lr){2-9}
         & Pipeline 3 & Agents & 99,640 & 6,227.5 & 5,012 & 3,124.3 & 3,145 & 11,670 \\
        \bottomrule
    \end{tabular}
    }
    \caption{Per-book word and token count statistics for each pipeline/model system, split into a \textbf{\textsc{Words}} sub-table (whitespace-delimited word counts) and a \textbf{\textsc{Tokens}} sub-table (\texttt{tiktoken} \texttt{o200k\_base} token counts). \textsc{Total} is the corpus sum across all 16 books; \textsc{Mean}, \textsc{Median}, \textsc{St.\ Dev.}, \textsc{Min}, and \textsc{Max} summarize the per-book distribution. Within each sub-table, rows are grouped by pipeline (Pipeline 1, Pipeline 2, Pipeline 3).}
    \label{tab:pipeline-book-count-stats}
\end{table}

\providecommand{\pass}{\textcolor{green!45!black}{\textbf{PASS}}}
\providecommand{\fail}{\textcolor{red!65!black}{\textbf{FAIL}}}
\setlength{\tabcolsep}{3pt}
\renewcommand{\arraystretch}{1.08}
\section{Agentic pipeline runs}
\label{app:agentic-pipe}

In this section of the appendix we provide more details on the agentic pipeline runs and artifacts it produced. All artifacts from agentic runs are shared at \href{https://github.com/Yves575/lait}{\path{github.com/Yves575/lait}}.

\paragraph{Local-cycle limits.} All 16 development-book runs used up to three local cycles. The third local cycle yielded limited marginal gains in the automatic gate outcomes, clearing 8 of 130 chunks still failing after cycle 2 (6.2\%); 12 of the 17 three-cycle runs had an unchanged failing-chunk set. We therefore capped subsequent main evaluation and multilingual target-language case-study runs at two local cycles. The exception is \textit{Make Me Famous}, which was generated during the earlier development batch and was later used in the evaluation set after replacing an evaluation book with sensitive content.

\paragraph{Runs summary.} We report the details of all runs in \autoref{tab:agentic-corpus-run-overview}, including the number of chunks and excerpts which pass acceptance gate within the enforced limits. We note that even when the final translation failed to pass within the enforced limits we observe that increasing the limits did not show justifiable improvements, while the ``failed'' translation was often acceptable.

\begin{table}[t]
\centering
\footnotesize
\resizebox{\columnwidth}{!}{%
\begin{tabular}{@{}lll@{}}
\toprule
Group & Value & Runs \\
\midrule
Run group & Evaluation books & 15 \\
Run group & Development books & 16 \\
Run group & Case study evaluation books & 4 \\
Source language & French & 10 \\
Source language & Japanese & 13 \\
Source language & Polish & 12 \\
Target language & English & 31 \\
Target language & French & 1 \\
Target language & Japanese & 1 \\
Target language & Polish & 1 \\
Target language & Spanish & 1 \\
Final Acceptance Gate & \fail & 22 \\
Final Acceptance Gate & \pass & 13 \\
Max chunk cycles & 2 & 18 \\
Max chunk cycles & 3 & 17 \\
\bottomrule
\end{tabular}
}
\caption{Corpus and run overview for the completed agentic pipeline runs.}
\label{tab:agentic-corpus-run-overview}
\end{table}

\begin{table}[t]
\footnotesize
\setlength{\tabcolsep}{3pt}
\begin{tabular}{@{}>{\raggedright\arraybackslash}p{0.30\columnwidth} r >{\raggedright\arraybackslash}p{0.46\columnwidth}@{}}
\toprule
Artifact / log unit & Count & Scope \\
\midrule
Completed pipeline runs & 35 & Evaluation, development, and multilingual target-language runs used in this paper. \\
Source-language chunks processed & 441 & All selected chunks entering the agentic pipeline. \\
Chunk-cycle translation/revision artifacts & 1,059 & Source chunks $\times$ completed chunk cycles. \\
Runs with job-metric logs & 31 & Excludes four development-book runs with unavailable metrics. \\
Logged agent jobs & 2,930 & All recorded style, translation, review, revision, and continuity jobs. \\
Failed agent jobs & 19 & Logged failed jobs across runs with metrics. \\
Final-gate flagged chunks & 74 & Chunks flagged by the final acceptance gate. \\
\bottomrule
\end{tabular}
\caption{Artifact counts generated by the selected agentic pipeline runs.}
\label{tab:artifact-footprint}
\end{table}

\paragraph{Artifact footprint.}
The selected agentic translation pipeline outputs comprise 35 completed runs: 15 evaluation-book runs, 16 development-book runs, and 4 multilingual target-language runs. Across these runs, the pipeline processed 441 source-language chunks and produced 1,059 chunk-cycle translation or revision artifacts, counting each completed translation or revision cycle for a chunk once. Job-level logs were available for 31 runs and record 2,930 agent jobs, including 19 failed jobs; four development-book runs were completed but untracked in the job-metric logs. At the automatic final acceptance gate, 13 runs passed and 22 retained flagged issues, with 74 chunks flagged in total (\autoref{tab:artifact-footprint}).

\definecolor{pOneRow}{HTML}{F4F8FF}
\definecolor{pTwoRow}{HTML}{F7F4FF}
\definecolor{pThreeRow}{HTML}{FFF7E8}
\definecolor{htRow}{HTML}{F2F2F2}

\section{Human evaluation}
\label{sec:app_humeval}

In this section of the appendix, we add more details about our human evaluation protocol, including participant recruitment, participant profiles, guidelines, and interface screenshots.

\paragraph{Participant recruitment.} The participants for the experiments were recruited using a freelancing platform Upwork. We limit our search to avid readers, who read at least one book in English per month and have native-level command of the language. We asked them to complete an entry questionnaire (\autoref{tab:entry_questionnaire}). We report their demographics in \autoref{tab:participant-demographics} and AI use frequency in \autoref{tab:participant-ai-use}.

\begin{table*}[t]
\centering
\scriptsize
\setlength{\tabcolsep}{3pt}
\renewcommand{\arraystretch}{1.12}

\begin{tabular}{@{}r l l l p{0.11\textwidth} l p{0.10\textwidth} p{0.11\textwidth} p{0.12\textwidth} r@{}}
\toprule
\shortstack{ID}
& Age
& Gender
& Education
& Nationality
& \shortstack{Native\\language}
& \shortstack{Additional\\languages}
& \shortstack{English literary\\fiction}
& \shortstack{Translated fiction\\in English}
& \shortstack{Books\\per year} \\
\midrule

1  & 25--34 & Woman & Undergraduate & American      & English & None           & Weekly or more       & Almost never         & 100 \\
2  & 25--34 & Woman & Undergraduate & American      & English & None           & Weekly or more       & A few times a year   & 40  \\
3  & 18--24 & Man   & Undergraduate & South African & English & None           & A few times a month  & Not sure             & 12  \\
4  & 45+    & Woman & Undergraduate & South African & English & None           & About once a month   & A few times a year   & 12  \\
5  & 45+    & Woman & High school    & American      & English & None           & Weekly or more       & Almost never         & 12  \\
6  & 45+    & Woman & Master's       & South African & English & None           & Weekly or more       & A few times a year   & 50  \\
7  & 35--44 & Woman & Doctoral       & Canadian      & English & French         & About once a month   & Almost never         & 24  \\
8  & 45+    & Woman & Undergraduate & American      & English & None           & Weekly or more       & A few times a month  & 75  \\
9  & 25--34 & Woman & Undergraduate & South African & English & None           & Weekly or more       & A few times a year   & 120 \\
10 & 25--34 & Woman & Undergraduate & American      & Spanish & None           & Weekly or more       & A few times a year   & 35  \\
11 & 25--34 & Woman & Undergraduate & American      & English & None           & Weekly or more       & A few times a year   & 100 \\
12 & 45+    & Woman & High school    & American      & English & None           & Weekly or more       & Almost never         & 103 \\
13 & 45+    & Woman & Undergraduate & British       & English & French         & A few times a month  & About once a month   & 20  \\
14 & 35--44 & Woman & Master's       & Polish        & Polish  & Polish; French & Weekly or more       & Weekly or more       & 80  \\
15 & 45+    & Woman & Doctoral       & American      & English & None           & Weekly or more       & About once a month   & 100 \\

\bottomrule
\end{tabular}

\caption{Participant demographics and reading profile. All participants reported that they primarily read in English and were very comfortable reading long literary texts in English. Nationality entries are shortened for display.}
\label{tab:participant-demographics}
\end{table*}

\begin{table*}[t]
\centering
\scriptsize
\setlength{\tabcolsep}{4pt}
\renewcommand{\arraystretch}{1.12}

\begin{tabular}{@{}r l p{0.28\textwidth} p{0.48\textwidth}@{}}
\toprule
\shortstack{ID}
& \shortstack{AI-use\\frequency}
& Models used most often
& Reported uses \\
\midrule

1  & Sometimes
   & ChatGPT
   & Deep research and summarization \\

2  & Rarely
   & Google
   & Does not use AI \\

3  & Often
   & ChatGPT; Grok Imagine; Nano Banana
   & Deep research and summarization; job search and career support; image generation \\

4  & Often
   & ChatGPT
   & Writing and editing \\

5  & Often
   & ChatGPT
   & Web search; personal advice; job search and career support \\

6  & Often
   & ChatGPT; Grok
   & Web search; image generation \\

7  & Often
   & Claude Opus; Claude Sonnet
   & Deep research and summarization; writing and editing \\

8  & Often
   & Claude; ChatGPT; Grok
   & Deep research and summarization; writing and editing; translation \\

9  & Always
   & ChatGPT; Claude; Kiro
   & Deep research and summarization; writing and editing \\

10 & Sometimes
   & Gemini
   & Deep research and summarization; writing and editing \\

11 & Never
   & None
   & Does not use AI \\

12 & Always
   & Claude Sonnet 4.6; ChatGPT 5.3
   & Deep research and summarization; writing and editing \\

13 & Never
   & None
   & Does not use AI \\

14 & Rarely
   & ChatGPT
   & Web search \\

15 & Always
   & ChatGPT; Claude; DeepSeek
   & Deep research and summarization; writing and editing \\

\bottomrule
\end{tabular}

\caption{Self-reported AI use.}
\label{tab:participant-ai-use}
\end{table*}

\paragraph{Procedure.}\label{sec:consent-form} Before the experiment all participants were asked to sign the consent form (see \autoref{tab:consent_form} and \autoref{tab:consent_form_questions}). They were also asked to fill in a demographic survey (\autoref{tab:entry_questionnaire}). All surveys were administered through Microsoft Forms. 
During the recruitment stage, we additionally ask them to report their reading habits.

\paragraph{Questions.} During the human evaluation, the readers fill in three types of questionnaires embedded in the interface. In the \textit{immersive reading} they are asked to fill in a questionnaire about each text separately (single-reading; \autoref{tab:isolated_reading_questions}), and then about both texts (comparison; \autoref{tab:comparison_questions}). In the \textit{close reading} each chunk pair is accompanied by a short questionnaire asking them to compare both translations (\autoref{tab:chunk_questions}).

\begin{table*}[p]
\centering
\scriptsize
\setlength{\tabcolsep}{4pt}
\renewcommand{\arraystretch}{0.94}
\setlist[itemize]{nosep,leftmargin=1.3em,topsep=0pt,parsep=0pt,partopsep=0pt}
\begin{tabular}{@{}>{\ttfamily\raggedright\arraybackslash}p{0.96\textwidth}@{}}
\toprule
\multicolumn{1}{c}{\bfseries Consent Form} \\
\midrule
{\bfseries Consent Form - Evaluating machine translation of literary texts} (Study \#[REDACTED])\\[0.15em]

{\bfseries Thank you for your interest in our study!}\\[0.15em]

{\bfseries Who is conducting the study?} The co-principal investigators (co-PIs) are [REDACTED] and [REDACTED]. Another researcher involved in data collection is [REDACTED], a student at [REDACTED].\\[0.15em]

{\bfseries Contact:} [REDACTED] or [REDACTED]. You can also reach out to us via the Upwork messaging tool.\\[0.15em]

{\bfseries Who is funding this study?} [REDACTED] awarded to [REDACTED].\\[0.15em]

{\bfseries Why are we doing this study?} We want to investigate the quality of literary machine translation (i.e., Can AI translate literary texts as well as humans)\\[0.15em]

{\bfseries Who can participate in this study?} Participants are adults who are comfortable reading long literary texts in English (up to 8,000 words) and who read fiction in English regularly.\\[0.15em]

{\bfseries Your participation is voluntary.} You have the right to refuse to participate in this study. Refusal to participate or withdrawal/dropout after agreeing to participate and signing this consent form will not have any adverse effects or consequences.\\[0.15em]

{\bfseries How is the study done?} This study takes place online and will last about 4 to 4.5 hours, divided into 2 sessions over a period of a few days. If you agree to take part: (1) You will first {\bfseries read this consent form} and indicate whether you agree to participate. (2) You will use your own device to complete the study online. {\bfseries A personal computer is recommended.} (3) You will complete a {\bfseries short questionnaire about your demographic}, language background, and reading habits (\textasciitilde{}5min). {\bfseries In Part 1 (about 1.5 hours),} you will read a long excerpt from a novel in translation (\textasciitilde{}8,000 words in English), answer a small number of questions about it, provide a written comment, and indicate whether you think it was translated by a person or produced using artificial intelligence. You will then read a second translation of the same excerpt and complete the same set of questions. At the end of Part 1, you will choose which version you preferred overall and explain why. You are encouraged to take a break between the readings. {\bfseries In Part 2 (about 2.5 hours),} which will take place a few days later, you will compare the two translations side by side in shorter chunks. For each chunk, you will highlight good and bad phrasing, choose the version you prefer, briefly explain why, and indicate whether the choice felt easy or difficult. You may take breaks during the study whenever needed. We may ask you for additional clarification if your comments are unclear.\\[0.15em]
{\bfseries Risks, benefits, and your privacy}\\[0.15em]

{\bfseries What are the risks of participating?} This study poses a minimal risk: none of the tasks involved in the experiment will have any direct or lasting impacts on participants. Possible inconveniences include fatigue or boredom, and we encourage taking breaks to prevent these.\\[0.15em]

{\bfseries What are the benefits of participating?} We do not think that taking part in this study will have direct benefits for you. However, in the future, others may benefit from what we learn from this study. Specifically, we think that this study might help us understand more about the quality of translation and how automatic machine translation works.\\[0.15em]

{\bfseries Will you be paid for taking part in this research study?} You will receive payment of \$110 USD for the entire task (part 1 and part 2). You will receive the money within one week of submitting your answers via the Upwork payment system. If you withdraw before completing the study, compensation will be prorated based on the amount of the study completed.\\[0.15em]

{\bfseries How will your privacy be maintained?} Your confidentiality will be respected. Information that discloses your identity will not be released without your consent.
\begin{itemize}
    \item All participants will be identified only by a unique code number.
    \item Participants will not be identified by name in any reports of the completed study. Name will be collected and stored by the PIs in a separate spreadsheet, and only for the purposes of matching name and code in case you decide to withdraw.
    \item The study involves a questionnaire. The initial demographic questionnaire will be collected via Microsoft Forms, which stores data on Canadian servers. This data will be stored in a secure [REDACTED]-hosted server to ensure your information remains within our local, protected environment. Consent forms and study responses will initially be collected through secure online tools used for this study. After collection, the data will be downloaded and stored on a password-protected [REDACTED] researcher's computer.
    \item Access to study data (identifiable only through participant codes) will be limited to the PI and research assistants, all of whom have completed the required ethics training.
\end{itemize}\\[-0.1em]

{\bfseries What if you decide to withdraw your consent to participate?} You may withdraw from this study at any time without giving any reasons, up to two weeks from the date next to your signature in the consent form. If you choose to enter the study and withdraw later, all data collected about you will be destroyed. Please contact us through email ([REDACTED] or [REDACTED]) and state that you wish to withdraw your data from this study. Please include the date of your participation. After two weeks have passed from the date of collection, we will not be able to accommodate a withdrawal request, as study data will already have been used in presentations and publications.\\[0.25em]

{\bfseries Results, contact information, and future data usage}\\[0.15em]

{\bfseries Study results.} The main findings will be reported in academic presentations and publications.\\[0.15em]

{\bfseries Contact for Complaints.} If you have any concerns about your rights as a research participant and/or your experiences while participating in this study, please contact the Director, Research Ethics at [REDACTED] or [REDACTED].\\[0.15em]

{\bfseries Future use of participant data.} The data will be coded and stored on [REDACTED] servers. No identifying information will be shared outside of our research group. All data that we collect will be used for research purposes only. De-identified study data will be made available to other researchers for future, unspecified use in academic publications and open-access servers. While limited study data will be accessible to others once published (e.g., age, gender, language background), {\bfseries we will never share identifying or confidential information.}\\
\bottomrule
\end{tabular}
\caption{Consent form presented to the participants prior to the evaluation.}
\label{tab:consent_form}
\end{table*}

\begin{table*}[p]
\centering
\scriptsize
\setlength{\tabcolsep}{4pt}
\renewcommand{\arraystretch}{0.94}
\begin{tabular}{@{}>{\ttfamily\raggedright\arraybackslash}p{0.96\textwidth}@{}}
\toprule
\multicolumn{1}{c}{\bfseries Consent Form Questions} \\
\midrule
{\bfseries Participant consent and signature.} Taking part in this study is entirely up to you: You have the right to refuse to participate. If you decide to take part, you may pull out of the study at any time without giving a reason and without any negative impact by sending an email to [REDACTED] or [REDACTED]. Your continuation on the online experiment platform indicates that you received a copy of this consent form for your own records and that you consent to participate in the study. By consenting, you acknowledge that you have not waived any rights to legal recourse in the event of research-related harm.\\[0.35em]

\midrule

{\normalfont
\begin{minipage}{0.96\textwidth}
\renewcommand{\arraystretch}{1.08}
\setlength{\tabcolsep}{3pt}
\footnotesize
\begin{tabularx}{\linewidth}{@{}
    >{\raggedright\arraybackslash}p{0.045\linewidth}
    >{\raggedright\arraybackslash}p{0.39\linewidth}
    >{\raggedright\arraybackslash}X
@{}}
     &
    \textbf{Primary Study Participation} (\textit{Required})
    &
    \textbf{YES} – I have read and understood this Consent Form and agree to participate in the study

    \textbf{NO} – I do not wish to participate in the study.
    \\

     &
    \textbf{Study Results} (\textit{Optional})
    &
    \textbf{YES} – I would like to be notified of the results or findings of this study (e.g., via a summary report or infographic). I understand that the researchers will contact me through the secure Upwork messaging app.

    \textbf{NO} – I do not wish to be contacted with the study results.
    \\

     &
    \textbf{Please type your full name} (\textit{Required})
    &
    Open-ended response
    \\

     &
    \textbf{Date} (\textit{Required})
    &
    Date input field
    \\
\end{tabularx}
\end{minipage}
}\\
\bottomrule
\end{tabular}
\caption{Consent-form response fields shown to participants.}
\label{tab:consent_form_questions}
\end{table*}

\newcolumntype{P}{>{\raggedright\arraybackslash\ttfamily\footnotesize}p{\linewidth}}

\begin{table*}
\renewcommand{\arraystretch}{0.92}
\begin{tabular}{@{}P@{}}

    \toprule
    \multicolumn{1}{c}{\bfseries\footnotesize Entry Questionnaire} \\
    \midrule

    {\bfseries Entry Questionnaire}\\[0.3em]

    {\bfseries Thank you for your interest in participating in our study!}\\[0.15em]

    This short questionnaire helps us understand your background, language experience, and reading habits. It should take only {\bfseries \textasciitilde{}5 minutes} to complete.\\[0.15em]

    Your responses will be used solely for research purposes and to help us ensure that you meet the requirements of this study.\\[0.15em]

    Thank you for your time and we appreciate your help!\\[0.25em]

    \midrule

    {\normalfont
    \begin{minipage}{\linewidth}
    \renewcommand{\arraystretch}{1.08}
    \setlength{\tabcolsep}{2pt}
    \scriptsize

    \begin{tabularx}{\linewidth}{@{}
        >{\centering\arraybackslash}p{0.025\linewidth}
        >{\raggedright\arraybackslash}p{0.19\linewidth}
        >{\raggedright\arraybackslash}p{0.245\linewidth}
        @{\hspace{0.015\linewidth}}
        >{\centering\arraybackslash}p{0.025\linewidth}
        >{\raggedright\arraybackslash}p{0.19\linewidth}
        >{\raggedright\arraybackslash}X
    @{}}
        \textbf{\#} & \textbf{Question} & \textbf{Response Options}
        &
        \textbf{\#} & \textbf{Question} & \textbf{Response Options}
        \\[0.25em]

        1 &
        \textbf{Your ID}
        &
        Open-ended response
        &
        8 &
        \textbf{How do you use AI models?} (if at all)
        &
        Open-ended response
        \\[0.25em]

        2 &
        \textbf{What is your age range?}
        &
        18-24; 25-34; 35-44; 45+; Prefer not to say
        &
        9 &
        \textbf{What is your native language?}
        &
        English; Other (open-ended response)
        \\[0.25em]

        3 &
        \textbf{What is your gender?}
        &
        Woman; Man; Non-binary; Prefer not to say; Other (open-ended response)
        &
        10 &
        \textbf{Do you primarily read in English?}
        &
        Yes; No
        \\[0.25em]

        4 &
        \textbf{What is the highest level of education you are currently in, or have completed?}
        &
        High school or equivalent; College / university undergraduate; Master's degree; Doctoral degree; Prefer not to say; Other (open-ended response)
        &
        11 &
        \textbf{How comfortable are you reading long literary texts in English?}
        &
        Very uncomfortable; Somewhat uncomfortable; Neither comfortable nor uncomfortable; Somewhat comfortable; Very comfortable
        \\[0.25em]

        5 &
        \textbf{What is your nationality?}
        &
        Open-ended response
        &
        12 &
        \textbf{Do you know any of these languages (pre-intermediate level or above)?} Select all that apply.
        &
        Japanese; Polish; French; None of these
        \\[0.25em]

        6 &
        \textbf{How often do you use AI models} (such as ChatGPT or Gemini)?
        &
        Never; Rarely; Sometimes; Often; Always
        &
        13 &
        \textbf{How often do you read literary fiction in English?}
        &
        Almost never; A few times a year; About once a month; A few times a month; Weekly or more
        \\[0.25em]

        7 &
        \textbf{What AI models do you use most often?} (if any)
        &
        Open-ended response
        &
        14 &
        \textbf{How often do you read translated fiction in English?}
        &
        Almost never; A few times a year; About once a month; A few times a month; Weekly or more; Not sure
        \\

    \end{tabularx}
    \end{minipage}
    }\\

    \bottomrule

\end{tabular}
\caption{Entry questionnaire presented to the participants prior to the evaluation.}
\label{tab:entry_questionnaire}
\end{table*}

\begin{table*}[t]
    \centering
    \small
    \renewcommand{\arraystretch}{1.15}
    \setlength{\tabcolsep}{6pt}
    \begin{tabularx}{\textwidth}{
        @{}
        >{\raggedright\arraybackslash}p{0.04\textwidth}
        >{\raggedright\arraybackslash}p{0.48\textwidth}
        >{\raggedright\arraybackslash}X
        @{}
    }
        \toprule
        \textbf{\#} & \textbf{Question} & \textbf{Response Options} \\
        \midrule

        1\phantomsection\label{i-Q1} &
        \textbf{How acceptable did this translation feel overall as a published version of the book?}\newline
        \emph{Please think about your overall impression of this version as a published translation.} &
        completely unacceptable\newline
        somewhat unacceptable\newline
        acceptable with some issues\newline
        mostly acceptable\newline
        fully acceptable \\
        \addlinespace[6pt]

        2\phantomsection\label{i-Q2} &
        \textbf{How smooth was this text to read?}\newline
        \emph{Please think about how naturally and clearly the sentences flowed while you were reading.} &
        very unsmooth\newline
        somewhat unsmooth\newline
        neutral\newline
        somewhat smooth\newline
        very smooth \\
        \addlinespace[6pt]

        3\phantomsection\label{i-Q3} &
        \textbf{To what extent did the wording and style help you stay immersed in the story?}\newline
        \emph{Please think about whether the wording and style helped you stay absorbed in the story, or pulled you out of it.} &
        interfered a lot with immersion\newline
        interfered somewhat\newline
        made little difference\newline
        supported immersion somewhat\newline
        supported immersion a lot \\
        \addlinespace[6pt]

        4\phantomsection\label{i-Q4} &
        \textbf{Would you want to continue reading the book in this version?}\newline
        \emph{If you were to continue reading, would you be willing to continue this version given its wording and style?} &
        definitely not\newline
        probably not\newline
        not sure\newline
        probably yes\newline
        definitely yes \\
        \addlinespace[6pt]

        5\phantomsection\label{i-Q5} &
        \textbf{What, if anything, stood out positively in the wording or style of this version?} &
        Open-ended response \\
        \addlinespace[6pt]

        6\phantomsection\label{i-Q6} &
        \textbf{What, if anything, stood out negatively in the wording or style of this version?} &
        Open-ended response \\
        \addlinespace[6pt]

        7\phantomsection\label{i-Q7} &
        \textbf{Do you think this version was\ldots} &
        human-translated\newline
        machine-translated \\
        \addlinespace[6pt]

        8\phantomsection\label{i-Q8} &
        \textbf{How confident are you in that guess?}\newline
        \emph{(Regarding Question 7)} &
        not at all confident\newline
        slightly confident\newline
        moderately confident\newline
        very confident\newline
        extremely confident \\

        \bottomrule
    \end{tabularx}

    \caption{Single-reading questionnaire shown after reading each excerpt in isolation. Questions Q1--Q4 collect ordinal ratings of acceptability, smoothness, immersion, and willingness to continue reading; Q5--Q8 collect free-text explanations, origin guesses, and confidence.}
    \label{tab:isolated_reading_questions}
\end{table*}

\begin{table*}[t]
    \centering
    \small
    \renewcommand{\arraystretch}{1.15}
    \setlength{\tabcolsep}{6pt}
    \begin{tabularx}{\textwidth}{
        @{}
        >{\raggedright\arraybackslash}p{0.04\textwidth}
        >{\raggedright\arraybackslash}X
        >{\raggedright\arraybackslash}p{0.25\textwidth}
        @{}
    }
        \toprule
        \textbf{\#} & \textbf{Question} & \textbf{Response Options} \\
        \midrule

        1\phantomsection\label{co-Q1} &
        \textbf{Which translation handled dialogue better overall?}\newline
        \emph{[Dialogue]} &
        Translation 1\newline
        Translation 2\newline
        No meaningful difference \\
        \addlinespace[6pt]

        2\phantomsection\label{co-Q2} &
        \textbf{Which translation's word choice felt more varied and/or expressive?}\newline
        \emph{[Word choice]} &
        Translation 1\newline
        Translation 2\newline
        No meaningful difference \\
        \addlinespace[6pt]

        3\phantomsection\label{co-Q3} &
        \textbf{Which translation would you prefer to continue reading more?} &
        Translation 1 (clearly better)\newline
        Translation 1 (slightly better)\newline
        Translation 2 (slightly better)\newline
        Translation 2 (clearly better) \\
        \addlinespace[6pt]

        4\phantomsection\label{co-Q4} &
        \textbf{Why? Please motivate your choice in 3--5 sentences.}\newline
        \emph{Tell us why you would prefer the translation you selected. How was it better than the other one? Use T1 to refer to Translation 1 and T2 to refer to Translation 2.} &
        Open-ended response \\
        \addlinespace[6pt]

        5\phantomsection\label{co-Q5} &
        \textbf{Which version is more likely to be AI-translated?} &
        Translation 1\newline
        Translation 2 \\
        \addlinespace[6pt]

        6\phantomsection\label{co-Q6} &
        \textbf{How confident are you in that choice?}\newline
        \emph{(Regarding Question 5)} &
        not at all confident\newline
        slightly confident\newline
        moderately confident\newline
        very confident\newline
        extremely confident \\
        \addlinespace[6pt]

        7\phantomsection\label{co-Q7} &
        \textbf{Why do you think it is more likely to be AI-translated?}\newline
        \emph{What are the indicators that make you think so?} &
        Open-ended response \\

        \bottomrule
    \end{tabularx}

    \caption{Comparison questionnaire shown after participants read the paired HT and MT excerpts for the same book. The questions collect translation preference, preference strength, reasoning, perceived machine-translation origin, and confidence.}
    \label{tab:comparison_questions}
\end{table*}

\begin{table*}[t]
    \centering
    \small
    \renewcommand{\arraystretch}{1.15}
    \setlength{\tabcolsep}{4pt}
    \begin{tabularx}{\textwidth}{
        @{}
        >{\raggedright\arraybackslash}p{0.035\textwidth}
        >{\raggedright\arraybackslash}p{0.43\textwidth}
        >{\raggedright\arraybackslash}X
        @{}
    }
        \toprule
        \textbf{\#} & \textbf{Question} & \textbf{Response Options} \\
        \midrule

        1\phantomsection\label{ch-Q1} &
        \textbf{Step 1.} Highlight wording quality\newline
        \emph{Read both translations and mark phrases that sound especially \textbf{good} or \textbf{poor}. Select text, then choose a label in the popover. Click an existing highlight to remove it.} &
        Open-ended reading and highlighting \\
        \addlinespace[6pt]

        2\phantomsection\label{ch-Q2} &
        \textbf{Step 2.} Which translation do you prefer? &
        T1\newline
        T2 \\
        \addlinespace[6pt]

        3\phantomsection\label{ch-Q3} &
        \textbf{Step 3.} Explain your choice (refer to translations as T1 and T2). &
        Open-ended response \\
        \addlinespace[6pt]

        4\phantomsection\label{ch-Q4} &
        \textbf{Step 4.} Was this choice difficult? &
        No, because the translation I chose was significantly better.\newline
        No, because the translation I chose was better.\newline
        Yes, because both translations were of similar quality. \\

        \bottomrule
    \end{tabularx}

    \caption{Close-reading per-chunk questionnaire. Participants choose the preferred translation for each aligned chunk, report preference strength, optionally highlight supporting spans, and explain the judgment.}
    \label{tab:chunk_questions}
\end{table*}

\paragraph{Interface.}\label{sec:eval_interface} 
Participant-facing guidelines and interface screenshots are provided in the supplementary materials section of the GitHub repository (\href{https://github.com/Yves575/lait}{\path{github.com/Yves575/lait}}).

\paragraph{Display normalization.} We applied lightweight display normalization to reduce typographic inconsistencies across texts, including normalizing quotation marks, guillemets, apostrophes, and ellipses. This normalization was used for presentation and annotation consistency and was applied directly in the evaluation interface without affecting the source files.  The same displayed text representation was used when recording span highlights, so participant offsets refer to the normalized interface text.

\section{Human evaluation results}
\label{sec:app_extra_humeval_res}

\begin{table}[t]
    \centering
    \scriptsize
    \setlength{\tabcolsep}{4pt}
    \renewcommand{\arraystretch}{1.08}
    \begin{tabular}{@{}lcc@{}}
        \toprule
        AI-use frequency & Correct / total & Accuracy \\
        \midrule
        Never     & 2 / 4  & 50\%  \\
        Rarely    & 4 / 4  & 100\% \\
        Sometimes & 2 / 4  & 50\%  \\
        Often     & 3 / 12 & 25\%  \\
        Always    & 6 / 6  & 100\% \\
        \bottomrule
    \end{tabular}
    \caption{MT-identification accuracy by AI-use frequency. Participant-level association between AI-use frequency and MT-identification accuracy was close to zero: Pearson $r = 0.013$; Spearman $\rho = 0.121$.}
    \label{tab:ai-use-mt-identification}
\end{table}

In this section of the appendix, we report additional results and descriptive statistics of the human evaluation data. 

\begin{figure}[t]
    \centering
    \includegraphics[
        width=0.94\columnwidth,
        height=0.24\textheight,
        keepaspectratio
    ]{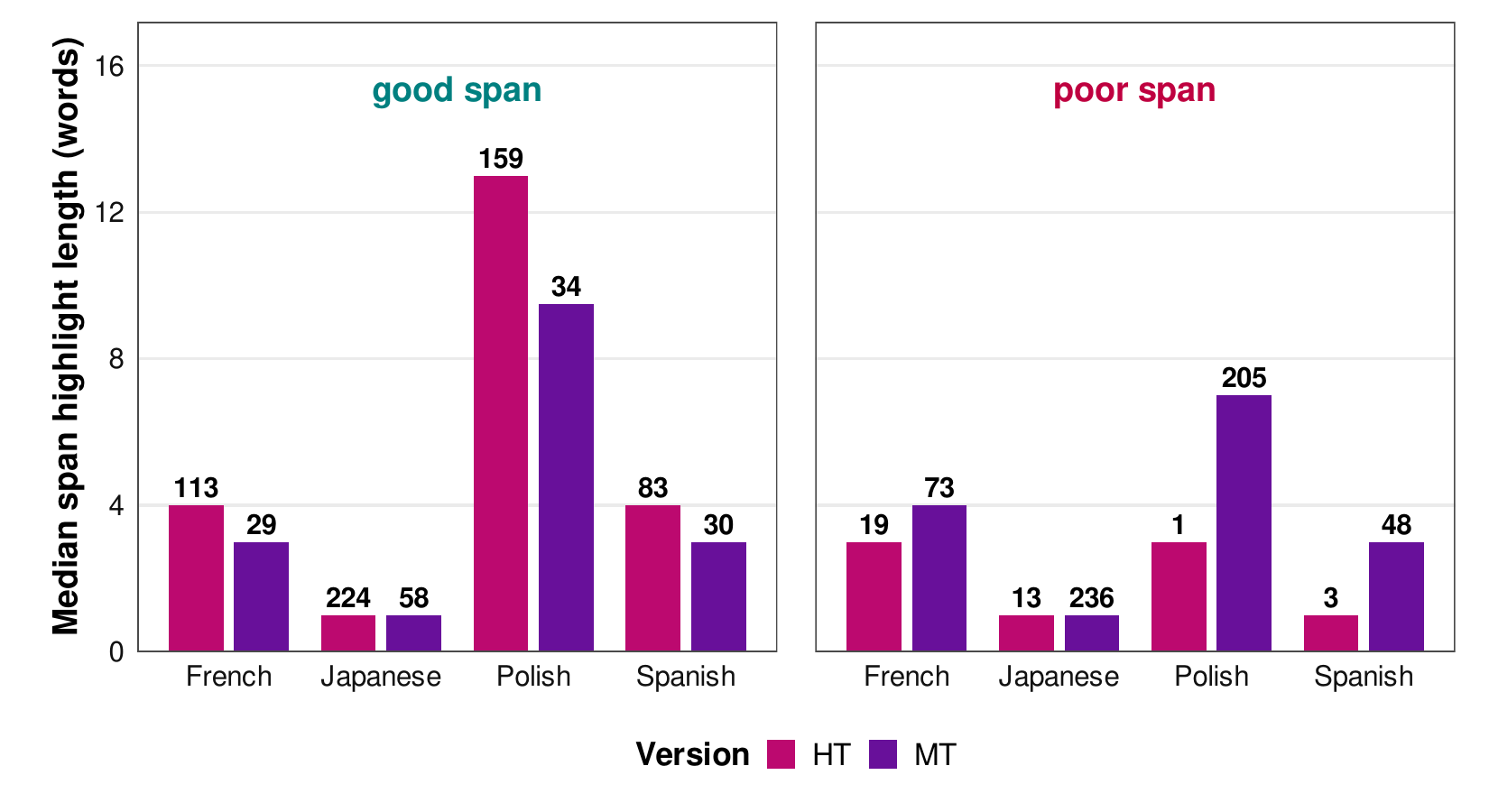}
    \caption{Median span highlight length in the multilingual case study. Bars show the median length of good and poor spans marked for HT and MT by target language.}
    \label{fig:span_highlight_median_length_by_language}
\end{figure}

\begin{figure}[t]
    \centering
    \includegraphics[
        width=0.94\columnwidth,
        height=0.24\textheight,
        keepaspectratio
    ]{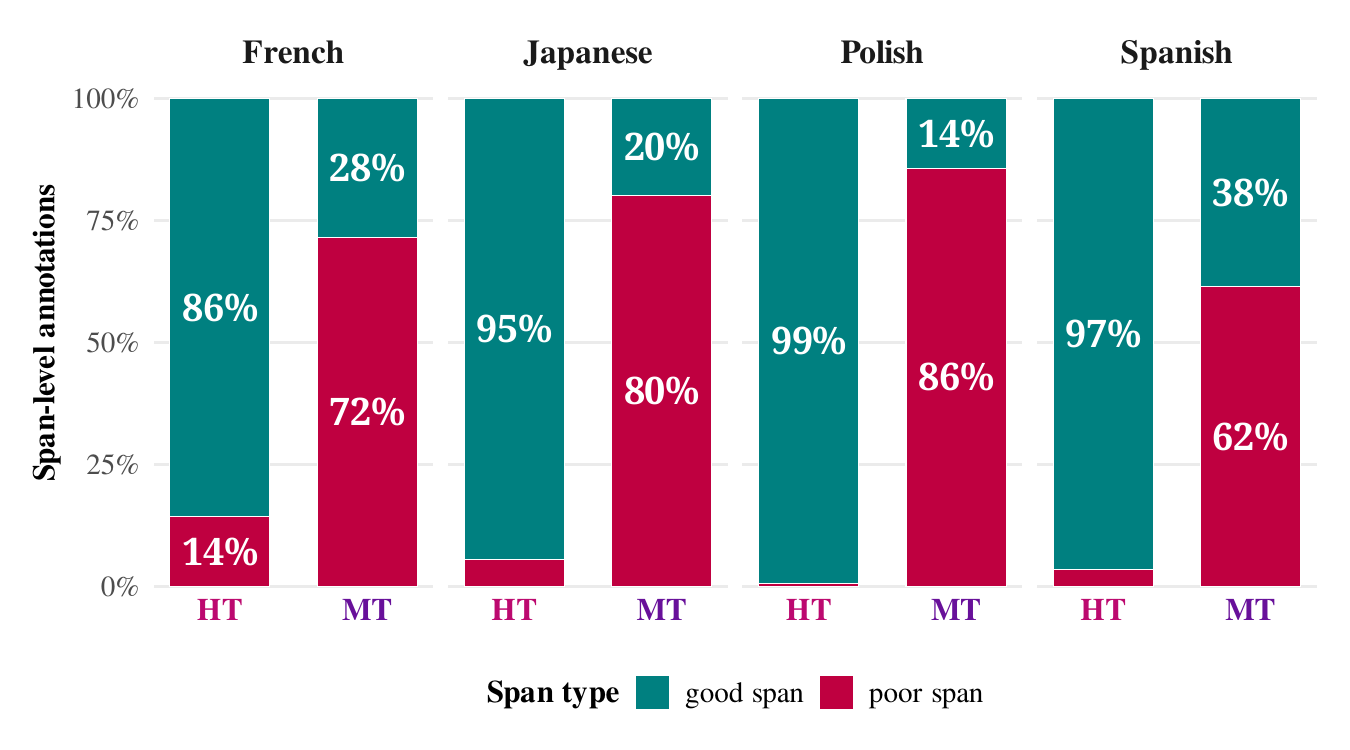}
    \caption{Span-level annotations by target language in the multilingual target-language case study. Bars show the proportion of good and poor spans marked in HT and MT for each target language.}
    \label{fig:span_highlights_by_language}
\end{figure}

\begin{figure}[t]
    \centering
    \includegraphics[
        width=0.94\columnwidth,
        height=0.30\textheight,
        keepaspectratio
    ]{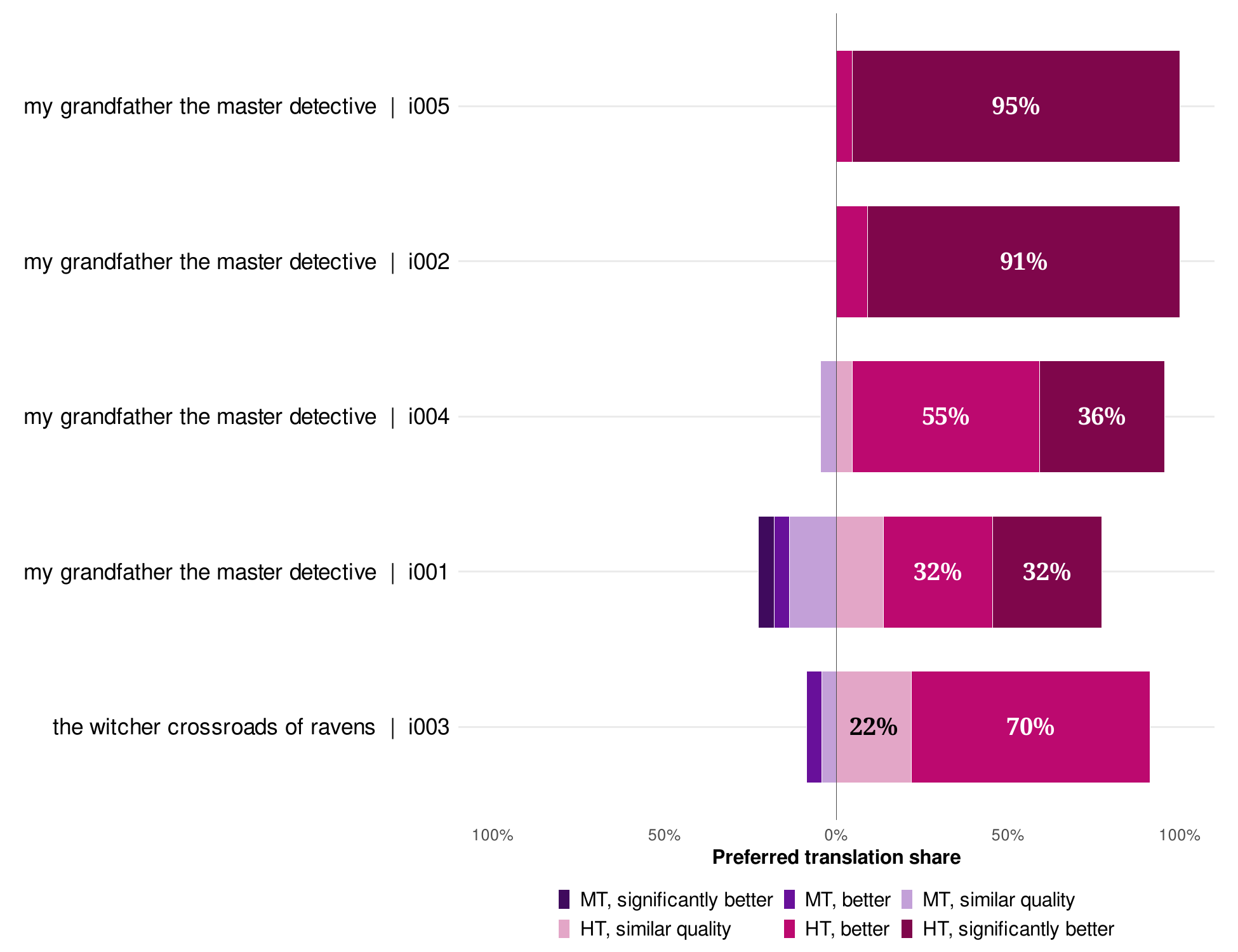}
    \caption{Book-level preferences in the multilingual case study.
    Each row is one book--participant pair; left/right bars indicate MT/HT preference, with darker shades marking stronger choices.}
    \label{fig:book_preference_diverging}
\end{figure}

\begin{figure*}[t]
    \centering
    \includegraphics[
        width=\textwidth,
        height=0.40\textheight,
        keepaspectratio
    ]{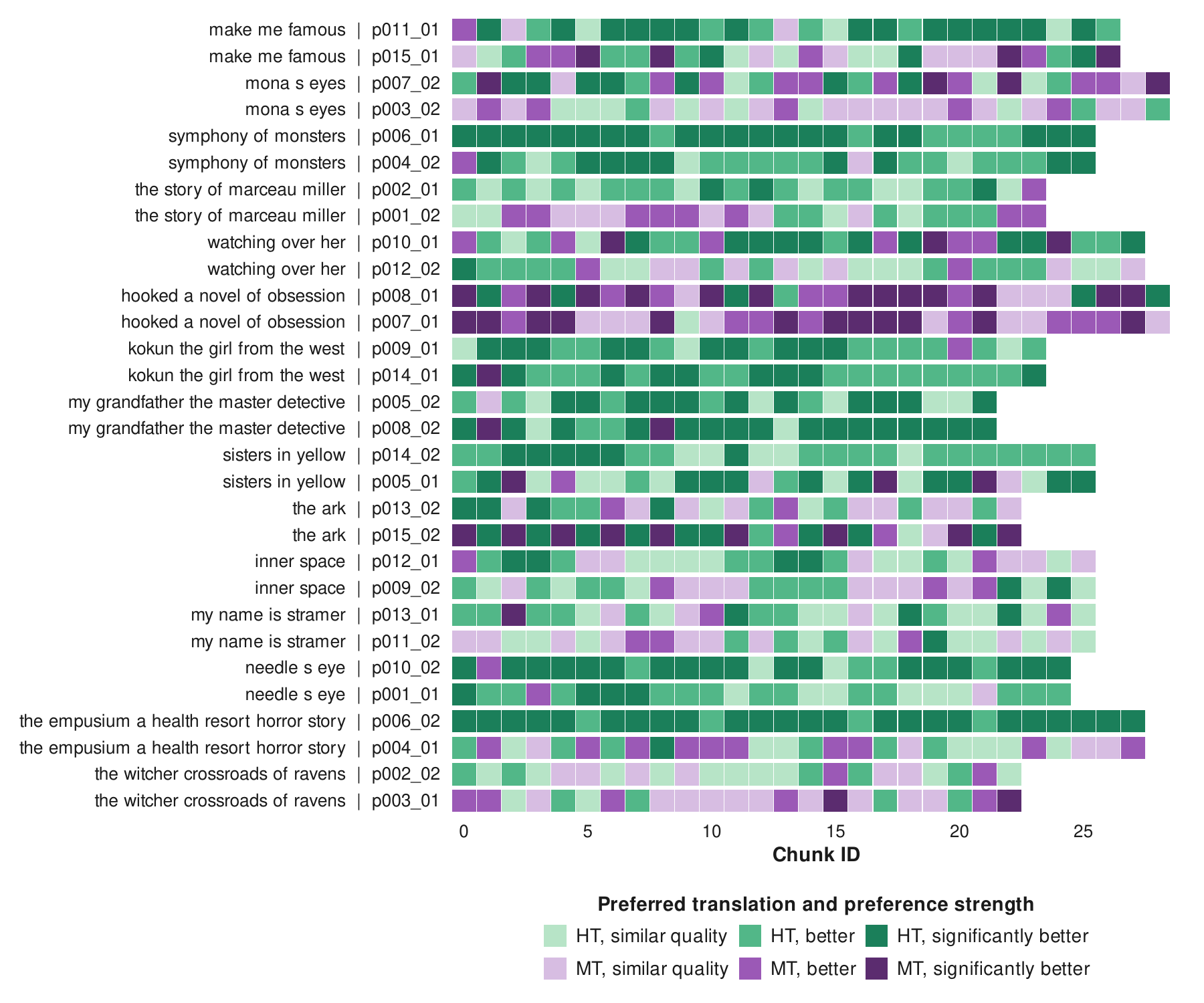}
    \caption{Close-reading chunk-level preferred translation. Each cell shows one chunk comparison for an excerpt and reader. Green indicates HT preference, purple indicates MT preference, and darker shades indicate stronger preference.}
    \label{fig:stat-part2_chunk_choice_heatmap}
\end{figure*}

\begin{figure}[t]
    \centering
    \includegraphics[
        width=0.94\columnwidth,
        height=0.30\textheight,
        keepaspectratio
    ]{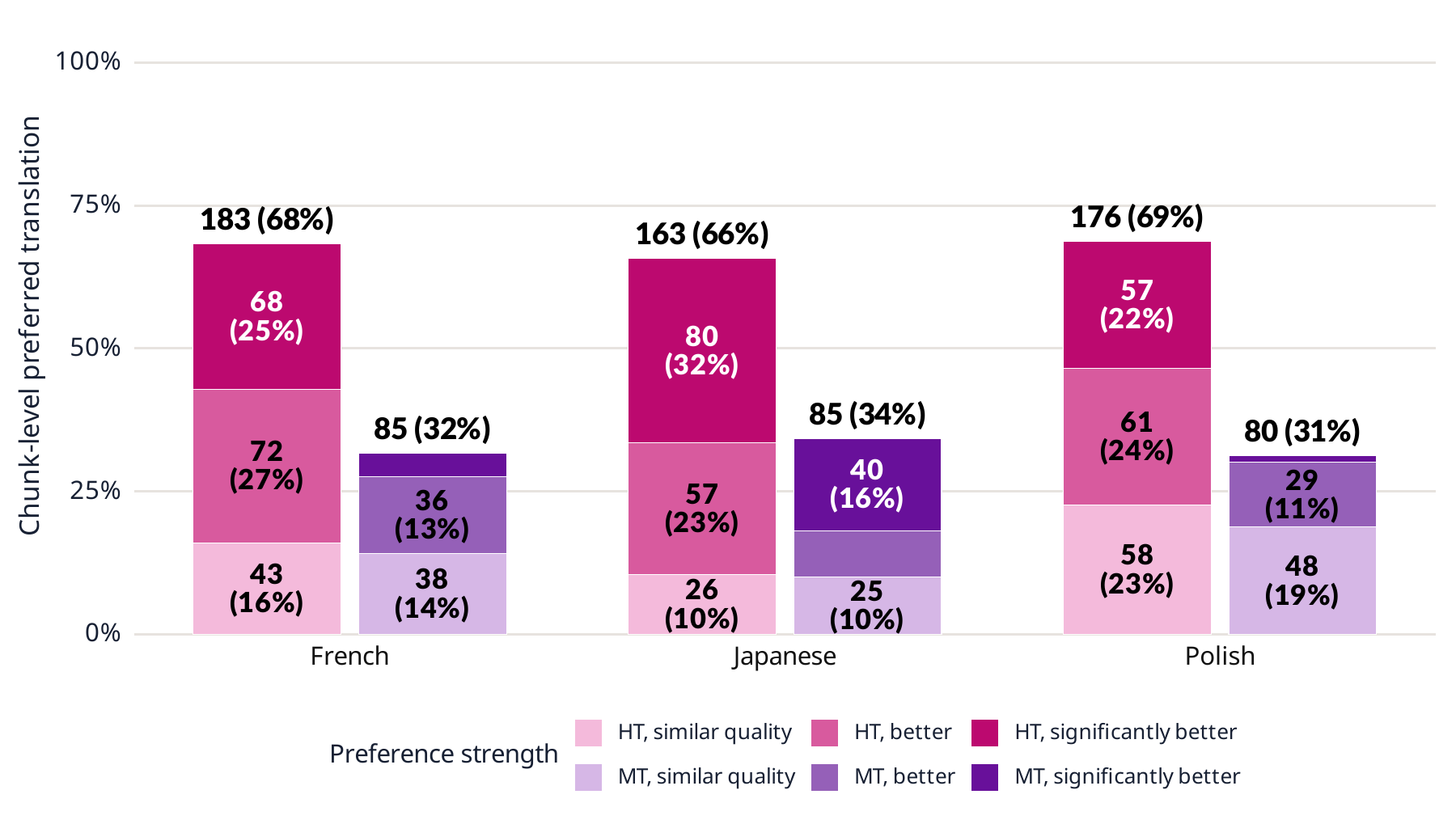}
    \caption{Close-reading chunk-level preferred translation by source language. Each pair of bars shows HT and MT preference shares within a source language; stacked segments show preference strength, and labels report counts and percentages.}
    \label{fig:stat-part2_preference_by_language}
\end{figure}

\paragraph{Overall results.} 
At the excerpt level, readers preferred HT more often for both dialogue handling and word choice (\autoref{fig:excerpt-attribute-pref}).
Close-reading judgments also favored HT overall: MT accounted for about one-third of preferences in each source language, although choices varied substantially across books and individual chunks (\autoref{fig:stat-part2_preference_by_language} and \autoref{fig:stat-part2_chunk_choice_heatmap}).
In the smaller multilingual case study, HT was preferred in 103 of 111 chunk judgments. 
The book-level results show the only excerpt-level MT preference, while the span annotations show mostly positive evidence for HT and negative evidence for MT across target languages (\autoref{fig:book_preference_diverging}, \autoref{fig:span_highlights_by_language}, and \autoref{fig:span_highlight_median_length_by_language}).

\begin{figure}[t]
    \centering
    \includegraphics[
        width=0.90\columnwidth,
        height=0.38\textheight,
        keepaspectratio
    ]{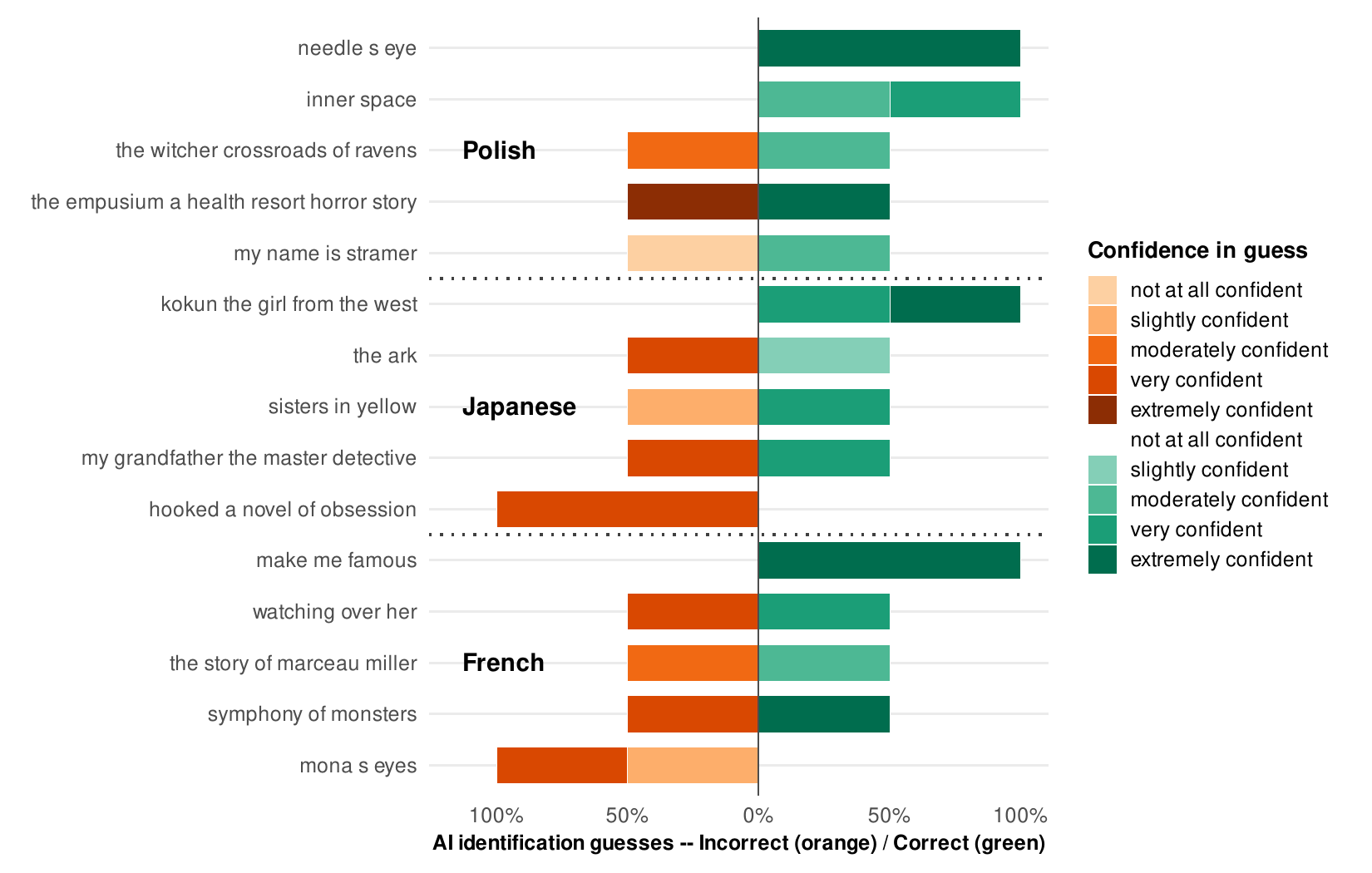}
    \caption{MT identification by book. Orange bars show HT identified as MT, green bars show MT identified as MT, and darker shades indicate higher confidence.}
    \label{fig:stat-book_ai_identification_diverging}
\end{figure}

\paragraph{Agreement.}\label{sec:app-agreement}
We report inter-reader agreement at three levels: (1) single-reading ratings, (2) binary preference and origin judgments (MT or HT), and (3) span highlights. For the \textbf{single-reading} 5-point ratings, we report Gwet's AC2 \citep{Gwet2021-ud} for \texttt{acceptability}, \texttt{smoothness}, \texttt{immersiveness}, \texttt{would continue} (\autoref{tab:single-reading-iaa-q1-q4}) as it is more fit for skewed distribution data than Krippendorff's alpha and handles ordinal variables. For \textbf{binary judgments}, we report the Gwet's AC1 (nominal data; \autoref{tab:comparison-chunk-binary-iaa}) for excerpt-level MT identification, excerpt-level preference, and chunk-level preference. Chunk-level agreement is reported for all chunks, for chunks excluding \textit{similar quality} judgments, and for the subset where both readers judged one translation as \textit{significantly better}. We also report chunk-level agreement measured separately by book (\autoref{tab:chunk-preference-iaa-by-book}), showing how it varies across the items and after excluding \textit{similar quality} judgments.

\begin{table}[t]
    \centering
    \begingroup
    \scriptsize
    \setlength{\tabcolsep}{2.2pt}
    \renewcommand{\arraystretch}{1.06}
    \newcommand{\acell}[2]{%
      \cellcolor[HTML]{#1}%
      \makebox[2.9em][r]{\textcolor{black}{\num{#2}}}%
    }
    \newcommand{\acslight}[1]{\acell{F8D7DA}{#1}}
    \newcommand{\acfair}[1]{\acell{FFF3CD}{#1}}
    \newcommand{\acmoderate}[1]{\acell{D1E7DD}{#1}}
    \newcommand{\acsubstantial}[1]{\acell{CFE2FF}{#1}}
    \newcommand{\acperfect}[1]{\acell{E2D9F3}{#1}}
    \begin{tabularx}{\columnwidth}{@{}Xrrrrr@{}}
        \toprule
        \textsc{Subset} & $n$ & \textsc{Exact} & $\leq 1$ & $M|d|$ & \textsc{AC2} \\
        \midrule
        \textbf{Q1--Q4 overall} & 120 & 35.0 & 64.2 & 1.16 & \acmoderate{0.426} \\
        {\color{teal}\hspace*{0.5em}HT only} & 60 & 38.3 & 70.0 & 0.98 & \acsubstantial{0.616} \\
        {\color{purple}\hspace*{0.5em}MT only} & 60 & 31.7 & 58.3 & 1.33 & \acfair{0.216} \\

        \addlinespace[0.2em]
        \textbf{Q1 acceptability} & 30 & 36.7 & 73.3 & 0.93 & \acsubstantial{0.632} \\
        {\color{teal}\hspace*{0.5em}HT only} & 15 & 40.0 & 80.0 & 0.80 & \acsubstantial{0.765} \\
        {\color{purple}\hspace*{0.5em}MT only} & 15 & 33.3 & 66.7 & 1.07 & \acmoderate{0.468} \\

        \addlinespace[0.2em]
        \textbf{Q2 smoothness} & 30 & 36.7 & 70.0 & 1.03 & \acmoderate{0.529} \\
        {\color{teal}\hspace*{0.5em}HT only} & 15 & 40.0 & 73.3 & 0.87 & \acsubstantial{0.725} \\
        {\color{purple}\hspace*{0.5em}MT only} & 15 & 33.3 & 66.7 & 1.20 & \acfair{0.290} \\

        \addlinespace[0.2em]
        \textbf{Q3 immersion} & 30 & 30.0 & 53.3 & 1.40 & \acfair{0.229} \\
        {\color{teal}\hspace*{0.5em}HT only} & 15 & 33.3 & 53.3 & 1.27 & \acmoderate{0.405} \\
        {\color{purple}\hspace*{0.5em}MT only} & 15 & 26.7 & 53.3 & 1.53 & \acslight{0.057} \\

        \addlinespace[0.2em]
        \textbf{Q4 continue} & 30 & 36.7 & 60.0 & 1.27 & \acfair{0.317} \\
        {\color{teal}\hspace*{0.5em}HT only} & 15 & 40.0 & 73.3 & 1.00 & \acmoderate{0.545} \\
        {\color{purple}\hspace*{0.5em}MT only} & 15 & 33.3 & 46.7 & 1.53 & \acslight{0.089} \\
        \bottomrule
    \end{tabularx}
    \caption{Inter-annotator agreement for single-reading ratings. Exact and $\leq 1$ are agreement percentages; $M|d|$ is mean absolute score difference; AC2 is Gwet's ordinal agreement coefficient.\protect\footnotemark}
    \label{tab:single-reading-iaa-q1-q4}
    \endgroup
\end{table}
\footnotetext{HT = human-translated; MT = machine-translated; $n$ = paired ratings. AC2 colors follow Landis and Koch heuristic bands: red = slight, yellow = fair, green = moderate, blue = substantial, purple = almost perfect.}

\begin{table}[t]
    \centering
    \begingroup
    \scriptsize
    \setlength{\tabcolsep}{2.2pt}
    \renewcommand{\arraystretch}{1.06}
    \newcommand{\acell}[2]{%
      \cellcolor[HTML]{#1}%
      \makebox[2.9em][r]{\textcolor{black}{\num{#2}}}%
    }
    \newcommand{\acslight}[1]{\acell{F8D7DA}{#1}}
    \newcommand{\acfair}[1]{\acell{FFF3CD}{#1}}
    \newcommand{\acmoderate}[1]{\acell{D1E7DD}{#1}}
    \newcommand{\acsubstantial}[1]{\acell{CFE2FF}{#1}}
    \newcommand{\acperfect}[1]{\acell{E2D9F3}{#1}}
    \begin{tabularx}{\columnwidth}{@{}Xrrrrrr@{}}
        \toprule
        \textsc{Measure} & $n$ & \textsc{Exact} & \textsc{Both HT} & \textsc{Both MT} & \textsc{Diff.} & \textsc{AC1} \\
        \midrule
        Excerpt: guessed MT & 15 & 40.0 & 2 & 4 & 9 & \acslight{-0.179} \\
        Excerpt: preferred & 15 & 40.0 & 5 & 1 & 9 & \acslight{-0.120} \\
        Chunk: preferred & 386 & 63.7 & 191 & 55 & 140 & \acfair{0.355} \\
        \hspace*{0.5em}excluding similar chunks & 194 & 72.2 & 116 & 24 & 54 & \acmoderate{0.546} \\
        \hspace*{0.5em}both significantly better & 44 & 79.5 & 28 & 7 & 9 & \acsubstantial{0.667} \\
        \bottomrule
    \end{tabularx}
    \caption{Inter-annotator agreement for excerpt- and chunk-level binary judgments.\protect\footnotemark}
    \label{tab:comparison-chunk-binary-iaa}
    \endgroup
\end{table}
\footnotetext{$n$ is paired judgments. Both HT, Both MT, and Diff. count paired annotator outcomes. For excerpt AI/MT identification, HT and MT refer to which version readers guessed was MT. AC1 colors follow the same heuristic bands as \autoref{tab:single-reading-iaa-q1-q4}.}
\begin{table*}[t]
    \centering
    \footnotesize
    \setlength{\tabcolsep}{4pt}
    \renewcommand{\arraystretch}{0.97}
    
    \begin{tabular}{llrrrrrr}
        \toprule
        Book & Lang & Pairs & Agr. & AC1 
        & \makecell[c]{Pairs excl.\ unsure\\[-1mm]\scriptsize$_{(\#p1/\#p2)}$}
        & \makecell[c]{Agr.\\excl.\ unsure}
        & \makecell[c]{AC1\\excl.\ unsure} \\
        \midrule
        
        Make Me Famous & FR & 27 & 44.4\% & 0.023 & 13$_{(5/11)}$ & 46.2\% & 0.112 \\
        Mona's Eyes & FR & 29 & 48.3\% & -0.015 & 8$_{(21/5)}$ & 50.0\% & 0.059 \\
        Symphony Of Monsters & FR & 26 & 92.3\% & 0.917 & 22$_{(4/0)}$ & 95.5\% & 0.952 \\
        The Story Of Marceau Miller & FR & 24 & 45.8\% & 0.050 & 9$_{(10/8)}$ & 55.6\% & 0.200 \\
        Watching Over Her & FR & 28 & 50.0\% & 0.113 & 11$_{(2/15)}$ & 54.5\% & 0.197 \\
        
        Hooked: A Novel Of Obsession & JA & 29 & 75.9\% & 0.694 & 19$_{(9/4)}$ & 73.7\% & 0.659 \\
        Kokun: The Girl From The West & JA & 24 & 91.7\% & 0.909 & 21$_{(3/0)}$ & 90.5\% & 0.895 \\
        My Grandfather, The Master Detective & JA & 22 & 95.5\% & 0.948 & 15$_{(6/2)}$ & 93.3\% & 0.929 \\
        Sisters In Yellow & JA & 26 & 76.9\% & 0.710 & 14$_{(9/5)}$ & 71.4\% & 0.622 \\
        The Ark & JA & 23 & 69.6\% & 0.391 & 11$_{(11/2)}$ & 63.6\% & 0.323 \\
        
        Inner Space & PL & 26 & 42.3\% & -0.076 & 8$_{(13/14)}$ & 75.0\% & 0.600 \\
        My Name Is Stramer & PL & 26 & 65.4\% & 0.354 & 5$_{(19/12)}$ & 60.0\% & 0.412 \\
        Needle's Eye & PL & 25 & 88.0\% & 0.865 & 17$_{(7/2)}$ & 88.2\% & 0.868 \\
        The Empusium: A Health Resort Horror Story & PL & 28 & 50.0\% & 0.200 & 17$_{(11/0)}$ & 41.2\% & -0.006 \\
        The Witcher: Crossroads Of Ravens & PL & 23 & 26.1\% & -0.467 & 4$_{(16/12)}$ & 75.0\% & 0.529 \\
        
        \bottomrule
    \end{tabular}
    
    \caption{Chunk-level preference agreement by book.}
    \label{tab:chunk-preference-iaa-by-book}
\end{table*}

For \textbf{span highlights}, we report overlap diagnostics rather than a chance-corrected coefficient.
A span is counted as overlapping when it intersects a span from the other reader on the same translation version; ``label-aware'' overlap additionally requires the same \textit{good}/\textit{poor} label.
Overall span overlap was 0.307, and label-aware overlap was 0.217. Readers who agreed on the preferred translation overlapped more in their highlighted evidence (0.322 span overlap; 0.263 label-aware overlap) than readers who disagreed (0.280 span overlap; 0.137 label-aware overlap).
\autoref{tab:quality-reason-preference-examples} shows examples and summarizes the top preference labels and representative comments.

\begin{table*}[p]
\centering
\scriptsize
\setlength{\tabcolsep}{4pt}
\renewcommand{\arraystretch}{1.05}
\begin{adjustbox}{max width=\textwidth, max totalheight=.88\textheight, keepaspectratio}
\begin{tabular}{@{}>{\raggedright\arraybackslash}p{0.12\textwidth} >{\raggedright\arraybackslash}p{0.055\textwidth} >{\raggedright\arraybackslash}p{0.15\textwidth} >{\raggedright\arraybackslash}p{0.50\textwidth} >{\centering\arraybackslash}p{0.05\textwidth}@{}}
        \toprule
        \textbf{Category} & \textbf{Code} & \textbf{Aspect} & \textbf{Definition} & \textbf{Pol.} \\
        \midrule
        \parbox[t]{0.12\textwidth}{\codelabel{A}.\\Language-level features} & \codelabel{A1} & Grammar \& spelling &
        Issues related to grammar such as verb-subject agreement, article use, or misspellings. &
        $+/-$ \\
         & \codelabel{A2} & Word choice: clarity &
        Whether individual word and phrase choices are clear or keep the reader guessing when it was not inteded. &
        $+/-$ \\
         & \codelabel{A3} & Word choice: naturalness &
        Whether the choice of specific words or phrases is natural in English at the local level. Covers words/phrases that are grammatically correct and even understandable, but sound off (e.g., awkward) in English. &
        $+/-$ \\
         & \codelabel{A4} & Word choice: idiomaticity &
        Correct use of English idioms, set phrases, conventional collocations, and figure-of-speech precision; the negative side covers both misused/garbled idioms and over-reliance on clich\'{e}d or trite expressions. &
        $+/-$ \\
         & \codelabel{A5} & Word choice: register \& contextual fit &
        Whether word choices match the context in aspects such as formality, period (e.g., historical vs contemporary), dialect/local speech, genre (e.g., sci-fi), speaker/situation (e.g., emotional context). &
        $+/-$ \\
         & \codelabel{A6} & Word choice: richness vs. blandness &
        Lexical interest of the vocabulary used, such as unusual, varied, atmospheric words and turns of phrase vs.\ generic, common, sterile, ``dumbed-down'' choices. Independent of B8 (a passage can be rich-and-concise or rich-and-wordy). &
        $+/-$ \\
         & \codelabel{A7} & Cultural adaptation &
        Translator's localization decisions in both directions: localizing/explicating foreign concepts for the target reader (e.g., glossing terms, Americanizing place names) vs.\ preserving foreignness (leaving names/terms in source language, retaining cultural markers). &
        $+/-$ \\
         & \codelabel{A8} & Sentence structure &
        Sentence length, complexity, word order, run-ons, fragments, choppiness, and variety vs.\ monotony across sentences. &
        $+/-$ \\
         & \codelabel{A9} & Consistency &
        Whether the translation remains the same choices across the text in terms of vocabulary, voice, register, etc.\ (e.g., whether character names are translated consistently). &
        $+/-$ \\
         & \codelabel{A10} & Formatting \& typography &
        The text's visual and typographic conventions, such as the typesetter-and-stylistic-choice layer that sits between grammar and content. It includes paragraph structure (breaks, dividers), dialogue formatting (quotation vs em-dashes). &
        $+/-$ \\
         & \codelabel{A11} & Repetition or redundancy &
        Overuse of the same words, phrases, structures whether variation or compression would be better. This includes unnecessary lexical repetition, redundant content, etc. &
        $-$ \\
         & \codelabel{A12} & Untranslated content &
        Source material (words or phrases) that appear untranslated in the target text and are perceived by the reader as translation failures not deliberate choices to retain the source word for culture preservation. &
        $-$ \\
        \midrule
        \parbox[t]{0.12\textwidth}{\codelabel{B}.\\Narrative-level features} & \codelabel{B1} & Dialogue: naturalness &
        Whether the dialogue sounds like how people actually speak in the specific situation, context, emotional moment. It covers cases such as word plausability in speech (of this character), belivable phrasing etc. &
        $+/-$ \\
         & \codelabel{B2} & Character voice \& potrayal &
        Whether characters come through as distinct, identifiable, and well-matched to who they are. B2 evaluates the voice of characters and narrators: both in dialogue and in prose attributed to a character's perspective. character distinctness, attribution \& character fit &
        $+/-$ \\
         & \codelabel{B3} & Description / vividness / visualizability &
        Whether the reader can picture the scene and whether word choices conjure clear mental images even when both options are technically correct; includes the craft distinction of showing vs.\ telling (scene-rendering that creates imagery vs.\ bare informational delivery). &
        $+/-$ \\
         & \codelabel{B4} & Figurative language &
        Use and effectiveness of metaphors, similes, imagery, and atmospheric or evocative phrasings; fires positively for evocative figures and negatively for misfiring ones that don't match the referent. &
        $+/-$ \\
         & \codelabel{B5} & Emotional conveyance &
        Effectiveness of conveying the character’s or scene’s emotions to the reader (distinct from C4: B5 is about how well the reader can feel the story’s emotional elements, while C4 is about whether the text itself feels alive or soulful). &
        $+/-$ \\
         & \codelabel{B6} & Narrative organization &
        Story structure: scene transitions, paragraph / section flow, balance of exposition vs.\ action, story cohesion / motif recall (themes and descriptions that call back to previous text in the story). &
        $+/-$ \\
         & \codelabel{B7} & Narration \& POV effects &
        Narrator stance: interior monologue, free indirect discourse (blending character’s thoughts into narration), narrator’s distance from the action, narrator addressing the reader directly; point-of-view handling like POV shifts. &
        $+/-$ \\
         & \codelabel{B8} & Pacing / information delivery / wordiness &
        Narrative-level pacing: info-dumps, lengthy descriptions that interrupt action; phrase-level wordiness (e.g.\ one word vs.\ whole phrase to describe the same thing). Independent of A6. &
        $+/-$ \\
        \midrule
        \parbox[t]{0.12\textwidth}{\codelabel{C}.\\Reader experience} & \codelabel{C1} & Comprehension / ease of following &
        Whether the reader could understand what was happening: local sentence-level understanding and global scene/plot-level comprehension (these can fire independently). &
        $+/-$ \\
         & \codelabel{C2} & Smoothness / reading effort &
        Whether the prose flowed or the reader kept getting stuck; i.e.\ how much friction was in the reading experience, separate from whether they understood it. &
        $+/-$ \\
         & \codelabel{C3} & Engagement / immersion &
        Whether the reader was pulled into the story: being in the world, losing awareness of reading, focused attention on the scene. &
        $+/-$ \\
         & \codelabel{C4} & Humanness / soul &
        Holistic judgement on whether the prose feels alive and human vs.\ robotic or mechanical (distinct from A3: individual word choices or sentences can feel natural, while the whole story can overall feel soulless). &
        $+/-$ \\
         & \codelabel{C5} & Enjoyment / overall positive affect &
        General reader satisfaction with the text: ``I liked it'', ``pleasant to read'', ``a pleasure'', ``I enjoyed both''; broader affective stance not related to a specific dimension, can be used as a catch-all for positive sentiment when no specific details given. &
        $+/-$ \\
        \midrule
        \parbox[t]{0.12\textwidth}{\codelabel{D}.\\Meta-translation} & \codelabel{D1} & Doesn't read as translated &
        Holistic judgment on whether the prose reads as if originally written in English vs.\ feels like it’s been translated. &
        $+/-$ \\
         & \codelabel{D2} & Adaptation vs. literalness &
        Whether the translator adapted to English idioms and general audience expectations (positive) vs.\ translated word-for-word from the source language (negative); can fire with D4 when literalness is interpreted as a cue for MT. &
        $+/-$ \\
         & \codelabel{D3} & Faithfulness to original &
        How well the text preserves the source meaning, voice, emotion, structure, subtle implications, connotations, and specific facts; usually perceived at the chunk-level, and often inferred by comparing small detail inconsistencies between two presented translations. &
        $+/-$ \\
        & \codelabel{D4} & MT/AI tell named &
        Reader explicitly articulates a specific cue that triggers the MT/AI impression: e.g., em-dash overuse, bullet-point styling, ``would'' overuse, US English defaults, anthropomorphizing concepts, and other details that are perceived as tells for AI-generated text. &
        $+/-$ \\
\bottomrule
\end{tabular}
\end{adjustbox}
\caption{Annotation scheme for participant comments on Single Q5, Q6, Compar. Q4, and Chunk Justif. Only positive codes are applied to comments for Single Q5, and only negative codes are applied for Single Q6. For Compar. Q4 and Chunk Justif., positive codes are applied to match the comments' positive aspects for the preferred translation, and negative codes applied to match the negative aspects of the non-preferred translation.}
\label{tab:annotation-scheme-q1-q2-q3-q5}
\end{table*}

\begin{table*}[p]
\centering
\scriptsize
\setlength{\tabcolsep}{4pt}
\renewcommand{\arraystretch}{1.2}
\begin{adjustbox}{max width=\textwidth, max totalheight=.88\textheight, keepaspectratio}
\begin{tabular}{@{}>{\raggedright\arraybackslash}p{0.06\textwidth} >{\raggedright\arraybackslash}p{0.14\textwidth} >{\raggedright\arraybackslash}p{0.50\textwidth} >{\raggedright\arraybackslash}p{0.24\textwidth} >{\raggedright\arraybackslash}p{0\textwidth}@{}}
        \toprule
        \textbf{Code} & \textbf{Aspect} & \textbf{Definition (AI-pointing)} & \textbf{Human-pointing examples} \\
        \midrule
        \codelabel{M1} & Literalism / source-bleed &
        Translation feels word-for-word or dictionary-like; source-language structure, idiom, or convention bleeds through; the method feels direct and unedited rather than adapted for English. &
        translated in a way that was corrected to make sense for English &
         \\[5pt]
        \codelabel{M2} & Word choice &
        A specific word or short phrase reads as off; wrong, stilted, archaic, jargon, bland, generic, or conversely vivid and colloquial. The reader points at the word itself. &
        in a frenzy' more accurate than 'with a fever' &
         \\[5pt]
        \codelabel{M3} & Sentence flow, structure \& grammar &
        How sentences are shaped and how they connect: choppy, truncated, run-on, abrupt, copy-paste structure, sentences that do not link, and grammatical mechanics such as tense inconsistency, verb agreement errors, modal misuse, or overuse of grammatical forms. &
        some things flow more smoothly in T1; more variation in sentence lengths &
         \\[5pt]
        \codelabel{M4} & Formatting \& surface artifacts &
        Visible page-level mechanics not folk-theory framed: punctuation, scene dividers, quotation marks, dialogue formatting, and encoding glitches. Em-dashes as an ``AI sign'' are coded as M10. &
        slightly better formatting with proper nouns &
         \\[5pt]
        \codelabel{M5} & Wrong tone or mixed register &
        Whole passage is in a wrong tone for a novel, such as manual, textbook, technical, or Wikipedia-mode prose, or register switches and mixes inside the passage, such as US/UK slang together, source-culture and target-culture mixing, or formal-informal swings. &
        older style consistent with fantasy &
         \\[5pt]
        \codelabel{M6} & Wordiness / over-explaining &
        Text uses more words than needed: padding, extra words, info-dumping, or lack of economy. Different from M5; wordiness can occur in any mode. &
        economical; tight prose &
         \\[5pt]
        \codelabel{M7} & Voice \& emotional life &
        Whether prose is alive, e.g., engages, immerses, evokes mood, conveys emotion, gives characters distinct voices. Fires on flatness, mechanical feel, characters sounding alike, failure to evoke scene, inability to picture/feel. Immersion lives here. &
        more natural and conveys more emotion; made me feel like a sassy idol &
         \\[5pt]
        \codelabel{M8} & Meaning errors &
        Text gets meaning, logic, embodied physics, or factual content wrong, e.g., impossible images, lost nuance, dumbed-down content. Error is in the text, not the reader's parsing. &
        T2 ``breathed through his mouth'' more logical; T2 treats speaking underwater as hypothetical rather than given &
         \\[5pt]
        \codelabel{M9} & Comprehension difficulty &
        Reader could not follow, had to reread, needed the other version, or was confused by a specific passage. About reader experience, not text defect. &
        T2 much easier to understand; version of the sentence clearer in T2 &
         \\[5pt]
        \codelabel{M10} & Folk theory invoked &
        Reader explicitly references what AI does, such as a named stereotype, prior experience, or capability claim. Triggers on language like ``classic sign of AI,'' ``AI tends to,'' ``I'd expect from AI,'' or ``beyond what AI is capable of.'' &
        Human translators manage fluidity AI does not match; would not think AI capable of that leap &
         \\[5pt]
\bottomrule
\end{tabular}
\end{adjustbox}
\caption{Annotation scheme for participant comments on Compar. Q7 (justifying decision for which translation is perceived as machine translation).}
\label{tab:annotation-scheme-q4}
\end{table*}

\label{subsec:app-anno-schema}
The coded comments are released in the  \href{https://github.com/Yves575/lait/blob/main/docs/COMMENT_CODING_SCHEMA_GUIDELINE.md}{\path{repository}}.

The category labels for translation quality and the reasons for judging a translation as AI-generated are described in \autoref{tab:annotation-scheme-q1-q2-q3-q5} and \autoref{tab:annotation-scheme-q4} respectively.
We present comment examples in \autoref{tab:example-contextual-judgment-examples}.

\paragraph{Comment analysis.}\label{app:preference-comment-evidence} \autoref{tab:q3-pos-ht} shows the top positive labels assigned to comments by readers who preferred HT in the immersive reading comparative questionnaire to indicate why they preferred it. \autoref{tab:q3-pos-mt} shows the top positive labels for comments from readers who preferred MT in that questionnaire. \autoref{tab:q5-pos-ht} shows the top positive labels assigned to comments by readers who preferred HT in the close reading per-chunk questionnaire to indicate why they preferred it. \autoref{tab:q5-pos-mt} shows the top positive labels for comments from readers who preferred MT in that questionnaire. 
\autoref{tab:quality-reason-preference-examples} summarizes the top preference labels and representative comments.

\begin{table*}[t]
\centering
\small
\begin{tabularx}{\textwidth}{r l r X}
\toprule
Rank & POSITIVE label assigned to MT & Count & Example \\
\midrule
1 & A3. Word choice: naturalness & 35 &
``Overall I preferred the wording used in MT. HT felt slightly simple at times, e.g., `being beaten with sticks' felt like over-explaining.'' \\

2 & A5. Word choice: register \& contextual fit & 29 &
``I slightly preferred the more formal/older English style of MT but both translations seemed to sufficiently convey the message.'' \\

3 & C2. Smoothness / reading effort & 26 &
``MT was a better translation as it was generally easier to read with a better natural flow.'' \\

4 & A6. Word choice: richness vs. blandness & 23 &
``MT uses more interesting, unusual words like `pellet of paper' instead of `ball of paper' and `loathsome' instead of `hideous.''' \\

5 & B3. Description / vividness / visualizability & 22 &
``I only picked MT as my preference because the descriptions, like of their home, gave me a fuller picture of the scene.'' \\

6 & C1. Comprehension / ease of following & 17 &
``HT is all mixed up in the narrative and it is not as logically rolled out as in MT. There is a significantly better flow in MT and it makes it more understandable and easier to follow.'' \\

7 & B8. Pacing / information delivery / wordiness & 17 &
``MT has a tighter feel that comes across as more of a conversation than HT does.'' \\

8 & B2. Character voice \& portrayal & 16 &
``I overall liked the wording of MT better because it sounded to me more like how the character of Sarah, who is the wife of an author, might speak.'' \\

9 & A2. Word choice: clarity & 16 &
``I thought HT was referring to a literal `band' of boys because of the instrument talk, so I prefer MT because the word `gang' was much clearer.'' \\

10 & D3. Faithfulness to original & 13 &
``MT has a marginally better flow and leans more faithfully on the Japanese, which results in a slightly awkward English reading experience, but it is not by any means bad.'' \\
\bottomrule
\end{tabularx}
\caption{Top Q5 POSITIVE labels assigned to MT in the chunk-level comparison task. Counts are response-level label presences.}
\label{tab:q5-pos-mt}
\end{table*}

\begin{table*}[t]
\centering
\small
\begin{tabularx}{\textwidth}{r l r X}
\toprule
Rank & POSITIVE label assigned to HT & Count & Example \\
\midrule
1 & C2. Smoothness / reading effort & 92 &
``HT simplified a lot of overcomplicated sentences we see in MT, which makes text in HT flow much better. Sentences in MT are too long and overly complicated.'' \\

2 & A2. Word choice: clarity & 87 &
``HT is more succinct, more precise, and more idiomatic. MT is unnecessarily wordy, e.g., `frightened her terribly' instead of `terrified her.''' \\

3 & A3. Word choice: naturalness & 80 &
``HT has more natural phrasing while MT sounds more like a literal translation. Better word choice makes HT easier to read and makes the text flow better when reading.'' \\

4 & C1. Comprehension / ease of following & 64 &
``I found myself having to reread sentences in MT to follow what it was saying. It was just far easier to understand in HT.'' \\

5 & A8. Sentence structure & 39 &
``HT had a better choice of splitting it up into smaller sentences which make it easier to read.'' \\

6 & B3. Description / vividness / visualizability & 39 &
``HT has a much clearer narrative picture and reads beautifully. The descriptive passages hit very well and it is just very enjoyable to read.'' \\

7 & B8. Pacing / information delivery / wordiness & 31 &
``HT has a better narrative and conversational flow, as well as tighter descriptions and a better arrangement of the action.'' \\

8 & A5. Word choice: register \& contextual fit & 26 &
``Some sentences in MT, while technically correct, are not how a person would say it in English. `We'll be happy to comply,' from HT is more fitting.'' \\

9 & A6. Word choice: richness vs. blandness & 24 &
``HT has more unusual and interesting word choice, such as `omnipresent water' instead of just `all that water' and `a sorrowful note appeared in his voice' rather than `sadness entered his voice.''' \\

10 & A1. Grammar \& spelling & 22 &
``HT has a much more natural use of grammar and punctuation. I struggled to understand how these two sentences are connected when I initially read them in MT.'' \\
\bottomrule
\end{tabularx}
\caption{Top Q5 POSITIVE labels assigned to HT in the chunk-level comparison task. Counts are response-level label presences.}
\label{tab:q5-pos-ht}
\end{table*}

\begin{table*}[t]
\centering
\small
\begin{tabularx}{\textwidth}{r l r X}
\toprule
Rank & POSITIVE label assigned to MT & Count & Example \\
\midrule
1 & C2. Smoothness / reading effort & 5 &
``MT has a better flow and word choice overall compared to HT, which had me rereading sentences to grasp their meaning more than once.'' \\

2 & C3. Engagement / immersion & 5 &
``The language in MT is much more natural than HT and is clearer to understand. MT pulls the reader in with more vivid descriptions and is generally well-written. HT was boring to read and I felt my mind wandering.'' \\

3 & C5. Enjoyment / overall positive affect & 5 &
``I enjoyed both translations immensely and found them easy to read. I prefer MT's word choices because they come off as more intricate and varied to me. It is more formal than HT's phrasing in some parts, without feeling unnatural.'' \\

4 & A6. Word choice: richness vs. blandness & 4 &
``MT uses richer language with more unusual turns of phrase and wording. This creates a better atmosphere and more interest. HT uses regular language that seems somehow sterile and more common.'' \\

5 & C1. Comprehension / ease of following & 4 &
``MT made more of an effort to explain foreign words than HT, like `the Pentateuch' in HT is translated as `the Five Books of Moses' in MT. I have no idea if that is an accurate translation, but it was easier for me to understand.'' \\

6 & B2. Character voice \& portrayal & 3 &
``I felt like I was able to pick up the essence of the characters more in MT, for instance, the difference between Sarah thinking `He melts me, the idiot' and `I'm touched.' MT just gave a clearer picture of personalities and relationships with its words.'' \\

7 & A5. Word choice: register \& contextual fit & 3 &
``MT also made language choices, particularly the use of an older style of English, that are generally more consistent with this style of fantasy, e.g. [MT] `Do not take offence' vs. [HT] `Don't be cross'.'' \\

8 & A3. Word choice: naturalness & 3 &
``The language in MT is much more natural than HT and is clearer to understand. MT pulls the reader in with more vivid descriptions and is generally well-written.'' \\

9 & A2. Word choice: clarity & 3 &
``When reading HT the last sentence confused me initially, `But what if he didn't when he could see it,' while in MT it seemed much clearer: `But what did it matter that he did not know, when he saw.''' \\

10 & A7. Cultural adaptation & 2 &
``MT made more of an effort to explain foreign words than HT, like `the Pentateuch' in HT is translated as `the Five Books of Moses' in MT. I have no idea if that is an accurate translation, but it was easier for me to understand as someone with very little knowledge of Jewish teachings.'' \\
\bottomrule
\end{tabularx}
\caption{Top Q3 POSITIVE labels assigned to MT in the Part 1 comparison questionnaire. Counts are response-level label presences.}
\label{tab:q3-pos-mt}
\end{table*}

\begin{table*}[t]
\centering
\small
\begin{tabularx}{\textwidth}{r l r X}
\toprule
Rank & POSITIVE label assigned to HT & Count & Example \\
\midrule
1 & C2. Smoothness / reading effort & 15 &
``HT was much easier for me to read than MT. The wording flowed better and made it a seamless reading experience, whereas in MT I felt like I had to reread sentences now and then to make sure I was understanding right.'' \\

2 & C1. Comprehension / ease of following & 8 &
``HT was far easier to follow in terms of flow and word choice. MT had at times, bizarre punctuation choices and was very choppy. MT made it difficult to determine who was speaking.'' \\

3 & A3. Word choice: naturalness & 6 &
``The HT translation is simply more human. MT translation is overall proper with small errors, but there is no emotional impact on the reader. HT flows better, the sentences are more real for lack of a better word.'' \\

4 & A8. Sentence structure & 6 &
``The wording in both translations was good, but where HT surpassed MT was in sentence structure, avoiding too many long sentences, and formatting, dividers between scenes.'' \\

5 & A2. Word choice: clarity & 5 &
``From the start of the story, HT was a lot clearer to me and I think a lot of that has to do with punctuation and word choice. MT does not necessarily use obscure or uncommon words, but the translation does not feel quite right at times.'' \\

6 & B6. Narrative organization & 5 &
``HT flowed better. Text was broken up and summarized. The dialogue tags were more natural. I like the shorter sentences in the first translation.'' \\

7 & C3. Engagement / immersion & 4 &
``HT had no formatting issues that I noticed, for one, and also had a richness/uniqueness to it that pulled me in. The dialogue flowed well and, while complicated at times, is not unusual for an epic fantasy book of this nature.'' \\

8 & A6. Word choice: richness vs. blandness & 3 &
``The descriptions were more 3 dimensional in HT. They interacted with one another and recalled past phrasing. They drew pictures more vividly and expressed thought in a more cohesive manner. The flow had a more nuanced cadence and was less flat than MT.'' \\

9 & C4. Humanness / soul & 2 &
``HT stands out to me as clearly more human translated. Translations, but also literary fiction in particular, can get away with a bit of a weird/removed style of writing that made MT not seem strange to me, but HT was more natural.'' \\

10 & D1. Doesn't read as translated & 2 &
``HT also felt more like reading a story in English rather than a translated story. MT definitely read like someone or something translating word by word; it did not feel natural while reading.'' \\
\bottomrule
\end{tabularx}
\caption{Top Q3 POSITIVE labels assigned to HT in the Part 1 comparison questionnaire. Counts are response-level label presences.}
\label{tab:q3-pos-ht}
\end{table*}

\begin{figure*}[t]
    \centering
    \begin{subfigure}[t]{0.49\textwidth}
        \centering
        \includegraphics[
            width=\linewidth,
            height=0.34\textheight,
            keepaspectratio
        ]{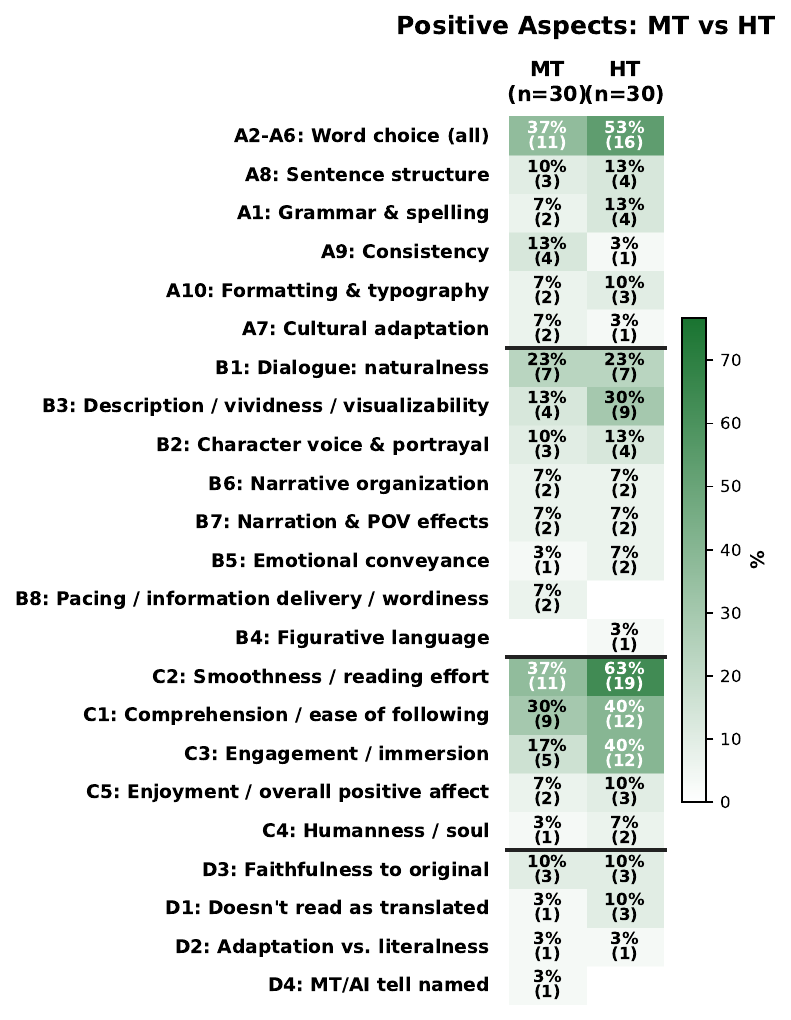}
        \caption{Positive aspects}
        \label{fig:stat-positive-aspects-mt-vs-ht}
    \end{subfigure}
    \hfill
    \begin{subfigure}[t]{0.49\textwidth}
        \centering
        \includegraphics[
            width=\linewidth,
            height=0.34\textheight,
            keepaspectratio
        ]{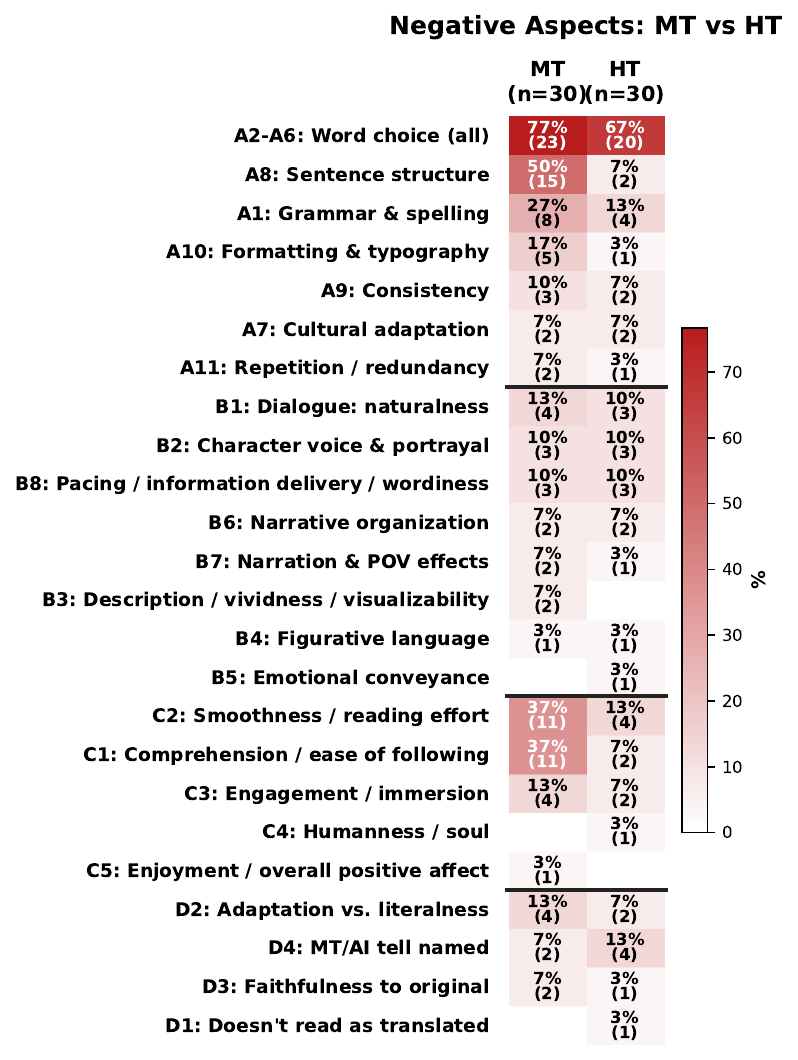}
        \caption{Negative aspects}
        \label{fig:stat-negative-aspects-mt-vs-ht}
    \end{subfigure}

    \caption{Positive and negative aspects mentioned by participants when comparing machine translation (MT) and human translation (HT).}
    \label{fig:stat-positive-negative-aspects-mt-vs-ht}
\end{figure*}

\begin{figure*}[!tbp]
    \centering
    \begin{subfigure}[t]{0.49\textwidth}
        \centering
        \includegraphics[
            width=\linewidth,
            height=0.34\textheight,
            keepaspectratio
        ]{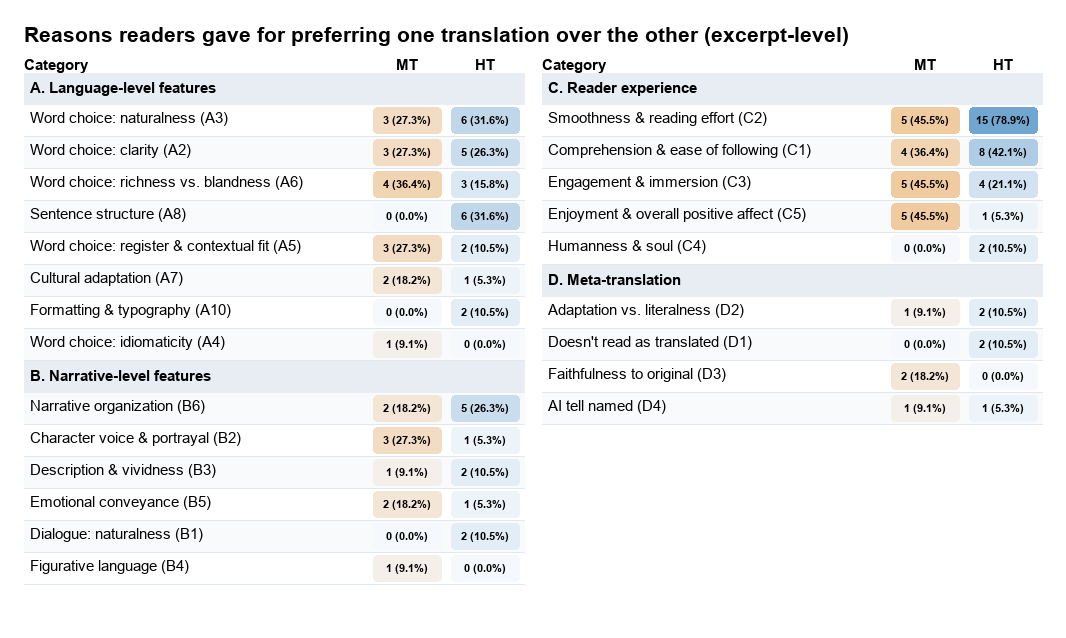}
        \caption{Excerpt-level preferences}
        \label{fig:pos-preference-reason-heatmap-excerpt}
    \end{subfigure}
    \hfill
    \begin{subfigure}[t]{0.49\textwidth}
        \centering
        \includegraphics[
            width=\linewidth,
            height=0.34\textheight,
            keepaspectratio
        ]{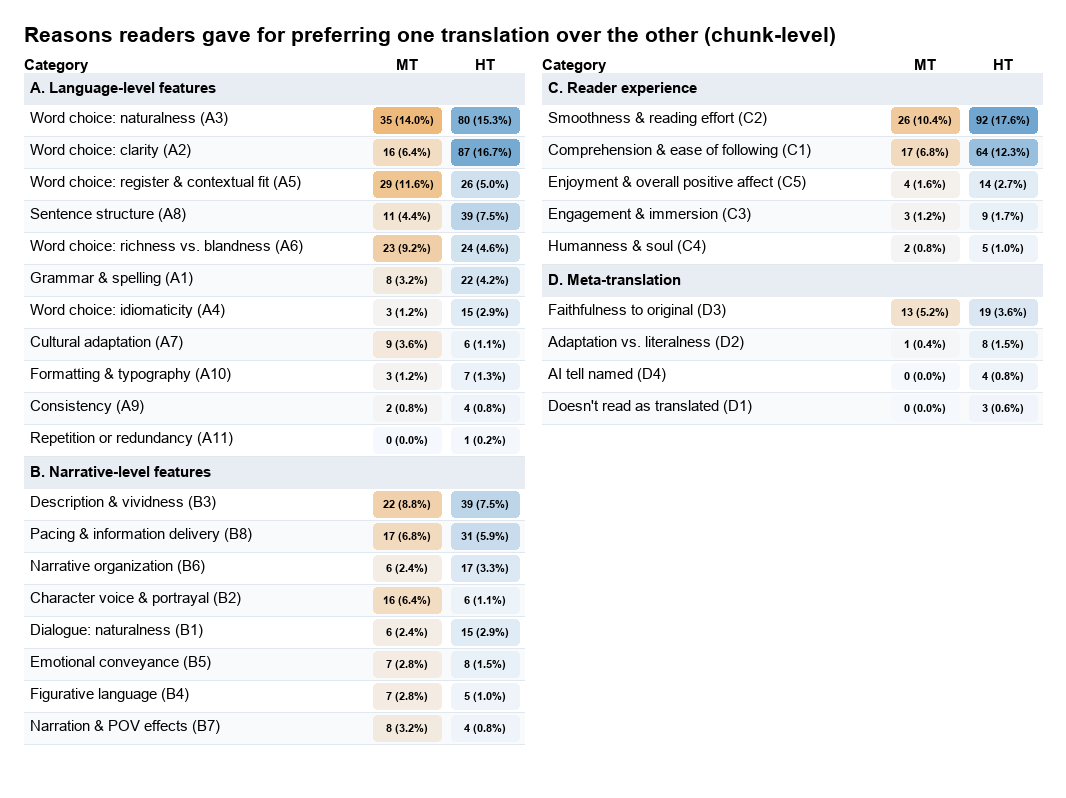}
        \caption{Chunk-level preferences}
        \label{fig:pos-preference-reason-heatmap-chunk}
    \end{subfigure}

    \caption{Positive reasons readers reported for preferring one translation over the other, separated by preferred system. Counts and percentages are calculated within each preferred-system group.}
    \label{fig:pos-preference-reason-heatmaps}
\end{figure*}

We further report tendencies mentioned in comments in \autoref{fig:stat-positive-negative-aspects-mt-vs-ht}, which summarizes positive and negative aspects readers mentioned for MT and HT in the comparison comments. Similarly, \autoref{fig:pos-preference-reason-heatmaps} shows the positive reasons readers gave for preferring HT or MT, separated by excerpt-level and chunk-level comparisons.

\begin{figure*}[t]
    \centering
    \begin{subfigure}[t]{0.90\textwidth}
        \centering
        \includegraphics[
            width=\linewidth,
            height=0.38\textheight,
            keepaspectratio
        ]{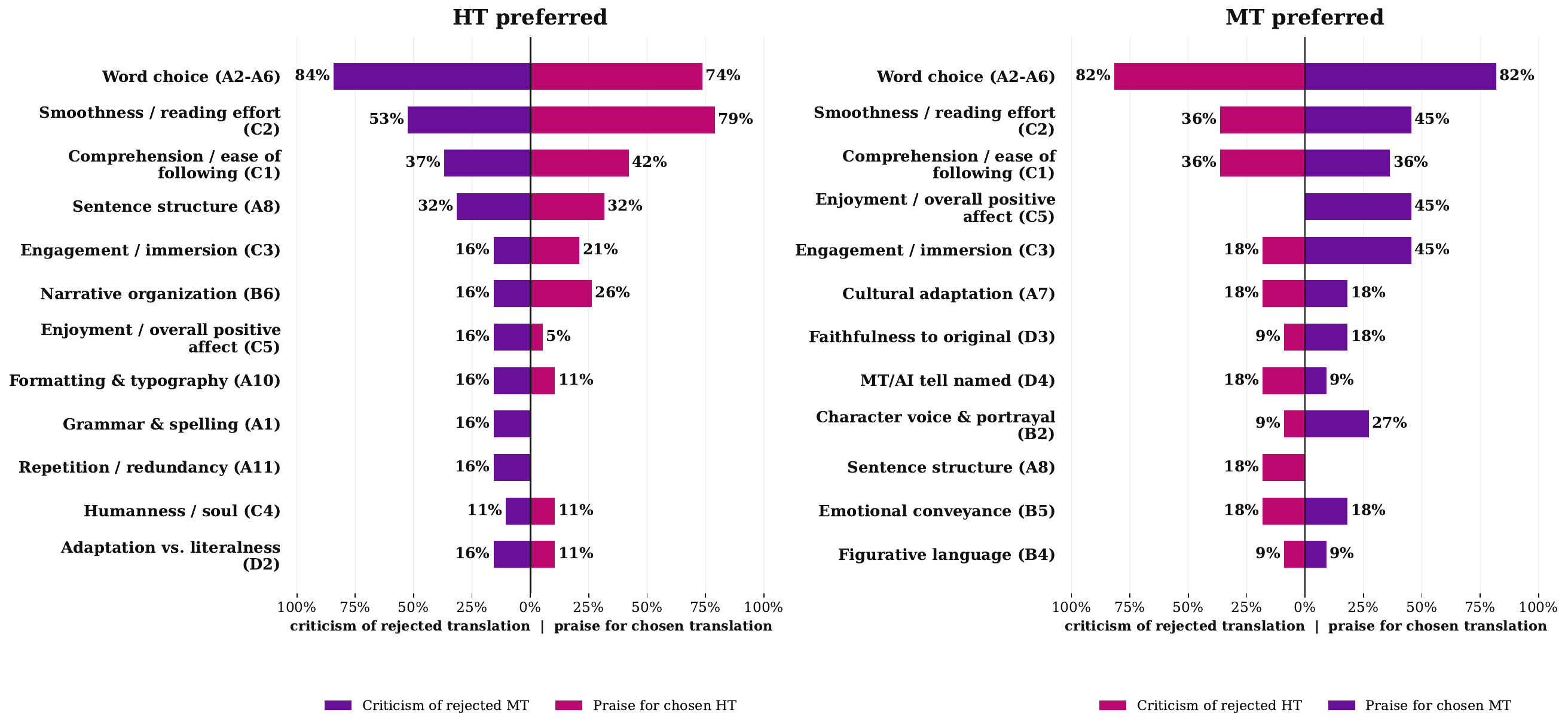}
        \caption{Excerpt-level preferences}
        \label{fig:stat-excerpt-preference-why-chosen-translation-won}
    \end{subfigure}

    \vspace{0.75em}

    \begin{subfigure}[t]{0.90\textwidth}
        \centering
        \includegraphics[
            width=\linewidth,
            height=0.38\textheight,
            keepaspectratio
        ]{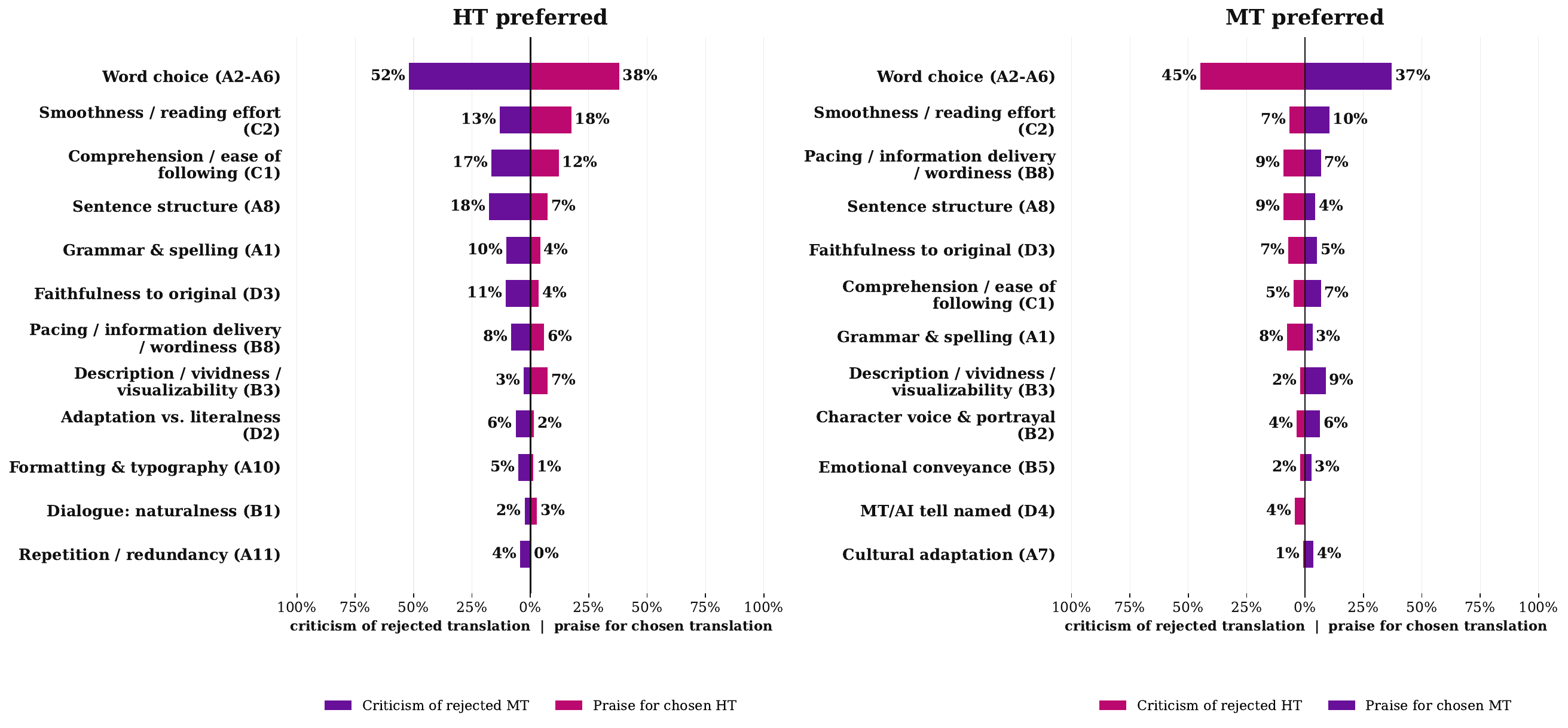}
        \caption{Chunk-level preferences}
        \label{fig:stat-chunk-preference-why-chosen-translation-won}
    \end{subfigure}

    \caption{Preference mechanisms explaining why the chosen translation was preferred. Diverging bars contrast attraction to the winning translation with rejection of the losing translation, separately for HT-preferred and MT-preferred cases.}
    \label{fig:stat-preference-why-chosen-translation-won}
\end{figure*}

As readers often mention not only positive things about their chosen translation but also negative things about the other candidate, we aim at visualizing these in \autoref{fig:stat-preference-why-chosen-translation-won}.

\begin{figure*}[t]
    \centering
    \includegraphics[
        width=0.82\textwidth,
        height=0.62\textheight,
        keepaspectratio
    ]{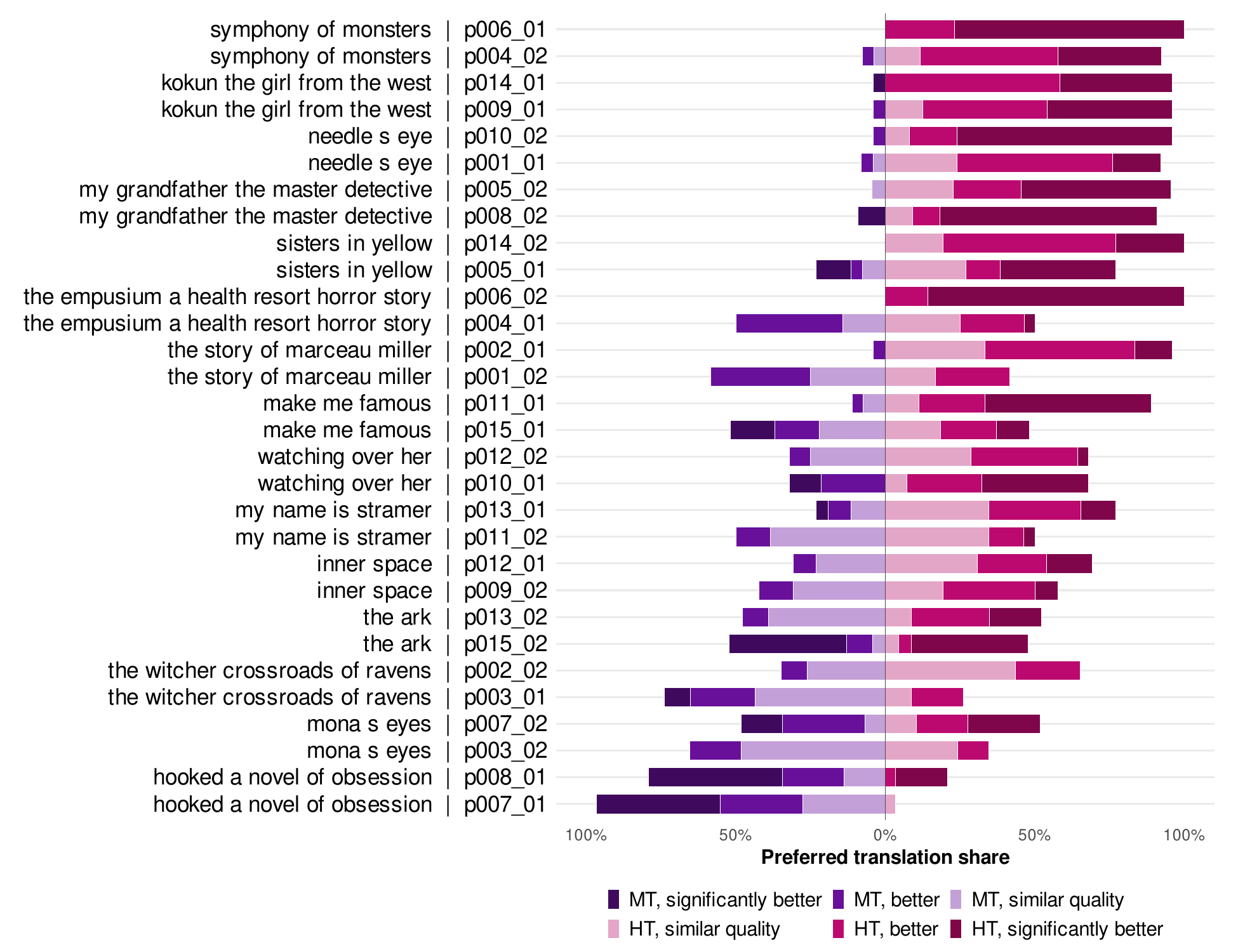}
    \caption{Preferred translation by excerpt and reader. Bars show the share of chunk-level choices favoring MT on the left and HT on the right; darker shades indicate greater preference strength.}
    \label{fig:stat-book_preference_diverging}
\end{figure*}

Finally, we present a book-level view of AI guesses per language (\autoref{fig:stat-book_ai_identification_diverging}), while \autoref{fig:stat-book_preference_diverging} shows book-level HT/MT preference direction and strength for each participant.

\begin{figure*}[t]
    \centering
    \includegraphics[width=0.95\textwidth]{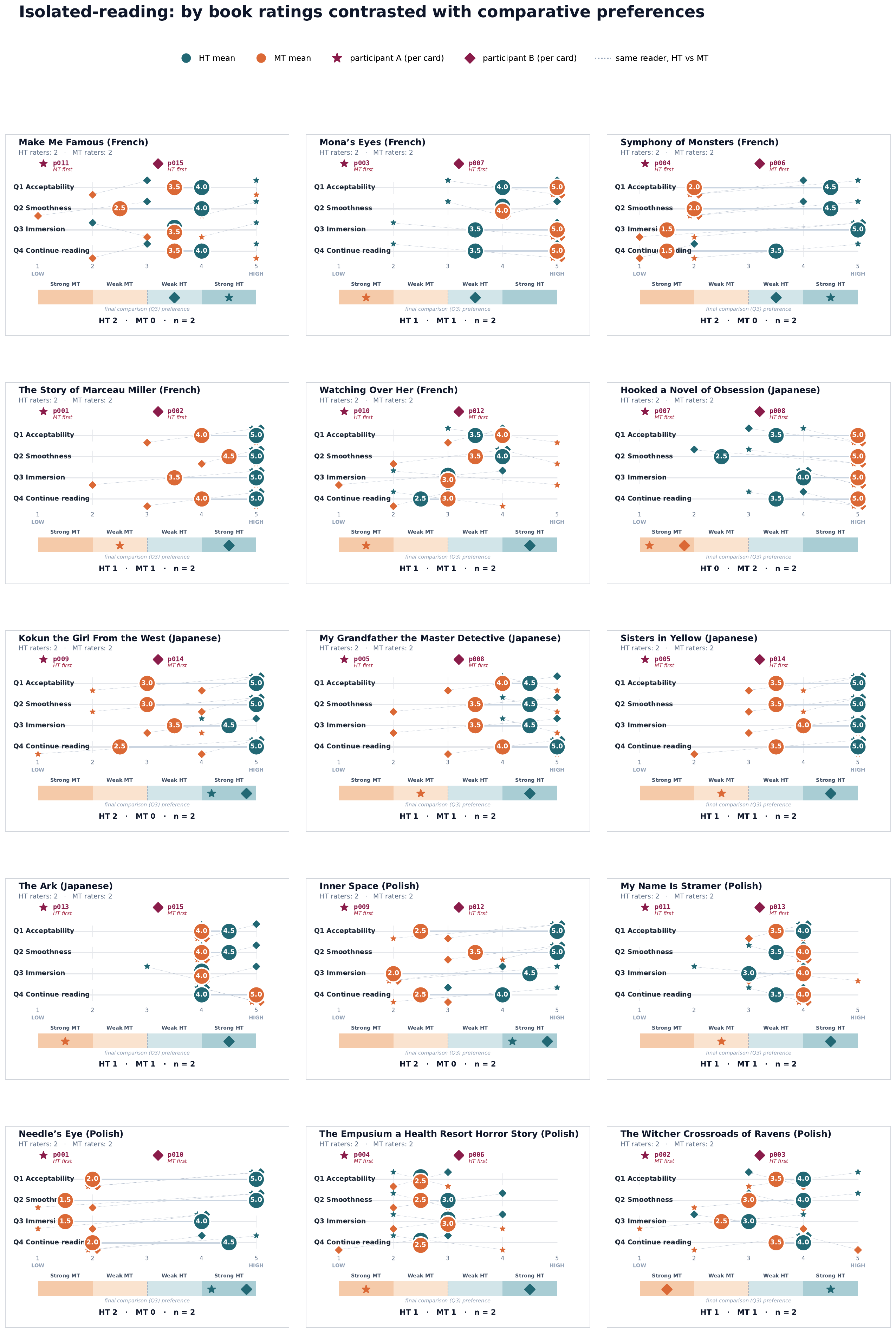}
    \caption{Per-book immersive-reading ratings for HT and MT. The figure breaks down participant ratings by book, making cross-book variation visible alongside the aggregate trends reported in the main paper.}
    \label{fig:by-book-individual-res}
\end{figure*}

\begin{figure*}[t]
    \centering
    \includegraphics[width=0.90\textwidth]{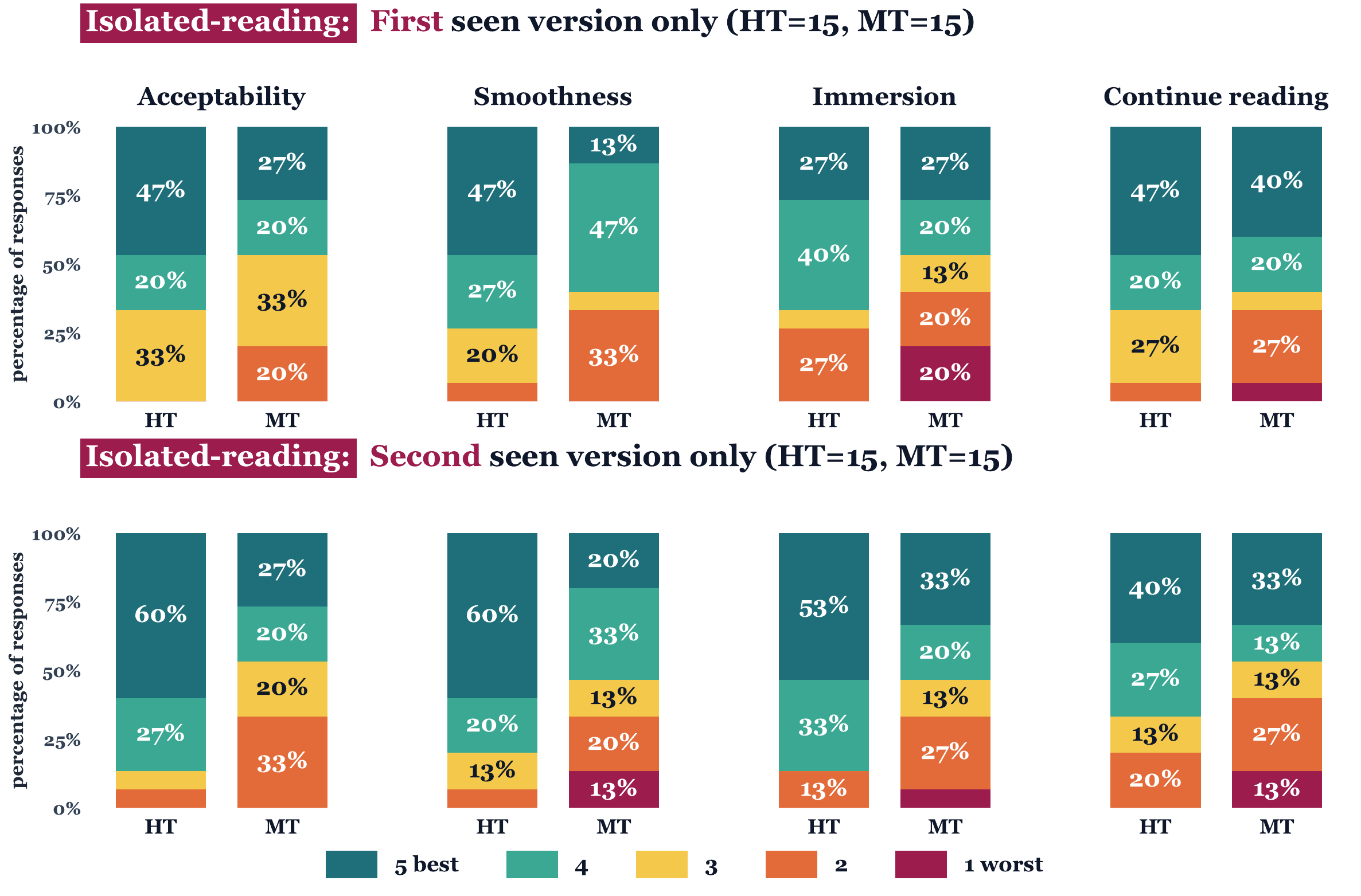}
    \caption{Immersive-reading ratings by presentation order. The figure compares ratings assigned after the first and second single readings, grouped by whether the human translation (HT) or machine translation (MT) was shown first.}
    \label{fig:single-reading-ratings-by-seen-order}
\end{figure*}

\paragraph{Single-reading ratings.} \autoref{fig:single-reading-ratings-by-seen-order} shows single-reading ratings for \texttt{acceptability}, \texttt{smoothness}, \texttt{immersiveness}, and \texttt{would continue} (reading), divided by the presentation order (i.e., ratings for what the readers saw first vs. second). We further report these ratings by individual preferences, indicating both the translation order and the readers' final preference for MT vs. HT in the excerpt-level evaluation (\autoref{fig:by-book-individual-res}).

\begin{figure*}[t]
    \centering
    \begin{subfigure}[t]{0.49\textwidth}
        \centering
        \includegraphics[height=0.15\textheight,keepaspectratio]{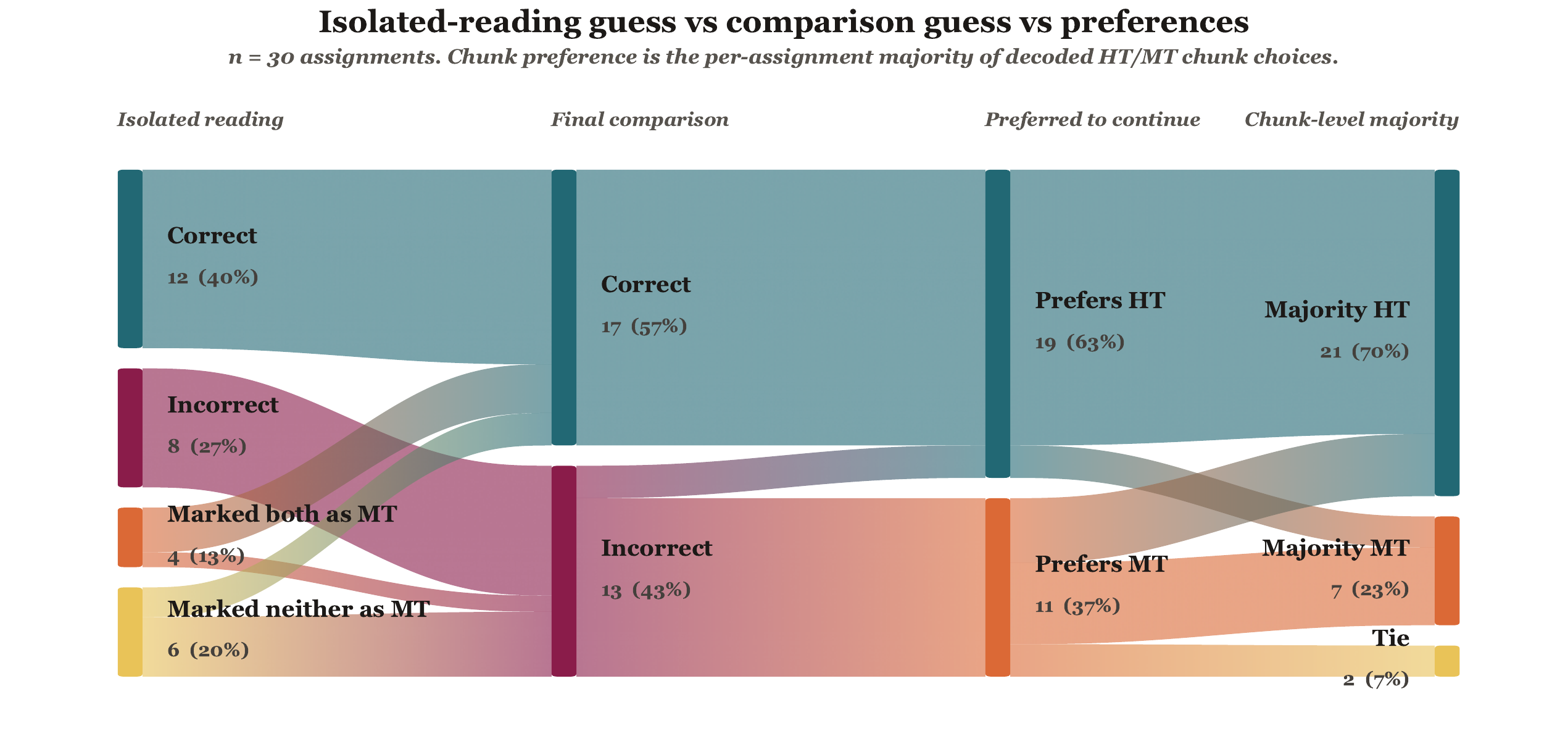}
        \caption{Guess correctness}
        \label{fig:sankey-correctness-guess}
    \end{subfigure}
    \hfill
    \begin{subfigure}[t]{0.49\textwidth}
        \centering
        \includegraphics[height=0.15\textheight,keepaspectratio]{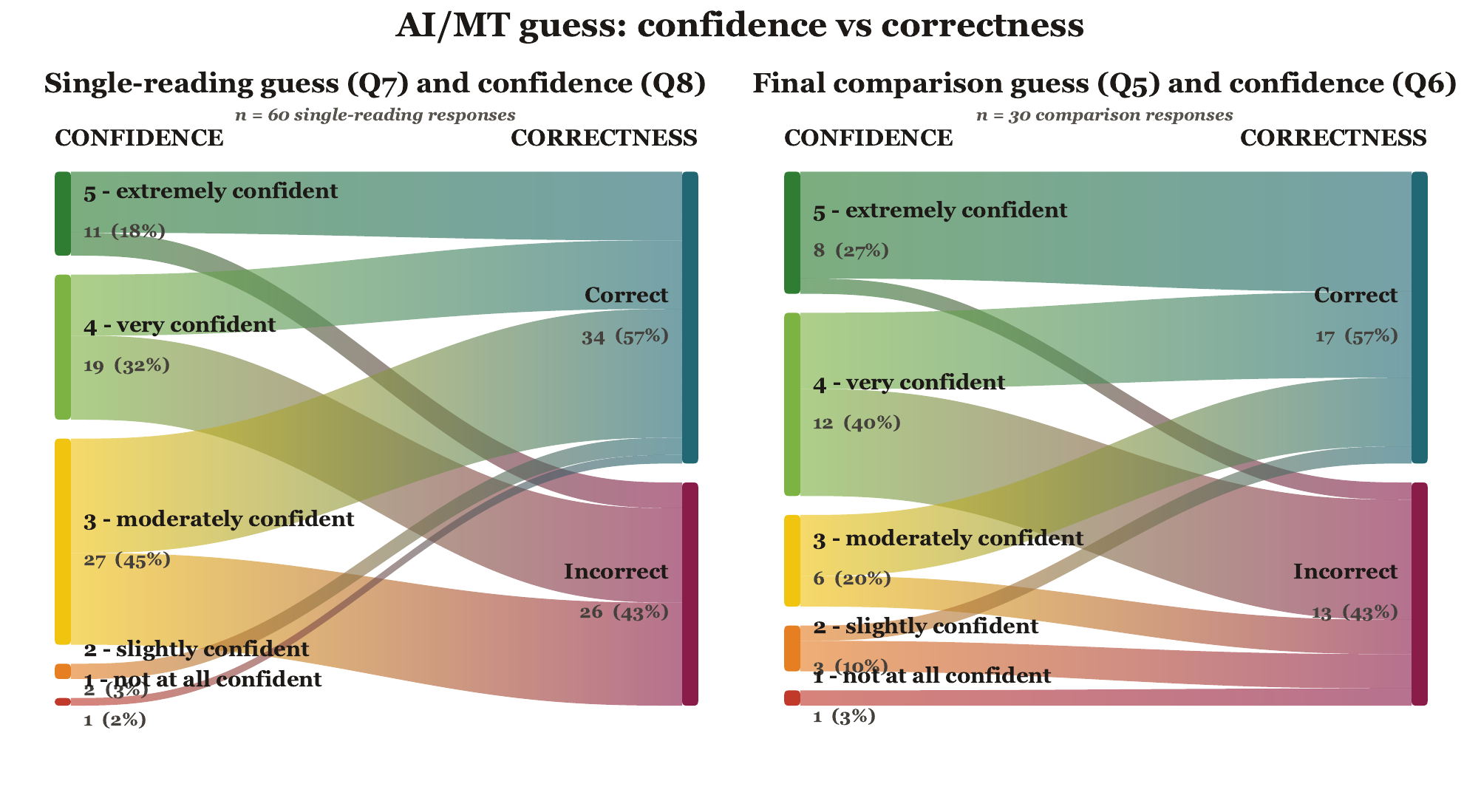}
        \caption{Guess confidence}
        \label{fig:sankey-conf-vs-corr-guess}
    \end{subfigure}
    \caption{Translation-origin guess flows after the comparison task. The first panel summarizes whether guesses were correct for HT and MT; the second panel shows reported confidence across those identification decisions.}
    \label{fig:sankey-guess-flows}
\end{figure*}

\paragraph{AI-detection guesses.} For AI-detection questions, we compare readers' single-reading guesses with their final comparison guesses. 
\autoref{fig:cues-ai} shows the most frequent cues readers mentioned when identifying a translation as machine-translated, separated by correct and incorrect guesses. 
Additional descriptive flows for guess correctness and confidence are shown in \autoref{fig:sankey-correctness-guess} and \autoref{fig:sankey-conf-vs-corr-guess}.

\paragraph{Readers who often use AI were NOT better at detecting AI translation.} Since we collect readers AI experience we check whether the more experienced readers were better at detecting AI translation. Overall, we do not see any apparent trends or correlation  between the AI experience and detection accuracy (\autoref{tab:ai-use-mt-identification}).

\paragraph{Good and poor span-level highlights.} In the \textit{close reading}, participants highlight spans they perceive as well and poorly worded. 
Poor evidence was more concentrated in MT chunks: 41.7\% had at least 100 poor-highlighted words per 1K words, compared with 11.9\% of HT chunks (\autoref{fig:span-spiky-poor-chunks}).

\paragraph{Post-study interviews.} Participants were asked to answer several questions about their opinions on LLM translations before and after the study. The questions can be found in \autoref{tab:post-study_questionnaire}. 
We report readers' post-study responses across four tables: \autoref{tab:post-study-responses-1}, \autoref{tab:post-study-responses-2}, \autoref{tab:post-study-responses-3} and \autoref{tab:post-study-responses-4}. 
The tables are split by participant group for readability.

Participants were initially skeptical about the quality of LLM translation: some, who have previously edited AI translations, recall them being of poor quality, and thus adjust their expectations accordingly, while other non-experts did not trust LLMs to produce a good narrative from anecdotal experience. However, most participants, even those who correctly discriminated HT from MT, were pleasantly surprised by MT's quality, despite their general preference towards HT, one participant even preferred MT over HT despite correctly identifying them! Furthermore, several participants noted that the LLM translations did not contain the usual cues that they typically associate with LLM translations and writing: one participant states that they would \textit{``have no inkling of it being translated by AI''} if it was surrounded by human-written books in their personal book collection. This view is not universal among participants, however. Some state that despite MT texts being semantically accurate, which is the goal of a general-purpose translation system, they write too literally, frequently miss subtle context cues in the story, mistranslate idioms or other culturally-sensitive phrases, and lack the stylistic and narrative requirements of fiction, \textit{``[MT] tends to flatten the author's voice.''} Conversely, many participants were disappointed in the quality of HT excerpts of some of the books, either due to mistakes that should have been caught during editing or overall poor quality. The participants' general consensus was that agentic LLM translation systems are not ready for standalone publication, but may be integrated to generate first drafts to be further edited by expert human translators.

\begin{table}[H]
    \centering
    \scriptsize
    \setlength{\tabcolsep}{3pt}
    \renewcommand{\arraystretch}{1.05}
    \begin{tabularx}{\columnwidth}{@{}>{\raggedright\arraybackslash}p{0.08\columnwidth}X@{}}
        \toprule
        \textbf{\#} & \textbf{Question} \\
        \midrule
        1 &
        What were your expectations for the quality of AI translation before starting this experiment?
        \\
        \midrule
        2 &
        You \textbf{\{correctly/incorrectly\}} identified the AI-translated version of Book 1, and \textbf{\{correctly/incorrectly\}} identified the AI-translated version of Book 2. Are you surprised by your accuracy?
        \\
        \midrule
        3 &
        What do you think about the quality of AI translation now, after reading the translation?
        \\
        \midrule
        4 &
        Any other comments?
        \\
        \bottomrule
    \end{tabularx}
    \caption{Post-study questionnaire. All responses are free text.}
    \label{tab:post-study_questionnaire}
\end{table}

\begin{table*}[p]
\centering
\footnotesize
\begin{tabularx}{\textwidth}{@{}l Y@{}}
\toprule
\multicolumn{2}{@{}l@{}}{\textbf{P1}\quad Identified Book 1: yes\quad Identified Book 2: no\quad Expected worse: yes}\\
\midrule
Q1 & I expected the AI translations to be more obviously written by AI. I was impressed even with the translations I didn't prefer. \\
\addlinespace[2pt]
Q2 & I'm not that surprised actually, book 2 was by far harder to pick a favorite of. I had a really hard time even finding flaws, and when I did I'd find them more on a preference level and then both translations had some. \\
\addlinespace[2pt]
Q3 & I think it's so good, hence my mistaking it as human. I think it read really smoothly and if it was one of my books, I would have no inkling of it being translated by AI \\
\addlinespace[2pt]
Q4 & no other comments come to mind! Thanks for having it, it was so fun! \\
\addlinespace[6pt]
\multicolumn{2}{@{}l@{}}{\textbf{P2}\quad Identified Book 1: yes\quad Identified Book 2: yes\quad Expected worse: yes}\\
\midrule
Q1 & My expectations for the quality of AI translation before starting this experiment were that they would be a little confusing to read because they would be directly translated, which sometimes does not mean translated correctly. It was my impression that usually with different languages, the English translations were not exactly correct. So I figured there would be some awkward or incorrect wording. \\
\addlinespace[2pt]
Q2 & I am not surprised by my accuracy in correctly identifying the AI translations of both Book 1 and Book 2. \\
\addlinespace[2pt]
Q3 & My opinion of AI and its quality is the same as it was before, which is that it can be helpful but at the end of the day only a real human can get something to the way it needs to be. I don't think AI has the ability to understand when something doesn't sound right or use context to correct something. A real person, on the other hand, can take something translated poorly and rework it in order for it to make sense. \\
\addlinespace[2pt]
Q4 & N/A \\
\addlinespace[6pt]
\multicolumn{2}{@{}l@{}}{\textbf{P3}\quad Identified Book 1: no\quad Identified Book 2: no\quad Expected worse: yes}\\
\midrule
Q1 & I expected the writing quality to be a lot worse. I thought that the resulting text would be more in line with that of AI-written text (not translated, entirely AI-generated from prompts). \\
\addlinespace[2pt]
Q2 & Not particularly. I leaned towards one being AI over the other mainly because of quality differences. I do still believe that with a few copies of each, humans could learn to pick up the patterns and correctly detect which is AI-generated. \\
\addlinespace[2pt]
Q3 & It's substantially better than I expected. I definitely feel that it's one of the best applications of LLMs that I've seen. \\
\addlinespace[2pt]
Q4 & The quality of the human translations were quite a bit worse than I expected. It's almost like they didn't use an editor to review it afterwards. There were pretty clear issues that shouldn't have slipped through.\par\noindent
I think that a version that's objectively higher quality than both could be made by combining the better parts of each translation. \\
\addlinespace[6pt]
\multicolumn{2}{@{}l@{}}{\textbf{P4}\quad Identified Book 1: no\quad Identified Book 2: no\quad Expected worse: yes}\\
\midrule
Q1 & My expectations of the AI translations were low, based on the poor quality of many of the AI-generated books I edit and humanize. \\
\addlinespace[2pt]
Q2 & I am somewhat surprised because I was fairly sure. However, the quality was much better than what I'm used to, and the books didn't have many of the usual ``tells'' I often see. \\
\addlinespace[2pt]
Q3 & I think it has improved remarkably in a very short time. \\
\addlinespace[2pt]
Q4 & N/A \\
\bottomrule
\end{tabularx}
\caption{Post-study responses grouped by participant (part 1 of 4; participants P1--P4). Q1 asks about expectations for AI translation quality before the study; Q2 asks whether participants were surprised by their AI-identification accuracy; Q3 asks for their post-study view of AI translation quality; Q4 collects other comments.}
\label{tab:post-study-responses-1}
\end{table*}

\begin{table*}[p]
\centering
\footnotesize
\begin{tabularx}{\textwidth}{@{}l Y@{}}
\toprule
\multicolumn{2}{@{}l@{}}{\textbf{P5}\quad Identified Book 1: no\quad Identified Book 2: no\quad Expected worse: no}\\
\midrule
Q1 & I expected them to be good. I use ChatGPT multiple times a day, for many reasons. A lot of times it's to get answers for random things that pop in my head that I'm curious about, (that happens quite a bit, maybe more than most people, my brain is funny like that). But I also have been using it a ton the past few months, daily, with my resume as I'm job hunting. I've seen AI give amazing responses, but I've also seen it be incorrect and inconsistent. So while my expectations were high, I had a constant hesitation wondering how much faith I wanted to put into the machine. If that makes sense. \\
\addlinespace[2pt]
Q2 & I'm not surprised now that I've completed them. For both books, when it came to Part 2 and I was annotating them and comparing them, I knew I'd made the wrong choices. Reading them one at a time and a day apart made a huge, huge difference and that surprised me. \\
\addlinespace[2pt]
Q3 & I think it's inconsistent, for starters. I don't think it's safe to hand projects over to and be confident that you'll get a flawless result. I've often wondered about how a machine can replace a human in instances like this, when tone and even the smallest nuances, make the biggest differences. They're also so literal with the words, they can't grasp the context. I think AI translations make for a great rough draft. Maybe that's the best way to put it. \\
\addlinespace[2pt]
Q4 & I'd be curious to know if my annotations, the translations that I critiqued and disliked the most, were AI. When I got to that point is where I feel I realized what was what. Or maybe I just didn't care for how a person translated them, and was critiquing them! Oh gosh.. haha! \\
\addlinespace[6pt]
\multicolumn{2}{@{}l@{}}{\textbf{P6}\quad Identified Book 1: yes\quad Identified Book 2: yes\quad Expected worse: no}\\
\midrule
Q1 & I expected AI to produce a competent literal translation but to slip up with idiomatic language and polysemous words, and its translation of scenes that are supposed to evoke emotion. \\
\addlinespace[2pt]
Q2 & Not really surprised (but certainly gratified) because translating fiction is tricky even for humans. I think it was with Henning Mankell that I first started paying attention to the names of translators when one of the books in his Wallander series sounded ``off''. And he used top-notch human translators! \\
\addlinespace[2pt]
Q3 & On the whole, I assume that AI accurately reproduced the text from the original to English but it tends to flatten the author's voice. That is not a problem when all the reader wants is a functional translation of information. But that is not why we read fiction. \\
\addlinespace[2pt]
Q4 & An AI translation is still a lot like CliffsNotes: function without form. I doubt ``War and Peace'' will get an AI makeover any time soon! But I shouldn't be too cock-a-hoop. AI learns 24/7 and I will probably have to buy another hat to eat in the not too distant future. \\
\addlinespace[6pt]
\multicolumn{2}{@{}l@{}}{\textbf{P7}\quad Identified Book 1: no\quad Identified Book 2: no\quad Expected worse: yes}\\
\midrule
Q1 & My expectations were pretty low. I was paid to ``clean up'' a machine translation of a non-fiction book about the outbreak of war in Ukraine in 2022 for Ibidem press, and it was pretty clunky. The idioms were very off. This was much better. \\
\addlinespace[2pt]
Q2 & I am only a little surprised by my accuracy, now that both tasks are finished. \\
\addlinespace[2pt]
Q3 & I think that rather than demonstrating the quality of AI translation, this project demonstrated to me how challenging it is to judge the quality of any translation of a piece of literature from the opening pages. The Japanese novel really seemed ``off'' to me in how it was depicting the protagonist -- she seemed much too squirrely to be a successful businesswoman. I attributed this to an AI-translator, but having looked up the book, I now know she was supposed to seem squirrelly because she is a stalker on the verge of giving in to obsession. After reading those texts, I can definitely see the appeal of working with an AI-translator rather than a human-translator for an author looking to truly preserve their voice in a piece of fiction. The two tasks have not impacted how i feel about AI writing at all. \\
\addlinespace[2pt]
Q4 & This project inspired me to dust off the now 15-year-old translation of a Renaissance Latin text in the style of Sallust that was a core component of my doctoral project and which, because the text seems just so unappealing to English readers, I had not ever planned on returning to... and the LLMs' analysis and discussion with me of the text confirmed that the text just isn't translateable today ... when no one wants to read Gibbons-esque prose. Thank you for this opportunity. It was a fun task, and I am enjoying reading Hooked so much that I have sent copies to friends. \\
\bottomrule
\end{tabularx}
\caption{Post-study responses grouped by participant (part 2 of 4; participants P5--P7). Questions are as in Table~\ref{tab:post-study-responses-1}.}
\label{tab:post-study-responses-2}
\end{table*}

\begin{table*}[p]
\centering
\footnotesize
\begin{tabularx}{\textwidth}{@{}l Y@{}}
\toprule
\multicolumn{2}{@{}l@{}}{\textbf{P8}\quad Identified Book 1: no\quad Identified Book 2: yes\quad Expected worse: yes}\\
\midrule
Q1 & I had a realistic expectation of accurate translations in terms of word choice, but I wasn't sure about the sentence construction on a grammar level. I wasn't expecting a 1v1 direct translation, because that doesn't work from Japanese to English, but I thought there would be more\ldots{} mechanical issues. \\
\addlinespace[2pt]
Q2 & I am and I'm not. I think it comes down to the nature of the translation. Book 2 was much easier to detect AI on because the human-led translation was so tightly done on a narrative level. It flowed very well, with some variance in terms of paragraph structures that I attributed to the nature of the original language.\par\noindent
Book One was hard to figure out, but I chose the version I did because of the work I currently do. Most of my authors use AI to wholly or in part generate a manuscript which they submit to me for humanizing. These manuscripts have certain linguistic conventions (truncated sentences, over-reliance on mechanical or stage direction language, certain buzzwords to describe human emotions: the architecture of a face, etc.) that made me view the translations through that lens. I'm trained to spot short, choppy sentences and odd word choices as being AI-generated. \\
\addlinespace[2pt]
Q3 & I'm pleasantly surprised at the quality. It's not perfect, but it's definitely not going to stop me from reading. \\
\addlinespace[2pt]
Q4 & I think the style of the human translation played a large part in how I viewed the AI version. The translator who chose a more Western style felt more natural to me, and that made it easier to pick out an AI version. Book One's translator felt more faithful to the tone and style of the Japanese original, which read oddly to my English eyes.\par\noindent
This was an honestly fascinating look at a side of my industry that I didn't have before. I greatly appreciated this chance to be part of the project, and it's solidified my desire to write down my observations on how AI handles fiction. \\
\addlinespace[6pt]
\multicolumn{2}{@{}l@{}}{\textbf{P9}\quad Identified Book 1: yes\quad Identified Book 2: yes\quad Expected worse: yes}\\
\midrule
Q1 & I expected the translation to be a lot worse than it actually was (almost like how Google Translate translates a sentence - usually in a way no English speaker would say it). \\
\addlinespace[2pt]
Q2 & I'm not surprised by the accuracy - I wasn't completely confident, of course, but I didn't expect a human to miss the obvious issues in the AI translations (dialogue issues and scene break/double hyphen issues). \\
\addlinespace[2pt]
Q3 & The issues in the AI translations were more to do with style than actual major wording problems, which I think is amazing. \\
\addlinespace[2pt]
Q4 & I think the books were translated very well in all versions, a few easily-fixable issues aside. If AI could do the brunt work in translations, then a human can just review it and fix those small issues (or train the model to fix it itself in the future!) \\
\addlinespace[6pt]
\multicolumn{2}{@{}l@{}}{\textbf{P10}\quad Identified Book 1: no\quad Identified Book 2: yes\quad Expected worse: yes}\\
\midrule
Q1 & My expectations for AI translation were pretty low coming into this experiment. I have seen some pretty bad translations in the past. \\
\addlinespace[2pt]
Q2 & I'm surprised! But I do think Book 1's translation was a lot better than Book 2's. It was easier in Book 2 to differentiate between the two because of the amount of dialogue, in my opinion, and the way things were phrased. \\
\addlinespace[2pt]
Q3 & I think there is some merit to exploring AI-assisted translations, but I think they need human revisions to streamline the overall translation. \\
\addlinespace[2pt]
Q4 & Thanks for giving me the opportunity to participate in the study! \\
\addlinespace[6pt]
\multicolumn{2}{@{}l@{}}{\textbf{P11}\quad Identified Book 1: yes\quad Identified Book 2: no\quad Expected worse: yes}\\
\midrule
Q1 & I expected the AI translation to be obvious. I read a lot of AI generated work and it has very specific tells and often devolves into nonsense. \\
\addlinespace[2pt]
Q2 & I am not surprised. Book 1 felt very obvious when comparing the two translations. Book 2 was much harder. I'm not even sure which one I chose because the translations were of similar quality in my opinion (both felt kind of off). \\
\addlinespace[2pt]
Q3 & I was surprised by the quality of the translation in Book 1, it was much better than expected, but the human-generated translation was even better and provided context I don't think an LLM is capable of. I can see a good use case for it. I struggled a little with both translations of Book2 so it's hard to draw any conclusions there. \\
\addlinespace[2pt]
Q4 & As a lover of books and languages I really enjoyed doing this! Thank you for the experience. Oh! And I'll also add that if I had only read T1 from book 1, I would not have guessed it was AI translated. It was only obvious by comparison. I was very suspicious of there being a double blind on the second book \emph{[face with tears of joy emoji]} \\
\bottomrule
\end{tabularx}
\caption{Post-study responses grouped by participant (part 3 of 4; participants P8--P11). Questions are as in Table~\ref{tab:post-study-responses-1}.}
\label{tab:post-study-responses-3}
\end{table*}

\begin{table*}[p]
\centering
\footnotesize
\begin{tabularx}{\textwidth}{@{}l Y@{}}
\toprule
\multicolumn{2}{@{}l@{}}{\textbf{P12}\quad Identified Book 1: yes\quad Identified Book 2: yes\quad Expected worse: yes}\\
\midrule
Q1 & I think I was expecting less than what I read. As in, I thought it would be easier to detect the AI. \\
\addlinespace[2pt]
Q2 & No, I'm not surprised. I do a LOT of work with AI and there are some tells that give it away. There is also the conversational tone that AI hasn't quite nailed down. \\
\addlinespace[2pt]
Q3 & I think the first book, the one about the spaceship, was much better. I don't know if it's the original author writing the book in a more conversational tone, or if the AI chose better words. The Michealangelo book was extremely difficult to read. Both version were challenging, which tells me that's the way the author wrote it. \\
\addlinespace[2pt]
Q4 & N/A \\
\addlinespace[6pt]
\multicolumn{2}{@{}l@{}}{\textbf{P13}\quad Identified Book 1: yes\quad Identified Book 2: no\quad Expected worse: yes}\\
\midrule
Q1 & I expected the quality of AI translation to be stilted and mechanical with the words not always in the right order. \\
\addlinespace[2pt]
Q2 & Yes, I'm surprised by incorrectly identifying the AI-translation version of Book 2. \\
\addlinespace[2pt]
Q3 & I still feel that the AI translation lacks a human aspect, with less vivid descriptions. \\
\addlinespace[2pt]
Q4 & The human translation can vary depending on the experience/quality of the translator. A human translator who is familiar with a particular place or culture can bring their own human experience to the translation. \\
\addlinespace[6pt]
\multicolumn{2}{@{}l@{}}{\textbf{P14}\quad Identified Book 1: yes\quad Identified Book 2: yes\quad Expected worse: yes}\\
\midrule
Q1 & I was expecting the quality to be somewhat worse than what I read in the provided texts. More missing context due to language/cultural differences, more literal translations without replacing idioms, etc. \\
\addlinespace[2pt]
Q2 & I thought it would be easier to identify the AI translation than it was in reality. When I had two texts side by side it was quite easy, but when I read them separately, it was quite similar to poorly done translation by a human amateur. \\
\addlinespace[2pt]
Q3 & Both AI translations in the text were more or less acceptable from the point of view of literal translation. However, they were missing emotional import, poetic sense and failed to take into consideration that some things should be changed in translation based on the cultural and language differences in the output language. \\
\addlinespace[2pt]
Q4 & I think AI can be a useful tool if we are stuck and are trying to see how something can be said or written, but it should only be treated as a tool or a guide, there is no way that AI can properly convey everything - emotion, language style, etc, and take into consideration many, many things around the language that we do not think of but notice the difference when reading the text. \\
\addlinespace[6pt]
\multicolumn{2}{@{}l@{}}{\textbf{P15}\quad Identified Book 1: yes\quad Identified Book 2: yes\quad Expected worse: no}\\
\midrule
Q1 & I expected a grammatically correct translation of AI. It has become much better in the past few months and I anticipated a robust and accurate interpretation of the narrative. In the first book, one translation was significantly better than the other. In the second book, I was disappointed by the grammar and use of words. There was lack of depth and engagement. \\
\addlinespace[2pt]
Q2 & I am not surprised by my accuracy. I read a lot of AI and work with it every day. So I can tell by grammar, use of colons and em dashes, and choice of words pretty clearly. \\
\addlinespace[2pt]
Q3 & I think the translations need work. The first book was superior to the second book. The translations were poorly organized and lacked compelling narrative. \\
\addlinespace[2pt]
Q4 & It would be fairly easy to fix the translations so they don't sound so AI driven. HUmanizing does not take that much time. Please feel free to reach out if you need help with anything else. \\
\bottomrule
\end{tabularx}
\caption{Post-study responses grouped by participant (part 4 of 4; participants P12--P15). Questions are as in Table~\ref{tab:post-study-responses-1}.}
\label{tab:post-study-responses-4}
\end{table*}

\begin{table*}[t]
\centering
\small
\begin{tabularx}{\textwidth}{@{}p{0.13\textwidth} p{0.18\textwidth} p{0.11\textwidth} X@{}}
\toprule
Level & Book & Preferred & Participant comment \\
\midrule
Excerpt & \textit{Needle's Eye} & HT &
HT was far easier to follow in terms of flow and word choice. MT had at times, bizarre punctuation choices and was very choppy. MT made it difficult to determine who was speaking, and while HT also lacked some context, it was just due to the story being short and its style. With MT, it was difficult to follow because it switched around, sometimes using quotation marks and sometimes using dashes. It did not have a flow to it. \\

Excerpt & \textit{The Story of Marceau Miller} & HT &
HT was much easier for me to read than MT. The wording flowed better and made it a seamless reading experience, whereas in MT I felt like I had to reread sentences now and then to make sure I was understanding right. HT also felt more like reading a story in English rather than a translated story. MT definitely read like someone or something translating word by word; it didn't feel natural while reading. \\

Excerpt & \textit{The Story of Marceau Miller} & MT &
I enjoyed both translations immensely and found them easy to read. I prefer MT's word choices because they come off as more intricate and varied to me. It is more formal than HT's phrasing in some parts, without feeling unnatural. I felt like I was able to pick up the essence of the characters more in MT, for instance, the difference between Sarah thinking ``He melts me, the idiot'' and ``I'm touched.'' MT just gave a clearer picture of personalities and relationships with its words, in my opinion. \\

Excerpt & \textit{The Witcher: Crossroads of Ravens} & MT &
MT also made language choices (particularly the use of an older style of English) that are generally more consistent with this style of fantasy e.g. [MT] ``Do not take offence'' vs. [HT] ``Don't be cross''. \\

Excerpt & \textit{Kokun: The Girl from the West} & HT &
HT had no formatting issues that I noticed, for one, and also had a richness/uniqueness to it that pulled me in. The dialogue flowed well and, while complicated at times, is not unusual for an epic fantasy book of this nature. \\

Chunk & \textit{Sisters in Yellow}, chunk 4 & MT &
I much preferred ``bento containers'' over ``plastic containers'' as it not only falls in line with the Japanese culture, it also doesn't sound cheap. \\

Chunk & \textit{The Empusium}, chunk 18 & MT &
The older-style language of MT better matches the period the story is set in. HT is a little too modern, using words like ``glasses'' instead of ``spectacles'' and ``frames'' instead of ``rims.'' \\

Chunk & \textit{The Witcher: Crossroads of Ravens}, chunk 14 & MT &
I preferred the chief digger's dialogue in MT, it seemed to fit his character better. It also contrasted with the style of Preston Holt better, while in HT both characters had more similar dialogue patterns. \\

Chunk & \textit{My Name Is Stramer}, chunk 10 & MT &
I could imagine the stories better in MT with more vivid language and I liked the phrase ``windmill his arms'' which made the wording come alive. \\

Chunk & \textit{Mona's Eyes}, chunk 21 & MT &
The dialogue in both feels clumsy in parts, especially the way Mona speaks, but I'm guessing that might just be how the original book was written. \\

Chunk & \textit{Kokun: The Girl from the West}, chunk 20 & HT &
A lot of the sentences were far too long, and this, combined with word choice, made the paragraphs difficult to follow (this could've just been a poorly written original scene, though, and not necessarily poorly translated). \\
\bottomrule
\end{tabularx}
\caption{Examples of contextual reader judgments in excerpt-level and chunk-level preference explanations. These examples include the full comments behind selected snippets quoted in the main text and illustrate how participants evaluated translation choices against perceived story context, including genre, period, cultural setting, character voice, and reading effort. Participants saw anonymized T1/T2 labels during evaluation; we replace those labels with HT and MT here for readability.}
\label{tab:example-contextual-judgment-examples}
\end{table*}

\section{Topic modeling support}
\label{sec:topic-model}
This section describes the label-blind topic-model analysis used to validate the human inductive coding.

\paragraph{Model-assisted inductive coding.}
We use GPT-5.4 as a topic model to support the inductive coding process. 
We pool 922 comments from the open-ended questions about translation quality (\questionref{i-Q5}, \questionref{i-Q6}, \questionref{co-Q4}, and \questionref{ch-Q3}).
We exclude 7 comments because they do not express an opinion about translation quality, leaving 915 comments.
We also analyze 30 AI-detection explanations from \questionref{co-Q7} separately. 
The model receives the participant comments and the question, but no manual annotations or codebooks. 
It assigns each comment to one, none or more topics, and a short description.
The exact prompts and run settings are available in our GitHub repository (\href{https://github.com/Yves575/lait/tree/main/analysis/human_eval}{\path{analysis/human_eval}}). 
The resulting topic inventories are reported in \autoref{tab:gpt-topics}.

\paragraph{Topic inventory selection.}
We repeat label discovery ten times for each inventory. 
Without consulting the manual labels, we match topics across runs using best-match Jaccard similarity. 
We then select the run whose topics are, on average, most similar to those from the other nine runs.
The selected inventories contain 24 translation-quality topics and 15 AI-detection-rationale topics (mean Jaccard: 0.568 and 0.687, respectively). 
Using the selected topic list, we ask GPT-5.4 to independently assign topics to every comment three times.
Across these runs, topics contain similar sets of comments (mean Jaccard: 0.653 for translation quality and 0.827 for AI detection), and individual comments receive similar sets of topics (0.746 and 0.814, respectively). 
Final assignments retain topics identified in at least two runs.

We compare these label-blind topics with the final human schemas (29 translation-quality labels and 10 AI-detection labels) to assess whether the themes identified through human inductive coding can be independently recovered. This provides schema-level validation, not annotator agreement.

\begin{table*}[t]
\centering
\small
\setlength{\tabcolsep}{3pt}
\begin{tabularx}{\textwidth}{@{}r r r >{\raggedright\arraybackslash}X
    @{\hspace{5pt}\vrule width 0.4pt\hspace{2pt}\vrule width 0.4pt\hspace{5pt}}
    r r r >{\raggedright\arraybackslash}X@{}}
\toprule
\multicolumn{4}{c}{\textbf{Translation quality}}
    & \multicolumn{4}{c}{\textbf{AI detection}} \\
\cmidrule(r){1-4}\cmidrule(l){5-8}
Count & $N$ & Stab. & GPT-5.4-generated topic
    & Count & $N$ & Stab. & GPT-5.4-generated topic \\
\midrule
1 & 361 & 0.78 & Word choice and diction fit
    & 1 & 10 & 0.65 & Lexical fit \\
2 & 226 & 0.67 & Clarity and ease of understanding
    & 2 & 9 & 0.74 & Source-text literalness \\
3 & 210 & 0.75 & Flow and readability
    & 3 & 9 & 0.93 & Sentence and paragraph flow \\
4 & 172 & 0.69 & Sentence structure and rhythm
    & 4 & 9 & 0.86 & Punctuation and dialogue formatting \\
5 & 170 & 0.66 & Naturalness and idiomatic English
    & 5 & 9 & 0.70 & Semantic and contextual resolution \\
6 & 114 & 0.60 & Meaning accuracy
    & 6 & 6 & 1.00 & Register and diction level \\
7 & 98 & 0.59 & Dialogue authenticity
    & 7 & 5 & 0.64 & Sentence-level correctness \\
8 & 97 & 0.73 & Verbosity and concision
    & 8 & 5 & 0.77 & Emotional and tonal conveyance \\
9 & 92 & 0.67 & Grammar and surface mechanics
    & 9 & 4 & 0.72 & Adaptive rephrasing and localization \\
10 & 76 & 0.68 & Imagery and descriptive vividness
    & 10 & 4 & 1.00 & Lexical richness and specificity \\
11 & 65 & 0.51 & Literary richness and atmosphere
    & 11 & 3 & 1.00 & Information density and verbal economy \\
12 & 57 & 0.66 & Formatting and presentation
    & 12 & 3 & 0.78 & Variety and register consistency \\
13 & 53 & 0.73 & Cultural and contextual mediation
    & 13 & 3 & 0.83 & Formatting artifact handling \\
14 & 53 & 0.81 & Engagement and immersion
    & 14 & 3 & 0.78 & Voice differentiation \\
15 & 51 & 0.79 & Machine-like or translationese feel
    & 15 & 2 & 1.00 & Terminology and naming consistency \\
16 & 47 & 0.74 & Character voice and viewpoint fit
    & & & & \\
17 & 44 & 0.56 & Coherence, transitions, and organization
    & & & & \\
18 & 42 & 0.57 & Emotional resonance and nuance
    & & & & \\
19 & 40 & 0.72 & Literalness versus adaptation
    & & & & \\
20 & 37 & 0.73 & Genre, setting, and period appropriateness
    & & & & \\
21 & 35 & 0.47 & Tense and grammatical perspective consistency
    & & & & \\
22 & 29 & 0.61 & Referent and pronoun clarity
    & & & & \\
23 & 28 & 0.64 & Consistency of recurring terms and details
    & & & & \\
24 & 25 & 0.34 & Register, tone, and stylistic fit
    & & & & \\
\bottomrule
\end{tabularx}
\caption{GPT-5.4-generated candidate topics for translation-quality comments
(\questionref{i-Q5}, \questionref{i-Q6}, \questionref{co-Q4}, and
\questionref{ch-Q3}) and AI-detection explanations (\questionref{co-Q7}). $N$ is the number of comments assigned to the topic by majority vote across three runs. Stab. is the mean pairwise Jaccard similarity of the topic's comment membership across those runs. Complete topic definitions and evidence phrases are
available in the GitHub repository.}
\label{tab:gpt-topics}
\end{table*}

\section{Exploratory multilingual case study}
\label{app:multilingual-case-study-results}

In this section of the appendix, we detail our case study targeting translation into languages other than English, specifically Spanish, French, Polish, and Japanese.

\begin{figure}[t]
    \centering
    \includegraphics[
        width=0.94\columnwidth,
        height=0.30\textheight,
        keepaspectratio
    ]{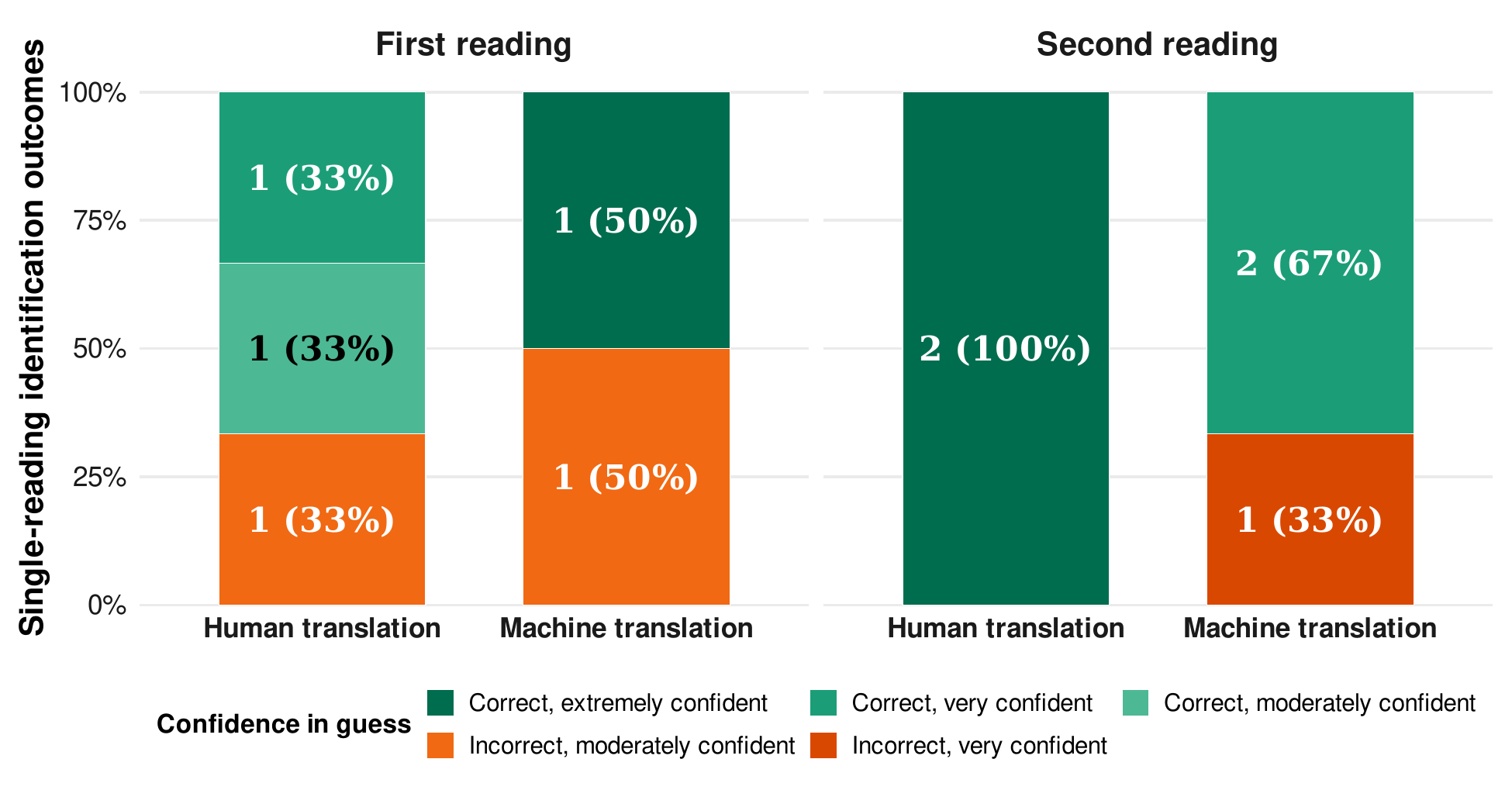}
    \caption{Multilingual case-study identification outcomes. Single-reading results for HT/MT identification in the multilingual case study; darker shades indicate higher confidence.}
    \label{fig:origin_guess_accuracy_by_reading}
\end{figure}

\begin{figure}[t]
    \centering
    \includegraphics[
        width=0.94\columnwidth,
        height=0.30\textheight,
        keepaspectratio
    ]{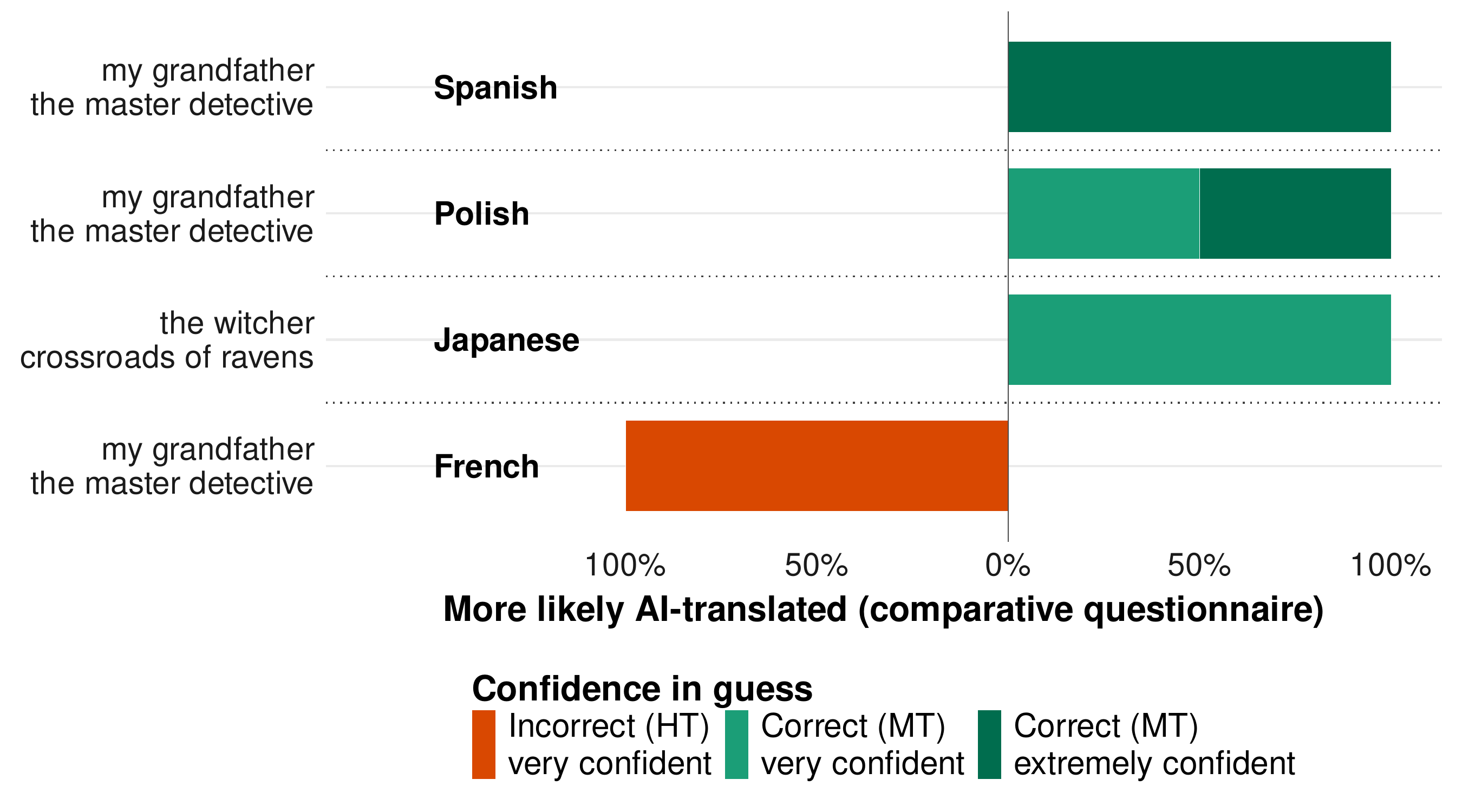}
    \caption{More likely AI-translated choices in the comparative questionnaire. After both readings, readers chose which version seemed more likely AI-translated. Orange bars show incorrect HT choices; green bars show correct MT choices, with darker shades indicating higher confidence.}
    \label{fig:book_ai_identification_diverging}
\end{figure}

\paragraph{MT identification.} Readers reliably identified MT for Polish (two raters), Spanish, and Japanese in the final excerpt-level comparison, while the French reader was incorrect (\autoref{fig:book_ai_identification_diverging}). We also note that the Spanish reader was initially incorrect, identifying MT as HT when they first encountered it in single-reading (\autoref{fig:origin_guess_accuracy_by_reading}), but they were able to reliably distinguish both during comparison, suggesting that perhaps the quality of translation into French and Spanish was higher than into Polish and Japanese.

\begin{figure*}[t]
    \centering
    \includegraphics[
        width=0.78\textwidth,
        height=0.33\textheight,
        keepaspectratio
    ]{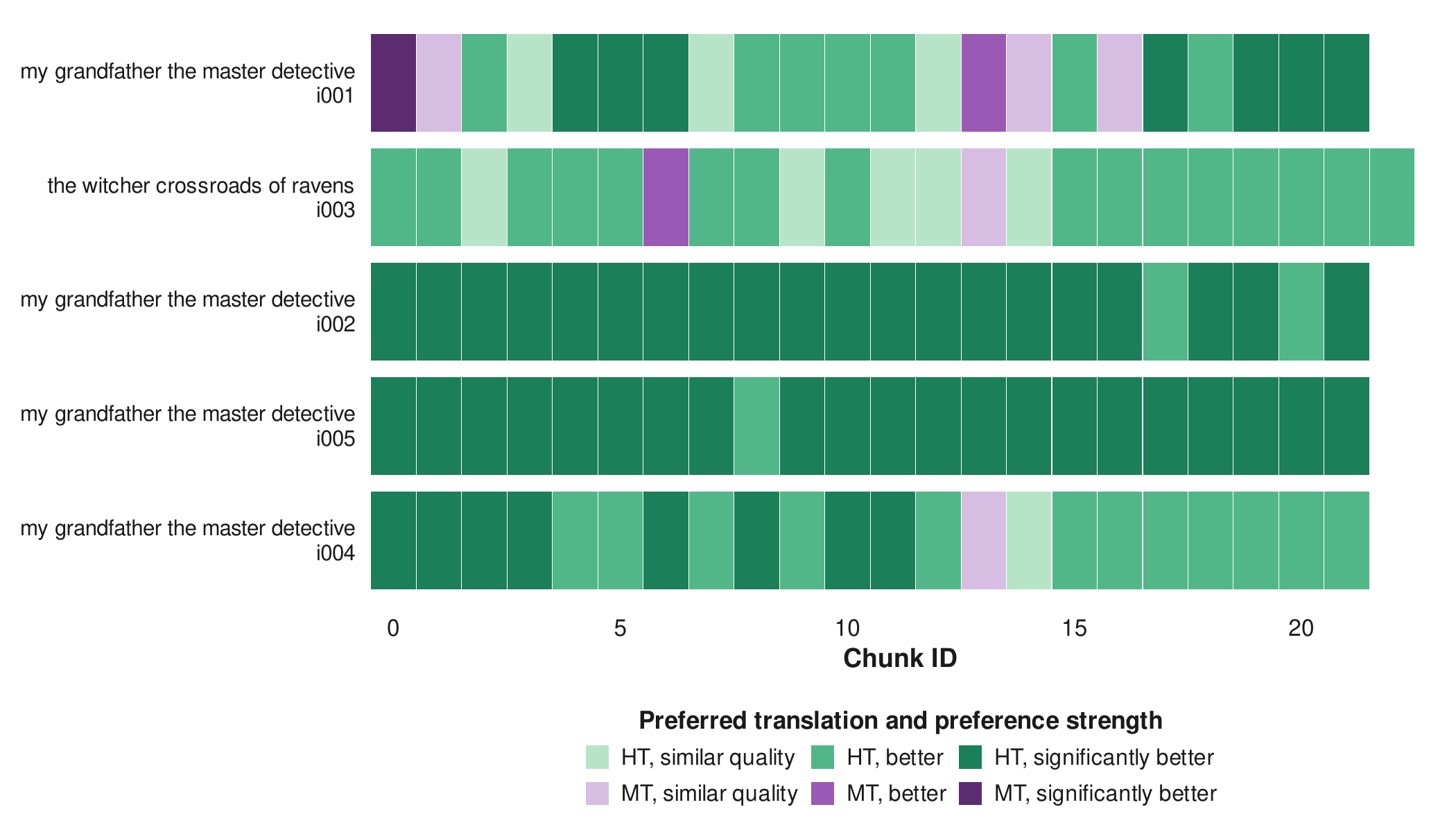}
    \caption{Chunk-level preferred translation in the multilingual target-language case study. Each row shows one excerpt-reader pair; green indicates HT preference, purple indicates MT preference, and darker shades indicate stronger preference.}
    \label{fig:part2_chunk_choice_heatmap}
\end{figure*}

\paragraph{Span annotations.} Across target languages, readers marked mostly positive evidence in HT and mostly negative evidence in MT (\autoref{fig:span_highlights_by_language}), with median span lengths reported to indicate how localized these judgments were (\autoref{fig:span_highlight_median_length_by_language}).

\section{Statistical analysis}
\label{sec:stat-test-results}

\begin{table*}[t]
    \centering
    \begingroup
    \scriptsize
    \setlength{\tabcolsep}{3.5pt}
    \renewcommand{\arraystretch}{1.12}
    \newcommand{\modelrow}[2]{%
        \addlinespace[0.35em]
        \rowcolor[gray]{0.95}%
        \multicolumn{5}{@{}>{\RaggedRight\arraybackslash}p{\textwidth}@{}}{%
            \textbf{\texttt{#1}}\quad\texttt{#2}}\\
        \addlinespace[0.15em]
    }
    \begin{tabular*}{\textwidth}{@{\extracolsep{\fill}}
        >{\RaggedRight\arraybackslash}p{0.13\textwidth}
        >{\RaggedRight\arraybackslash}p{0.30\textwidth}
        >{\RaggedRight\arraybackslash}p{0.30\textwidth}
        r
        r
    @{}}
        \toprule
        \textsc{Counts} & \textsc{Outcome coding} & \textsc{Effect size [95\% CI]} & $p$ & \textsc{Order} $p$ \\
        \midrule
        \modelrow{excerpt\_preference}{glm(prefer\_HT \textasciitilde{} order, family = binomial, data = excerpt)}
            19/11
            & \texttt{prefer\_HT}: HT=1, MT=0
            & HT probability = 63.4\%; OR = 1.73 [0.82, 3.65]
            & .148
            & .705 \\
        \modelrow{chunk\_preference}{glmer(prefer\_HT \textasciitilde{} order + (1|reader) + (1|book), family = binomial, data = chunks)}
            522/250
            & \texttt{prefer\_HT}: HT=1, MT=0
            & HT probability = 76.2\%; OR = 3.20 [1.31, 7.82]
            & .011
            & .536 \\
      \modelrow{chunk\_preference (primary)}{glmer(prefer\_HT \textasciitilde{} order + (1|reader) + (1|book) + (1|reader:book), family = binomial, data = chunks)}
        522/250
        & \texttt{prefer\_HT}: HT=1, MT=0
        & OR = 3.28 [1.41, 7.59]
        & .0056
        & .327 \\
        \midrule
        \modelrow{excerpt\_strength}{lm(signed\_strength \textasciitilde{} order, data = excerpt)}
            19/11
            & signed score: HT positive, MT negative; max $|\cdot|=2$
            & signed mean = 0.57 [-0.10, 1.23]
            & .092
            & .612 \\
        \modelrow{chunk\_strength}{lmer(signed\_strength \textasciitilde{} order + (1|reader) + (1|book), data = chunks)}
            522/250
            & signed score: HT positive, MT negative; max $|\cdot|=3$
            & signed mean = 0.91 [0.24, 1.58]
            & .010
            & .176 \\
        \midrule
        \modelrow{dialogue}{lm(signed\_attribute \textasciitilde{} order, data = excerpt)}
            17/9/4
            & signed score: HT=1, no difference=0, MT=-1
            & signed mean = 0.27
            & .109
            & .109 \\
        \modelrow{word\_choice}{lm(signed\_attribute \textasciitilde{} order, data = excerpt)}
            18/10/2
            & signed score: HT=1, no difference=0, MT=-1
            & signed mean = 0.27
            & .131
            & .253 \\
        \bottomrule
    \end{tabular*}
    \caption{Statistical analyses for HT-vs-MT preferences and excerpt-level attribute preferences. Counts are HT/MT for binary and strength analyses, and HT/MT/no-difference for \texttt{dialogue} and \texttt{word\_choice}. Binary models code HT as 1 and MT as 0. Strength and attribute models use signed scores, with HT coded positive and MT coded negative. Excerpt-level mixed-effects models were singular, so simpler order-adjusted fixed-effect models are reported for excerpt outcomes; chunk-level models include random intercepts for reader and book. Effects are order-adjusted; brackets give 95\% confidence intervals.}
    \label{tab:preference-model-effects}
    \endgroup
\end{table*}

\begin{table*}[t]
    \centering
    \begingroup
    \scriptsize
    \setlength{\tabcolsep}{4pt}
    \renewcommand{\arraystretch}{1.12}
    \newcommand{\modelrow}[2]{%
        \addlinespace[0.35em]
        \rowcolor[gray]{0.95}%
        \multicolumn{5}{@{}>{\RaggedRight\arraybackslash}p{\textwidth}@{}}{%
            \textbf{\texttt{#1}}\quad\texttt{#2}}\\
        \addlinespace[0.15em]
    }
    \begin{tabular*}{\textwidth}{@{\extracolsep{\fill}}
        r
        r
        >{\RaggedRight\arraybackslash}p{0.42\textwidth}
        r
        r
    @{}}
        \toprule
        \textsc{HT count} & \textsc{MT count} & \textsc{Effect size [95\% CI]} & $p$ & \textsc{Order} $p$ \\
        \midrule
        \modelrow{positive\_spans}{glmer.nb(count \textasciitilde{} version + order + (1|reader) + (1|book), data = highlights)}
            2099
            & 1260
            & HT/MT rate ratio = 1.69 [1.55, 1.85]
            & <.001
            & .596 \\
        \modelrow{negative\_spans}{glmer.nb(count \textasciitilde{} version + order + (1|reader) + (1|book), data = highlights)}
            1147
            & 2728
            & HT/MT rate ratio = 0.43 [0.39, 0.48]
            & <.001
            & .052 \\
        \bottomrule
    \end{tabular*}
    \caption{Chunk-level highlight count statistical analysis. Positive and negative evidence spans were modeled separately with negative-binomial mixed-effects count models with random intercepts for \textit{reader} and \textit{book}. Rate ratios compare HT to MT; values above 1 indicate more spans on HT, and values below 1 indicate fewer spans on HT.}
    \label{tab:highlight-count-model-effects}
    \endgroup
\end{table*}

\begin{table*}[t]
    \centering
    \begingroup
    \scriptsize
    \setlength{\tabcolsep}{3.5pt}
    \renewcommand{\arraystretch}{1.12}
    \newcommand{\modelrow}[2]{%
        \addlinespace[0.35em]
        \rowcolor[gray]{0.95}%
        \multicolumn{8}{@{}>{\RaggedRight\arraybackslash}p{\textwidth}@{}}{%
            \textbf{\texttt{#1}}\quad\texttt{#2}}\\
        \addlinespace[0.15em]
    }
    \begin{tabular*}{\textwidth}{@{\extracolsep{\fill}}
        >{\RaggedRight\arraybackslash}p{0.15\textwidth}
        >{\RaggedRight\arraybackslash}p{0.18\textwidth}
        rrrr
        >{\RaggedRight\arraybackslash}p{0.17\textwidth}
        >{\RaggedRight\arraybackslash}p{0.17\textwidth}
    @{}}
        \toprule
        \textsc{Response} & \textsc{Predictor} & \textsc{Estimate} & SE & $z$ & $p$ & OR [95\% CI] & \textsc{HT advantage} \\
        \midrule
        \modelrow{acceptability}{clmm(q1 \textasciitilde{} type + order + (1|reader) + (1|book), data = model\_data, link = "logit")}
            & \texttt{typeMT} & -1.38 & 0.51 & -2.70 & .0069 & 0.25 [0.09, 0.68] & $4.0\times$ odds \\
            & \texttt{orderMT-first} & 0.41 & 0.49 & 0.85 & .3978 & 1.51 [0.58, 3.90] & -- \\
        \modelrow{smoothness}{clm(q2 \textasciitilde{} type + order, data = model\_data, link = "logit")}
            & \texttt{typeMT} & -1.49 & 0.50 & -2.98 & .0029 & 0.23 [0.08, 0.59] & $4.3\times$ odds \\
            & \texttt{orderMT-first} & 0.30 & 0.48 & 0.63 & .5279 & 1.36 [0.53, 3.51] & -- \\
        \modelrow{immersion}{clm(q3 \textasciitilde{} type + order, data = model\_data, link = "logit")}
            & \texttt{typeMT} & -0.83 & 0.48 & -1.74 & .0815 & 0.44 [0.17, 1.10] & $2.3\times$ odds \\
            & \texttt{orderMT-first} & 0.27 & 0.47 & 0.56 & .5724 & 1.30 [0.52, 3.30] & -- \\
        \modelrow{continue\_reading}{clmm(q4 \textasciitilde{} type + order + (1|reader) + (1|book), data = model\_data, link = "logit")}
            & \texttt{typeMT} & -0.79 & 0.50 & -1.58 & .1150 & 0.45 [0.17, 1.21] & $2.2\times$ odds \\
            & \texttt{orderMT-first} & 0.07 & 0.49 & 0.14 & .8860 & 1.07 [0.41, 2.79] & -- \\
        \bottomrule
    \end{tabular*}
    \caption{Cumulative-link model results for single-reading 5-point ratings ($n=60$ per response, 30 HT and 30 MT). HT is the reference for \texttt{type}; odds ratios below 1 for \texttt{typeMT} indicate lower odds of a higher rating for MT than HT. \textsc{HT advantage} reports the inverse odds ratio for translation effects only.}
    \label{tab:single-reading-model-effects}
    \endgroup
\end{table*}

\begin{table*}[t]
\centering\small
\setlength{\tabcolsep}{3pt}
\begin{tabular*}{\textwidth}{@{\extracolsep{\fill}}l rrrrrr@{}}
\toprule
& \multicolumn{6}{c}{$p$ for \texttt{typeMT} by random-effects specification} \\
\cmidrule(lr){2-7}
\textsc{Response} & \textsc{primary} & \textsc{+b slope} & \textsc{+r slope} & \textsc{+b slope}$^0$ & \textsc{+r slope}$^0$ & \textsc{maximal}$^\dagger$ \\
\midrule
\textbf{Acceptability }    & .0069 & .0382 & .0139 & .0154 & .0121 & .0428 \\
\textbf{Smoothness }       & .0029 & .0180 & .0029 & .0044 & .0029 & .0180 \\
\textbf{Immersion }        & .0815 & .1196 & .1461 & .1076 & .1279 & .1988 \\
\textbf{Continue reading}  & .1150 & .1883 & .2764 & .1732 & .2726 & .3072 \\
\bottomrule
\end{tabular*}
\caption{Sensitivity of the translation-type effect to random-effects structure (cumulative-link models; slopes on book (b) or reader (r), with $^0$ marking zero-correlation variants). All slope-bearing fits are at the boundary (intercept-slope correlation $=-1$); fixed effects are reported at face value following the reduction protocol of \citet{barr2013random}. For smoothness, the reader-slope fit reduces to the fallback model. $^\dagger$Saturated maximal model (60 random effects for 60 observations).}
\label{tab:ratings-spec-grid}
\end{table*}

\begin{table*}[t]
    \centering
    \begingroup
    \scriptsize
    \setlength{\tabcolsep}{3.5pt}
    \renewcommand{\arraystretch}{1.12}
    \begin{tabular*}{\textwidth}{@{\extracolsep{\fill}}
        >{\RaggedRight\arraybackslash}p{0.10\textwidth}
        >{\RaggedRight\arraybackslash}p{0.36\textwidth}
        >{\RaggedRight\arraybackslash}p{0.16\textwidth}
        >{\RaggedRight\arraybackslash}p{0.15\textwidth}
        r
        r
    @{}}
        \toprule
        \textsc{Analysis} & \textsc{Model} & \textsc{Data} & \textsc{Effect} & \textsc{Model} $p$ & \textsc{Exact} $p$ \\
        \midrule
        Single-reading origin accuracy &
        \texttt{glmer(correct \textasciitilde{} 1 + (1|person\_id) + (1|book\_id), family = binomial)} &
        34/60 correct &
        accuracy = 58.3\% &
        .413 &
        .366 \\
        \midrule
        Single-reading HT vs.\ MT accuracy &
        \texttt{glmer(correct \textasciitilde{} version + (1|person\_id) + (1|book\_id), family = binomial)} &
        MT: 16/30; HT: 18/30 &
        no version difference &
        .557 &
        -- \\
        \midrule
        Direct-comparison origin accuracy &
        \texttt{glm(correct \textasciitilde{} 1, family = binomial)} &
        17/30 correct &
        accuracy = 56.7\% &
        .467 &
        .585 \\
        \midrule
        Correct-guess confidence &
        \texttt{lmer(confidence \textasciitilde{} stage + (1|person\_id) + (1|book\_id))} &
        51 correct guesses &
        comparison--single = 0.40 &
        .129 &
        -- \\
        \bottomrule
    \end{tabular*}
    \caption{Statistical analysis for AI detection. In single reading, a correct response means labeling an MT excerpt as machine-translated or an HT excerpt as human-translated. In direct comparison, a correct response means selecting the actual MT version as more likely AI-translated. The requested direct-comparison crossed mixed-effects model was singular, so the intercept-only logistic model is reported; the exact binomial test is included because the comparison is against chance.}
    \label{tab:origin-identification-models}
    \endgroup
\end{table*}

In this section of the appendix, we report details on the statistical models used for the quantitative results. 
The statistical models were chosen based on the type of response variable and data distribution. All analyses were conducted in R using \texttt{lme4} for mixed-effects models, \texttt{lmerTest} for inference in linear mixed-effects models, \texttt{emmeans} for order-adjusted marginal means and contrasts, and \texttt{ordinal} for cumulative-link models. The exact formulas and outputs are provided in \autoref{tab:preference-model-effects} for preference judgments, \autoref{tab:highlight-count-model-effects} for span highlight, \autoref{tab:single-reading-model-effects} for Likert-type ratings, and \autoref{tab:origin-identification-models} for AI detection. 

\paragraph{Binary preferences.}
For binary choices (HT vs MT), we use logistic generalized linear mixed-effects models (GLMMs), with HT coded as 1 and MT as 0. These models estimate whether readers are more likely to choose HT than MT. When estimable, we include random intercepts for \textit{reader} and \textit{book} to account for repeated judgments by the same readers and repeated judgments of the same books.

For dialogue and word-choice comparisons, responses had three levels: HT, MT, and no meaningful difference. We coded these as signed attribute preferences, with HT = +1, no difference = 0, and MT = -1. Crossed mixed-effects models with \textit{reader} and \textit{book} random intercepts were singular, hence we report order-adjusted fixed-effect linear models.

\paragraph{Preference strength.}
For preference strength, we convert choices into signed numeric scores. HT preferences are coded as positive and MT preferences as negative, with larger absolute values indicating stronger preferences. We analyze these signed scores with linear mixed-effects models when the random-effects structure is estimable (random intercepts for \textit{reader} and \textit{book}).

\paragraph{Single-reading ratings.}
For 5-point Likert ratings, such as \texttt{acceptability}, \texttt{smoothness}, \texttt{immersion}, and willingness to \texttt{continue reading}, we use cumulative-link mixed models with random intercepts for \textit{reader} and \textit{book} when estimable. These models are designed for ordered categorical responses, where the difference between adjacent scale points should not be treated as necessarily equal. 
We additionally tested by-book and by-reader random slopes for translation type. Across the six tested random-effects specifications, the worst-case \textit{p}-values were .043 for acceptability, .018 for smoothness, .199 for immersion, and .307 for willingness to continue reading. Acceptability and smoothness therefore remained significant, whereas immersion and willingness to continue remained non-significant.

\paragraph{AI detection.}
For AI-detection questions, responses were coded as binary correctness outcomes with 1 if the reader correctly identified the translation origin and 0 otherwise. In the single-reading task, MT excerpts were counted as correct when labeled machine-translated, and HT excerpts when labeled human-translated. We analyzed these responses with logistic generalized linear mixed-effects models, including random intercepts for \textit{reader} and \textit{book} when estimable. In the direct-comparison task, correctness meant selecting the actual MT version as more likely AI-translated; the crossed mixed-effects model was singular, so we report the corresponding intercept-only logistic model and an exact binomial test against chance. Confidence ratings were analyzed with linear mixed-effects models, restricted to correct guesses when testing whether correct responses were more confident after direct comparison.

\paragraph{Highlight counts.}
For highlighted evidence spans, we use negative-binomial mixed-effects count models. These models are appropriate for count data when the variance is larger than expected under a simple Poisson model.

\paragraph{Random effects.}
Across mixed-effects analyses, we include random intercepts for \textit{reader} and \textit{book} whenever estimable (i.e., does not result in a \textit{singular fit}). If a mixed-effects model is singular, meaning that one or more random-effect variances are estimated as zero or near-zero, we report the corresponding simpler model.

\section{Automatic metrics preferences}
\label{app:auto-eval}

In this section, we explore how well automatic metrics can capture readers' preferences for the chunk-level judgments.

\paragraph{Paragraph-level alignment.}
Because several automatic metrics are constrained by context-window size, we evaluate the translations on shorter aligned paragraph units rather than complete 8K-word excerpts or even 300-word chunks.
We follow the \texttt{par3} alignment procedure \cite{thai-etal-2022-exploring}.
For each excerpt, source paragraphs are paired with paragraph-level Microsoft Translate output, which serves as an English pivot.
Each candidate English version, including professional \textsc{HT} and each MT system output, is sentence-aligned to this pivot.
The candidate sentences aligned to each pivot paragraph are then grouped under the original source paragraph boundaries.
After alignment, we compute diagnostics for each recovered paragraph pair and manually inspect flagged cases.
We flag a length issue when the larger paragraph is more than three times longer than the smaller paragraph, ignoring pairs where both sides are shorter than 10 characters.
We also flag lexical-overlap issues when a candidate paragraph has a pairwise BLEU score below 1.5, against a majority of the other MT outputs aligned to the same source paragraph.
We employ these heuristics to filter misaligned paragraphs.

\paragraph{Paragraph-level metrics.}
On the aligned paragraph pairs, we run two metric families: \textsc{COMET} and \textsc{MetricX}.
For each family, we use one reference-based metric and one reference-free quality-estimation metric: \textsc{COMET-22} and \textsc{MetricX} use the professional \textsc{HT} as reference, while \textsc{COMETKiwi} \cite{rei-etal-2022-cometkiwi} and \textsc{MetricX-QE} \cite{juraska-etal-2024-metricx} use only the source and candidate translation.
\textsc{COMET} scores are higher-is-better, whereas \textsc{MetricX} scores are lower-is-better.
The metric input limits are 512 tokens for the \textsc{COMET} family and 1,536 tokens for the \textsc{MetricX} family.
For reference-based metrics, this budget includes the source, candidate translation, and reference.
For reference-free metrics, it only includes the source and candidate translation.
Paragraph pairs that exceed the corresponding model limit are skipped for that metric.
Reference-based metrics are not reported for the \textsc{HT} row, since this would score the reference against itself.
For each book and system, we report the mean of all remaining paragraph scores for each metric.

Paragraph-level \texttt{par3} automatic-evaluation results for all systems are summarized in
\autoref{tab:appendix_par3_model_metric_results}.

\paragraph{Chunk-level comparison with human judgments.}
We evaluate the same approximately 300-word aligned chunks used in the chunk-level \textit{close reading} task.
We score the \textsc{HT} and agentic/P3 \textsc{MT} chunks directly with \textsc{MetricX-QE} and with \textsc{LiTransProQA} \cite{zhang-etal-2025-litransproqa}, using Gemini 3.1 Pro as the question-answering judge \cite{googledeepmind2026gemini31pro}.
In addition, we map the paragraph-level \textsc{COMETKiwi} and \textsc{MetricX-QE} scores onto the close-reading chunks.
This mapping matches paragraph-level source text to the chunk source text in order and averages all paragraph scores assigned to a chunk.
When both \textsc{HT} and \textsc{MT} have paragraph scores for the same metric, we average over the shared paragraph indices so that the two systems are compared on the same source material.
For every metric, we compute an oriented \textsc{HT}--\textsc{MT} delta: positive values favor \textsc{HT}, and negative values favor the agentic/P3 \textsc{MT}.
Chunk-level HT--MT metric deltas used in the human-judgment comparison are summarized in
\autoref{tab:appendix_chunk_metric_results}.

\begin{table*}[t]
    \centering
    \small
    \setlength{\tabcolsep}{5pt}
    \resizebox{\textwidth}{!}{%
        \begin{tabular}{llcrrrr}
            \toprule
            \multicolumn{3}{l}{\textsc{System}} & \multicolumn{4}{c}{\textsc{Automatic metric}} \\
            \cmidrule(lr){1-3}\cmidrule(lr){4-7}
            Model & Pipeline & Human eval. & COMET-22 $\uparrow$ & COMETKiwi $\uparrow$ & MetricX $\downarrow$ & \textbf{MetricX-QE $\downarrow$} \\
            \midrule
            \rowcolor{pTwoRow}
            Gemini & P2 & & \textbf{0.805} & \textbf{0.773} & \textbf{5.182} & \textbf{4.758} \\
            \rowcolor{pOneRow}
            Gemini & P1 & & 0.804 & 0.772 & 5.243 & 4.813 \\
            \rowcolor{pTwoRow}
            GPT-5.4 & P2 & & 0.800 & 0.768 & 5.482 & 5.020 \\
            \rowcolor{pOneRow}
            GPT-5.4 & P1 & & 0.798 & 0.766 & 5.542 & 5.073 \\
            \rowcolor{pThreeRow}
            \textbf{Agents} & P3 & \faCheck & 0.793 & 0.766 & 5.656 & 5.109 \\
            \rowcolor{htRow}
            \textbf{\textsc{HT}} & -- & \faCheck & -- & 0.726 & -- & 5.726 \\
            \bottomrule
        \end{tabular}
    }
    \caption{Paragraph-level automatic evaluation results on the full par3 development and evaluation corpus. Rows are sorted by MetricX-QE, ascending. Row colors distinguish P1, P2, P3, and HT systems. Scores are unweighted means over book-level system scores across both splits. Higher is better for COMET-22 and COMETKiwi; lower is better for MetricX and MetricX-QE. Bold indicates the best MetricX-QE score. The Human eval. column marks systems used in the human evaluation. Missing HT reference-based scores are shown as ``--''.}
    \label{tab:appendix_par3_model_metric_results}
\end{table*}

\FloatBarrier

\FloatBarrier

\end{document}